\documentclass{article}
\usepackage[final]{colm2026_conference}

\usepackage{hyperref}
\usepackage{url}
\usepackage{booktabs}
\usepackage{lineno}
\usepackage{color}
\usepackage{tcolorbox}
\usepackage{caption}
\usepackage{multicol}
\usepackage{titletoc}
\tcbuselibrary{breakable, skins}

\newtcolorbox{promptbox}[1]{
    breakable,
    colback=white,
    colframe=black!70,
    fonttitle=\scriptsize\bfseries,
    title=#1,
    toprule=1.5pt,
    bottomrule=0.5pt,
    left=3pt, right=3pt, top=3pt, bottom=3pt,
}
\newtcolorbox{systembox}{
    breakable,
    colback=white,
    colframe=black!40,
    leftrule=3pt,
    rightrule=0.4pt,
    toprule=0.4pt,
    bottomrule=0.4pt,
    fontupper=\scriptsize,
    title={\scriptsize\textbf{System}},
    left=3pt, right=3pt, top=2pt, bottom=2pt,
    before skip=2pt, after skip=2pt,
}
\newtcolorbox{userbox}{
    breakable,
    colback=white,
    colframe=black!20,
    leftrule=3pt,
    rightrule=0.4pt,
    toprule=0.4pt,
    bottomrule=0.4pt,
    fontupper=\scriptsize,
    title={\scriptsize\textbf{User}},
    left=3pt, right=3pt, top=2pt, bottom=2pt,
    before skip=2pt, after skip=2pt,
}

\newcommand{\hintcell}[1]{%
  \begin{minipage}[t]{\linewidth}
  \raggedright
  \ttfamily\scriptsize
  #1
  \end{minipage}%
}

\definecolor{ForestGreen}{rgb}{0.133,0.545,0.133}
\definecolor{BrickRed}{rgb}{0.796,0.255,0.329}
\definecolor{darkblue}{rgb}{0,0,0.5}
\definecolor{mygreen}{rgb}{0.133,0.545,0.133}

\usepackage{soul}

\definecolor{enthl}{RGB}{221,235,247}   % soft blue   - O_ent atomic entity
\definecolor{anshl}{RGB}{237,226,244}   % soft purple - O_ans atomic entity
\definecolor{comphl}{RGB}{219,242,231}  % soft green  - composed substitution

\DeclareRobustCommand{\hlent}[1]{{\sethlcolor{enthl}\hl{#1}}}
\DeclareRobustCommand{\hlans}[1]{{\sethlcolor{anshl}\hl{#1}}}
\DeclareRobustCommand{\hlcomp}[1]{{\sethlcolor{comphl}\hl{#1}}}

\hypersetup{
  colorlinks=true,
  citecolor=darkblue,
  linkcolor=darkblue,
  urlcolor=darkblue
}

\usepackage{amsmath,amsfonts,bm}

\def\eqref#1{equation~\ref{#1}}

\def\1{\bm{1}}

\DeclareMathAlphabet{\mathsfit}{\encodingdefault}{\sfdefault}{m}{sl}
\SetMathAlphabet{\mathsfit}{bold}{\encodingdefault}{\sfdefault}{bx}{n}

\usepackage{xcolor}
\usepackage{graphicx}
\usepackage{tabularx}
\usepackage{amsmath}
\usepackage[capitalize]{cleveref}
\usepackage{mathtools}
\usepackage{multirow}
\usepackage{setspace}
\hypersetup{colorlinks,linkcolor={blue},citecolor={blue},urlcolor={blue}}

\usepackage{latexsym}
\usepackage{comment}
\usepackage{tabulary}
\usepackage{amsfonts}
\usepackage{array}

\usepackage[T1]{fontenc}
\usepackage[utf8]{inputenc}
\usepackage{listings}
\usepackage{pifont}
\usepackage{inconsolata}

\usepackage{colortbl}
\usepackage{enumitem}
\usepackage{makecell}
\usepackage{arydshln}

\usepackage{subcaption}
\usepackage{wrapfig}
\usepackage{float}

\newcommand{\algname}{\texttt{Retail\-RAG}}
\newcommand{\dataname}{\texttt{Retail\-Bridge}}

\definecolor{absgray}{RGB}{242,243,245}
\definecolor{metablue}{RGB}{0,102,204}
\definecolor{lightred}{RGB}{255, 204, 204}
\definecolor{lightgreen}{RGB}{204, 255, 204}
\definecolor{lightyellow}{RGB}{255, 255, 204}

\newcommand{\customabstractpage}{
\begin{tcolorbox}[
    enhanced,
    colback=absgray,
    colframe=absgray,
    boxrule=0pt,
    arc=8pt,
    left=3mm,
    right=3mm,
    top=3mm,
    bottom=3mm
]

{\Large\bfseries
Toward Agentic Retrieval-Augmented Generation for Enterprise Retail Workflows over Heterogeneous Data Sources
\par}

\vspace{3mm}

\textbf{Jeonghwan Choi}$^{1*}$, \textbf{Jibin Hwang}$^{1*}$, \textbf{Hayoon Park}$^{1}$, \\ \textbf{Hyeonjae Cheon}$^{3}$, \textbf{Hwanjun Song}$^{1,2\dagger}$\par

\vspace{1mm}

$^{1}$Korea Advanced Institute of Science and Technology (KAIST), $^{2}$Cluvion, $^{3}$GS Retail\\\par

\vspace{4mm}

\noindent
Enterprise retail workflows require agents to chain evidence across heterogeneous sources, from structured product and order records to unstructured inquiries and reviews, yet existing agentic RAG benchmarks rarely capture this challenge. To enable controlled evaluation of such workflows, we introduce \dataname{}, a compositional retail benchmark built from persona-driven simulations. Its questions are constructed by linking entities and intermediate answers across heterogeneous sources, requiring agents to identify the relevant sources, carry evidence between them, and produce a grounded answer. To systematically evaluate LLMs on these retail workflows, we develop \algname{}, a retail-aware framework equipped with auto-schema briefing, source-routed plan chaining, and diagnostic plan revision. \algname{} consistently outperforms existing agentic RAG baselines across all evaluation axes, serving as a strong reference system for assessing LLM agents on retail workflows. On \dataname{}, evaluating ten LLM backbones with \algname{} reveals multi-source chaining as the dominant bottleneck across providers and model scales, suggesting that reliable retail agents require explicit cross-source orchestration rather than scaling alone. Code and data are publicly available at \url{https://retailbridge-kdd.github.io}.

\vspace{4mm}

\noindent
\begin{minipage}[t]{0.7\textwidth}
{\small
\textbf{Date:} Oct 16, 2026 \par
\textbf{Correspondence:} Hwanjun Song at {\color{metablue}\href{mailto:songhwanjun@kaist.ac.kr}{songhwanjun@kaist.ac.kr}} \par
\textbf{First Authors:} 
\\Jeonghwan Choi at {\color{metablue}\href{mailto:hwani.choi@kaist.ac.kr}{hwani.choi@kaist.ac.kr}}, \\ Jibin Hwang at {\color{metablue}\href{mailto:jibinhwang@kaist.ac.kr}{jibinhwang@kaist.ac.kr}} \par
\textbf{Code \& Data:} {\color{metablue} \url{https://retailbridge-kdd.github.io}}
}
\end{minipage}
\hfill
\begin{minipage}[t]{0.27\textwidth}
\vspace*{-0.3cm}
\raggedleft
\includegraphics[width=1.0\linewidth]{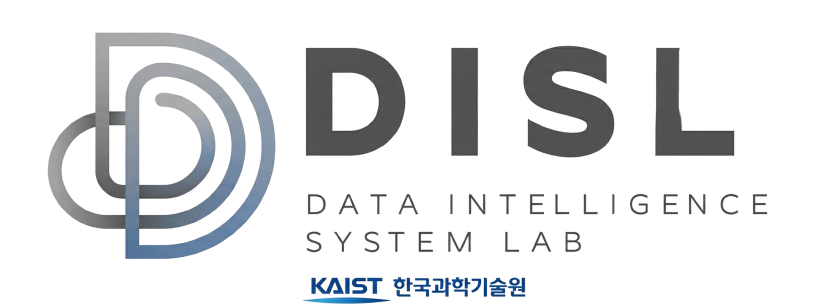}
\end{minipage}

\end{tcolorbox}
}

\begin{document}

\lhead{Accepted to KDD 2027 Datasets and Benchmarks Track.}
\thispagestyle{empty}

\vspace*{-1.3cm}
\customabstractpage
{\renewcommand{\thefootnote}{}\footnotetext{$^{*}$Equal contribution. $^{\dagger}$Corresponding author.}}
%\newpage

\section{Introduction}
\label{sec:introduction}

Retail has become a major application domain for large language model (LLM) agents, spanning e-commerce assistance, seller analytics, and customer support \cite{bianchi2024negotiationarena, wang2025customerr1, choi2026retailsim}. In this setting, reliable responses require grounding in internal enterprise records such as product catalogs, order histories, and customer interactions, making retrieval-augmented generation (RAG) \cite{gao2024retrieval, yan2025hetarag} the core paradigm. However, real retail queries rarely reside in a single repository; they span \textit{heterogeneous} sources combining structured tables and unstructured text. Answering them requires agents to route across sources, bridge isolated evidence, and synthesize a final answer.

Several agentic RAG methods have been proposed, including reasoning-and-acting agents \cite{yao2023react, nahid2026prism}, plan-then-retrieve agents \cite{lee2024planrag, yang2025pasemiqa}, and iterative retrieval agents \cite{fang2025kirag, asl2025fair}. However, they remain limited in two ways. First, they operate on \emph{single} homogeneous corpora, unlike real retail workflows that require chaining evidence across structured tables and unstructured text. Second, they are evaluated mainly on \emph{open-domain} sources such as the Web or Wikipedia, leaving their behavior on enterprise retail data largely unverified. %
Recent enterprise-RAG benchmarks highlight this gap, with top agentic systems achieving only $32\%$ on heterogeneous deep-search tasks \cite{choubey2025benchmarking} and text-to-SQL agents only $20$--$25\%$ on realistic database interactions \cite{huo2025bird}. Therefore, current LLM agents struggle to navigate complex schemas, coordinate evidence across heterogeneous sources, and execute multi-step enterprise workflows reliably.

\begin{figure*}[t]
\centering
\includegraphics[width=\textwidth]{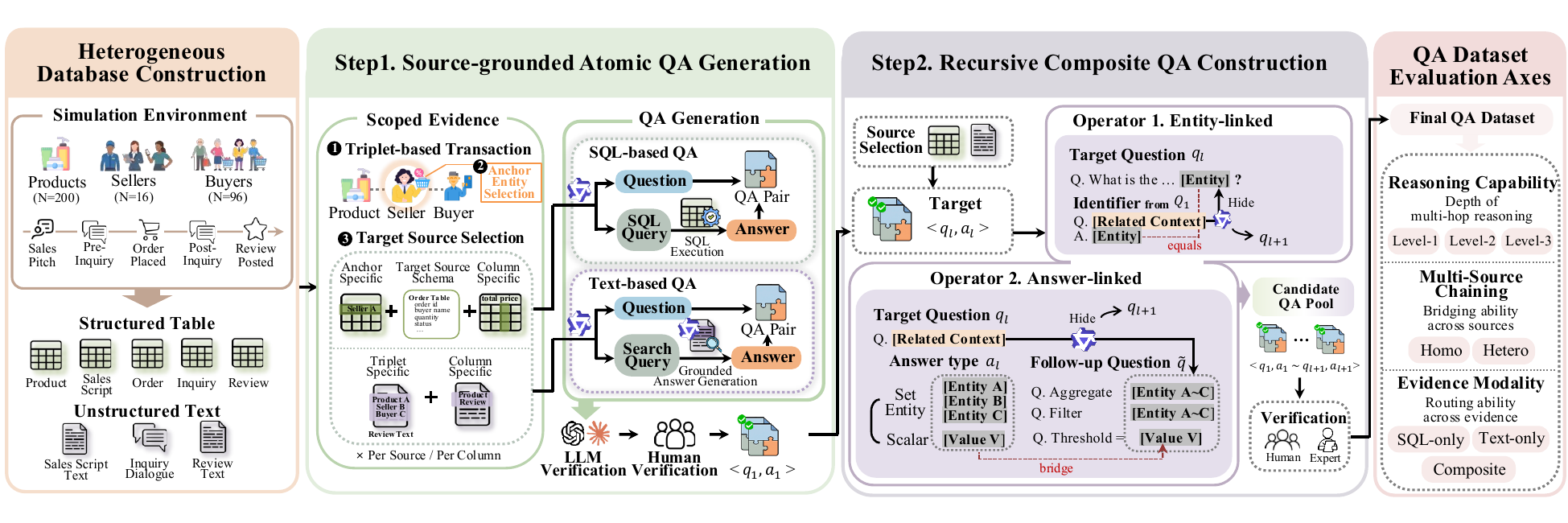}
\vspace{-0.8cm}
\caption{Overview of \dataname{}. Starting from persona-driven retail simulations, \dataname{} constructs a \emph{heterogeneous} retail database, then (Step 1) generates source-grounded \emph{atomic} QAs and (Step 2) recursively composes them via \emph{entity-linked} and \emph{answer-linked} operators into multi-source, multi-hop queries stratified along three evaluation axes.}
\label{fig:overview}
\vspace{-0.6cm}
\end{figure*}

%On the "method" side, existing agentic RAG frameworks are not designed for heterogeneous, schema-rich enterprise environments, often assuming a single corpus or supporting only one source modality. Addressing this gap requires both a dedicated benchmark and framework for enterprise retail workflows.
The core bottleneck is twofold. On the "data" side, real enterprise retail data and high-difficulty workflow queries are rarely accessible, limiting controllable evaluation of multi-source reasoning. On the "framework" side, existing agentic RAG systems are poorly suited to heterogeneous, schema-rich retail environments, often assuming a single corpus or a single source modality. Consequently, benchmarking current LLMs on enterprise retail workflows requires both a realistic, controllable data testbed and a retail-aware agentic framework that can faithfully orchestrate cross-source reasoning; otherwise, measured performance may conflate LLM capability with data or framework limitations.

We first address the data bottleneck by introducing \dataname{}, a multi-source retail QA benchmark for evaluating agentic RAG systems on enterprise retail workflows. Unlike generic agentic RAG benchmarks such as HERB \cite{choubey2025benchmarking} and SPARTA \cite{park2026sparta}, \dataname{} is built around two key ideas, as illustrated in Figure \ref{fig:overview}. First, we simulate realistic enterprise retail environments by mapping persona-driven seller-buyer interactions into \emph{heterogeneous} retail data, where each interaction is stored as either structured database records or unstructured text. Second, we construct retail queries with controllable difficulty through a two-step compositional synthesis pipeline. In the first step, we generate and verify source-grounded \emph{atomic QAs} from individual structured or unstructured sources, providing reliable single-source questions and a controllable pool of composition anchors.  In the second step, we recursively compose these atomic QAs using \emph{entity-linked} and \emph{answer-linked} operators, which respectively replace an explicit entity or intermediate result with evidence that must be resolved from another source. This process yields $1{,}680$ multi-source, multi-hop queries across three difficulty levels, with difficulty systematically controlled by reasoning depth and source transitions. Consequently, \dataname{} supports diagnostic evaluation beyond aggregate accuracy, revealing performance degradation across difficulty levels, cross-source joining ability, and pipeline failure points.

On the framework side, we develop \algname{}, the first retail-aware agentic RAG framework for systematically benchmarking how different LLM backbones handle enterprise retail workflows. Rather than assuming a single corpus or source modality, \algname{} operationalizes the key capabilities required by enterprise retail workflows: (i) \emph{auto-schema briefing}, which provides the planner with database schemas, entity links, and analytical constraints before planning; (ii) \emph{source-routed plan chaining}, which explicitly routes each step to structured or unstructured sources and carries intermediate results across SQL-to-text-to-SQL chains; and (iii) \emph{diagnostic plan revision}, which identifies and locally repairs failed steps without restarting the entire workflow. By reducing framework-induced failures in schema understanding, cross-source orchestration, and plan recovery, \algname{} enables a more faithful evaluation of LLM capabilities on enterprise retail workflows.

%On the framework side, we introduce \algname{}, a retail-aware agentic RAG framework designed for the heterogeneous data environment of real-world enterprise retail. Unlike prior agentic RAG systems that assume a single corpus, aggregate evidence post-hoc, or handle only structured tables, \algname{} is built on three retail-specific designs: (i) \emph{auto-schema briefing} that gives the planner a schema-aware view of the retail database before planning begins; (ii) \emph{source-routed plan chaining}, where each step declares its target source and passes results as inline variables, enabling explicit SQL-to-text-to-SQL chains within a single planner; and (iii) \emph{diagnostic plan revision} that locally repairs failed steps instead of restarting. Together, these make \algname{} the first agentic RAG framework natively designed for enterprise retail workflows.

We conduct two complementary evaluations on \dataname{}. First, we compare \algname{} with representative agentic RAG systems to establish a strong reference system for enterprise retail workflows. \algname{} consistently outperforms baselines such as MAC-SQL \cite{wang2025macsql} and HetaRAG \cite{yan2025hetarag}, particularly on queries requiring multi-source bridging. Second, using \algname{} as a common evaluation framework, we benchmark a diverse set of open-source and proprietary LLM backbones. Across models, accuracy drops substantially as queries require more source transitions and deeper reasoning chains, with failures concentrated in multi-source chaining (from Level-1 to Level-3). These results identify cross-source orchestration, rather than model scale alone, as the central bottleneck in enterprise retail RAG.
%We use \dataname{} to evaluate \algname{} from two angles: against representative agentic RAG baselines to validate its retail-specific design, and as a backbone for benchmarking a wide range of LLMs. From the framework perspective, \algname{} significantly outperforms baselines such as MAC-SQL \cite{wang2025macsql} and HetaRAG \cite{yan2025hetarag}, with consistent gains on queries requiring multi-source bridging. From the model perspective, both open-source and proprietary LLMs lose roughly half of their accuracy as queries demand more source joins and longer reasoning chains (Level-1 to Level-3), with errors concentrated in multi-source chaining rather than single-source retrievals, indicating that reliable retail LLM agents will require explicit cross-source orchestration, not scaling alone.
Our main contributions are: 

\begin{itemize}[leftmargin=*]
    \item \dataname{}, the first multi-source retail QA benchmark, built by mapping persona-driven seller-buyer simulations into a heterogeneous enterprise database without privacy constraints.
    \item A compositional QA synthesis pipeline whose entity-linked and answer-linked operators recursively yield  multi-source, multi-hop retail queries across three levels of compositional depth.
    \item A multi-stage validation of benchmark quality and practical relevance through human review, retail domain-expert assessment, and comparison with a real retail dataset.
    \item \algname{}, the first retail-aware agentic RAG framework, integrating auto-schema briefing, source-routed plan chaining, and diagnostic plan revision.
    \item A comprehensive benchmark of diverse agentic RAG baselines and LLM backbones, exposing multi-source chaining as the fundamental bottleneck of current retail LLM agents.
\end{itemize}

\section{Related Work}
\label{sec:related-work}

\paragraph{LLM Agents for Enterprise Retail.} Retail has become a major application domain for LLM agents, spanning product search \cite{wang2025ecomscriptbench, wang2025shoppingbench}, negotiation \cite{xia2024bargaining, bianchi2024negotiationarena}, and shopping assistance \cite{wang2025customerr1, zhou2026mixecom}, while persona-driven simulation increasingly enables controlled seller--buyer interactions across diverse retail scenarios \cite{bansal2025magentic, zhang2026shopr1, choi2026retailsim}. These efforts target the customer-facing surface, deepening the realism of a single interaction touchpoint with self-contained inputs. A complementary direction instead aims to raise the productivity of sellers and analysts who run the platform, through business-oriented text-to-SQL agents over enterprise databases \cite{li2025agent, lei2025spider, huo2025bird} and heterogeneous enterprise RAG frameworks \cite{choubey2025benchmarking, yan2025hetarag}. Yet they target generic business or software schemas rather than retail-specific seller, product, and buyer structures, and rely on a single retrieval mode, leaving the cross-source chains central to retail workflows unaddressed.

\paragraph{Agentic RAG Systems.} RAG grounds language model outputs in external evidence \cite{lewis2020rag, gao2024retrieval}, and agentic extensions enable multi-hop retrieval through reasoning-and-acting agents that interleave retrieval with reasoning traces \cite{yao2023react, nahid2026prism}, plan-then-retrieve agents that decompose a query before retrieving \cite{wang2023planandsolve, lee2024planrag, yang2025pasemiqa}, and iterative retrieval agents that refine subsequent searches with intermediate evidence \cite{harsh2023ircot, zhang2024beam, fang2025kirag, asl2025fair}. Yet these systems assume a single homogeneous text corpus over open-domain passages such as Wikipedia, ignoring heterogeneous stores where structured databases and unstructured text coexist, and aggregate evidence post-hoc rather than chaining intermediate results as inline arguments. 

Such limitations make them ill-suited for enterprise retail, where queries must traverse heterogeneous sources with intermediate evidence carried across them.

\paragraph{Heterogeneous QA Benchmarks.} Evaluating agents on realistic enterprise workflows requires QA benchmarks in which evidence is genuinely scattered across heterogeneous sources, rather than packaged within a single document. Prior work has progressed from single-passage extraction \cite{pranav2016squad} and text-only multi-hop reasoning \cite{yang2018hotpotqa, harsh2022musique} to benchmarks that pair structured tables with linked passages \cite{chen2020hybridqa, chen2021ottqa} and add numerical reasoning over tables and paragraphs \cite{zhu2021tatqa, zhao2022multihiertt}. More recent efforts move toward independently stored sources in enterprise or tree-structured settings \cite{choubey2025benchmarking, park2026sparta}. However, their heterogeneity is co-located and explicitly linked through hyperlinks or shared headers, so multi-source chaining reduces to following an existing link. Enterprise retail instead requires resolving such links implicitly across independently maintained sources.

\section{RetailBridge: Multi-Source QA over Heterogeneous Retail Sources}
\label{sec:method}
\dataname{} is a multi-source retail QA benchmark whose queries require bridging evidence across independently maintained structured and unstructured sources. We construct it through (i) a heterogeneous retail database from persona-driven simulations (Section \ref{subsec:source_construction}), and (ii) a compositional pipeline that synthesizes {atomic} QAs and composes them via {entity-linked} and {answer-linked} operators into multi-source, multi-hop queries (Section \ref{subsec:compostional_qa}).

\subsection{From Simulation to Retail Database}
\label{subsec:source_construction}
Real-world enterprise retail data are inaccessible due to privacy constraints, yet faithfully reproducing their heterogeneity is essential for evaluating multi-source retail QA. Inspired by recent persona-driven retail simulators \cite{choi2026retailsim}, we simulate multi-stage seller--buyer interactions, with each stage mapped onto a heterogeneous database that mirrors a real enterprise stack. Notably, the simulator has been validated to reproduce real-world economic regularities, grounding our database in realistic retail dynamics.

% We address this by mapping RetailSim's multi-stage seller--buyer interactions \cite{choi2026retailsim} onto a heterogeneous database that mirrors a real enterprise stack.

\paragraph{Persona-Driven Retail Trajectories.}
Following \citet{choi2026retailsim}, we simulate seller--buyer interactions conditioned on structured personas specifying {each participant's demographics and behavioral tendencies.} Anchored on a shared product catalog, each trajectory progresses through five stages---sales pitch, pre-purchase inquiry, order, post-purchase inquiry, and review---with each stage building on prior outcomes so that a trajectory forms a coherent retail history. These trajectories provide the basis for the multi-source heterogeneous database $\mathcal{D}$, with every record traceable to its originating interaction. Details on the simulation are in Appendix \ref{appendix:simulation_setup}.

% Following \citet{choi2026retailsim}, we simulate seller--buyer interactions conditioned on structured personas. Anchored on a shared product catalog, each trajectory progresses through five stages---sales pitch, pre-purchase inquiry, order, post-purchase inquiry, and review---with each stage building on prior outcomes so that a trajectory forms a coherent retail history. These trajectories provide the basis for multi-source heterogeneous database $\mathcal{D}$; details on the simulation setup are provided in Appendix \ref{appendix:simulation_setup}.

\paragraph{Heterogeneous Source Mapping.}
We construct $\mathcal{D}$ from the simulated trajectories together with their anchoring product catalog through two steps: (i) \emph{projection}, which labels each item with its retail stage $s$ (\emph{e.g.}, sales pitch, order) and its available heterogeneity axes $h \in \{\textsc{tab}, \textsc{txt}\}$, and (ii) \emph{aggregation}, which collects items sharing $(s, h)$ into a source $D_{s,h}$, yielding $\mathcal{D} = \{D_{s,h}\}$ of eight sources:

\begin{itemize}[leftmargin=*, topsep=1pt]
    \item \textbf{Tabular Sources} ($h{=}\textsc{tab}$): five tables---two for product catalogs and order logs, and three for the relational metadata of sales scripts, reviews, and customer-service inquiries.
    \item \textbf{Textual Sources} ($h{=}\textsc{txt}$): three text collections with only the raw text of sales scripts, reviews, and customer-service inquiries.
\end{itemize}
{Crucially, these sources remain independently structured yet linkable implicitly through shared identifiers of sellers, products, and buyers, which is exactly the cross-source joining that \dataname{} queries are designed to exercise. See Figure \ref{fig:dbschema} in Appendix \ref{appendix:schema_database} for the detailed heterogeneous database schema.}
% These sources can be linked through shared sellers, products, and buyers, enabling QA to traverse evidence across them. See Figure \ref{fig:dbschema} in Appendix \ref{appendix:schema_database} for the detailed heterogeneous database schema.

\subsection{Compositional QA Synthesis}
\label{subsec:compostional_qa}
We synthesize \dataname{}'s queries in two steps: (Step 1) generating \emph{atomic} QAs answerable from a single source (Section \ref{subsec:atomic_qa}), and (Step 2) recursively composing them via \emph{entity-linked} and \emph{answer-linked} operators into multi-source, multi-hop queries across three difficulty levels (Level-1 to Level-3) (Section \ref{subsec:multi_source_qa}).

\begin{table}[t]
\centering
\small
\caption{Examples of the two composition operators in \dataname{}. Colors mark each operator's bridged value (\hlent{$O_\mathrm{ent}$}, \hlans{$O_\mathrm{ans}$}) and the referring expression \hlcomp{$q^{\text{ref}}$} hiding it in $q_{\ell+1}$.}
\vspace*{-0.2cm}
\begin{tabularx}{\columnwidth}{@{}c l X@{}}
\toprule
\textbf{Operator} & \textbf{Query} & \textbf{Example} \\
\midrule
\multirow{9}{*}{$O_\mathrm{ent}$}
& \makecell[tl]{$q_1$\\\tiny(Level-1)}
& Which seller broadcast the Libby's Mixed Vegetables (8.5 oz, Pack of 12)?\quad\textit{A:} \hlent{Robert A.} \\
& \makecell[tl]{$q_\ell$\\\tiny(Level-$\ell$)}
& What is the average product review rating for \hlent{Robert A.}?\quad\textit{A:} 3.41 \\
& \makecell[tl]{$q_{\ell+1}$\\\tiny(Composite)}
& What is the average product review rating for \hlcomp{the seller who broadcast the Libby's Mixed Vegetables}?\quad\textit{A:} 3.41 \\
& \textit{Effect}
& Single-entity bridge (\textsc{Script} $\rightarrow$ \textsc{Review}); a singular $q^{\text{ref}}$ hides one entity \\
\midrule
\multirow{10}{*}{$O_\mathrm{ans}$}
& \makecell[tl]{$q_\ell$\\\tiny(Level-$\ell$)}
& Which of Mark A.'s products received the highest rating in product reviews?\quad
\textit{A:} \{\hlans{Stretchable Candy Necklace, \ldots}\} \\
& \makecell[tl]{$\tilde{q}_\ell$\\\tiny(Follow-up)}
& What is the total revenue from orders of \hlans{these products}?\quad\textit{A:} \$20{,}712.52 \\
& \makecell[tl]{$q_{\ell+1}$\\\tiny(Composite)}
& What is the total revenue from orders of \hlcomp{the products that received the highest rating in product reviews among Mark A.'s products}?\quad\textit{A:} \$20{,}712.52 \\
& \textit{Effect}
& Set bridge (\textsc{Review} $\rightarrow$ \textsc{Order}); a plural $q^{\text{ref}}$ hides the whole answer set \\
\bottomrule
\end{tabularx}
\label{tab:composition_operators}
\vspace*{-0.2cm}
\end{table}

\subsubsection{Atomic QA Generation}
\label{subsec:atomic_qa}
Atomic QAs serve two roles in \dataname{}: as the Level-1 base (answerable from a single source) and as anchors for multi-source composition. To ensure their reliability, {we (i) select a piece of evidence from a single source, (ii) generate each atomic QA grounded in it, and (iii) validate the result with two complementary filters, retaining only QAs that pass both.}

\paragraph{Evidence Selection.}
To cover varied perspectives, we sample triplets $t = \langle\text{seller}, \text{product}, \text{buyer}\rangle$ from the simulated trajectories and treat each entity in $t$ as a question anchor. For each anchor, we identify a source $D_{s,h}$ in which the anchor appears and select a specific piece of evidence $e$ from the source---a table cell when $h{=}\textsc{tab}$ (\emph{e.g.}, the product's price), or a self-contained snippet when $h{=}\textsc{txt}$ (\emph{e.g.}, a phrase from the buyer's product review)---yielding a gold answer $a$. For numerical questions, where LLM-inferred answers are prone to hallucination \cite{li2025exposing}, we construct an additional source $D_{\text{metric}}$ storing metrics (\emph{e.g.}, aggregations, rankings, rates) derived from the tabular sources, from which we draw $e$; this source is used \emph{only} during QA construction and excluded from $\mathcal{D}$ at test time.

\paragraph{QA Generation.}
From each piece of evidence $e$, an LLM generates a Level-1 question $q_1$ (atomic) and a heterogeneity-specific search query $q_{\text{search}}$, executed on $D_{s,h}$ to yield an answer $a_1$:
% , and a heterogeneity-specific search query $q_{\text{search}}$:
\[
(q_1, q_{\text{search}}, a_1) = \mathrm{LLM}(e),
\]
where $q_{\text{search}}$ is a SQL query when $h{=}\textsc{tab}$ and a retrieval query when $h{=}\textsc{txt}$. To diversify phrasing and reduce model-specific bias, we split the sampled triplets evenly across three strong LLMs (Qwen3.6-Max, DeepSeek-V4-Pro, Gemini-3.1-Pro) under the same prompt\footnote{For reproducibility, all prompts used throughout the \dataname{} pipeline are provided in Tables \ref{tab:atomic_generation}--\ref{tab:final_answer_generation} of Appendix \ref{appendix:prompt}.}. We store each atomic QA as a tuple $(q_1, q_{\text{search}}, a_1, s, h)$, {tagging it with the originating stage $s$ and heterogeneity axis $h$, which enables \dataname{} to precisely control the source diversity and evidence modality of every query. }

\paragraph{Atomic QA Verification Filters.}
Since unreliability in atomic QAs would propagate through composition, we retain $q_1$ only if it passes two complementary filters:{
\begin{itemize}[leftmargin=*, topsep=1pt]
    \item \textbf{Executable Verification}: we automatically filter out QAs whose $q_{\text{search}}$ fails to execute on $D_{s,h}$ (or $D_{\text{metric}}$ for numerical questions), ensuring every retained QA is answerable from its source.
    \item \textbf{LLM Verification}: we further filter QAs through two judges (GPT-5.5, Claude Opus 4.7), retaining only QAs where both independently confirm that $q_{\text{search}}$ targets the entity queried by $q_1$ and that $a_1$ is grounded in the retrieved evidence.
\end{itemize}
The verified atomic QAs form the atomic pool $Q_1$ for multi-source composition, spanning all eight sources and both heterogeneity axes so that composition can bridge any source combination.}

% \paragraph{Atomic QA Verification Filters.}
% We retain $q_1$ only if it passes two complementary filters:
% \begin{itemize}[leftmargin=*, topsep=1pt]
%     \item \emph{\textbf{Executable verification}}: $q_{\text{search}}$ executes on $D_{s,h}$ (or $D_{\text{metric}}$ for numerical questions) and yields a non-empty, substantive result $a_1$.
%     \item \emph{\textbf{LLM verification}}: both judges (GPT-5.5, Claude Opus 4.7) independently confirm that $q_1$, $q_{\text{search}}$, and $a_1$ are mutually consistent: $q_{\text{search}}$ targets the entity queried by $q_1$, and $a_1$ is grounded in the retrieved evidence.
% \end{itemize}
% The verified atomic QAs form the base atomic pool $Q_1$ for multi-source composition.

\definecolor{enthl}{RGB}{221,235,247}   % soft blue   - O_ent atomic entity
\definecolor{anshl}{RGB}{237,226,244}   % soft purple - O_ans atomic entity
\definecolor{comphl}{RGB}{219,242,231}  % soft green  - composed substitution

\subsubsection{Recursive QA Composition}
\label{subsec:multi_source_qa}
Atomic QAs alone stay within a single source, whereas real retail questions chain evidence across sources. To bridge this gap, we build a Level-$L$ composite query $q_L$ from an atomic $q_1$ by recursively lifting $q_\ell$ to $q_{\ell+1}$, within or across sources, via one of two operators. The operator is chosen by the type of the value being bridged: (i) \emph{entity-linked} composition $O_\mathrm{ent}$ hides a single entity, while (ii) \emph{answer-linked} composition $O_\mathrm{ans}$ hides a set- or scalar-valued answer (see Table \ref{tab:composition_operators} for operator examples).

\paragraph{Entity-Linked Composition.}
$O_\mathrm{ent}(q_1, q_\ell)$ generates $q_{\ell+1}$ when $a_1$ equals an entity in $q_\ell$. Since an entity admits a noun-phrase referent, we prompt an LLM to rewrite $q_\ell$, hiding that entity behind a referring expression $q^{\text{ref}}_{1}$ derived from $q_1$:
\[
q_{\ell+1} = q_{\ell}\bigl[\,\text{entity} \mapsto q^{\text{ref}}_{1}\,\bigr],
\]
where $[x \mapsto y]$ denotes hiding $x$ behind $y$. The chain links $q_1$, which identifies the entity, with $q_\ell$, which produces the final answer.

\paragraph{Answer-Linked Composition.}
$O_\mathrm{ans}(q_\ell)$ generates $q_{\ell+1}$ when $a_\ell$ is a set or scalar. Since such answers admit no noun-phrase referent, we first prompt an LLM to write a follow-up question $\tilde{q}_\ell$ that takes $a_\ell$ as an explicit answer entity (\emph{e.g.}, asking the total revenue from orders of the answered product set), and then rewrite $\tilde{q}_\ell$, hiding $a_\ell$ behind a referring expression $q^{\text{ref}}_{\ell}$ derived from $q_\ell$:
\[
q_{\ell+1} = \tilde{q}_\ell\bigl[\,a_\ell \mapsto q^{\text{ref}}_{\ell}\,\bigr].
\]
The chain links $q_\ell$, which produces the bridging value, with $\tilde{q}_\ell$, which yields the final answer. Further examples across answer types are provided in Appendix \ref{appendix:answertype_qa}.

\paragraph{Recursive Extension.}
Each step lifts $q_\ell$ to $q_{\ell+1}$ via $O_\mathrm{ent}$ or $O_\mathrm{ans}$. The recursion stops at the maximum level $L{=}3$, beyond which composed queries become unrealistic for retail. Candidates pass the same two-filter verification scheme as atomic QAs (Section~\ref{subsec:atomic_qa}), augmented with composite-specific checks on chain-correctness and referent unambiguity. For diversity in the final benchmark, we embed each query with Qwen3-8B-Embedding and apply farthest-point sampling per level. {Throughout this process, every query carries its reasoning level, source diversity, and evidence modality, stratifying \dataname{} along three diagnostic axes.}

\subsection{Human Verification of RetailBridge}
\label{subsec:quality_assurance}
{To ensure the reliability of \dataname{} beyond automated verification, we apply a two-stage human verification process with expert annotators, verifying the correctness of each query (Stage 1) and its alignment with real retail workflows (Stage 2). Detailed verification processes are in Appendix \ref{appendix:human_annotation}.}

\paragraph{Stage 1: NLP-Expert Selection.}
{All the queries are independently reviewed by three graduate-level NLP experts along four criteria---query quality, answer correctness, evidence-question alignment, and modality-source consistency. Their judgments show high inter-annotator agreement (Gwet's AC1 $=$ 0.89), and only unanimously approved queries are retained, leaving the final $1{,}680$ queries---a scale comparable to recent RAG benchmarks \cite{lei2025spider, huo2025bird}--- with three query difficulties from Level-1 to Level-3.}

\paragraph{Stage 2: Domain-Expert Verification.}
{To verify that composed queries reflect questions asked in real retail operations rather than arbitrary compositional depth, a domain expert with over ten years of industry experience at GS Retail\footnote{GS Retail is a major South Korean retailer operating GS25 convenience stores, GS THE FRESH supermarkets, and GS SHOP, with approximately 7,500 employees and annual consolidated revenue of about KRW 12 trillion.} audited a level-stratified sample of 120 queries drawn from the final set, judging each on whether a seller-side analyst would ask it in practice. The expert marked 90\% as questions a seller-side analyst would ask in practice, and this holds even at the deepest composition level, with $97.5\%$ at Level-1, $90.0\%$ at Level-2}, and $82.5\%$ at Level-3, confirming that our composed queries remain aligned with real retail workflows.% (see Appendix \ref{appendix:human_annotation} for details).}

% \paragraph{Diversity Sampling and Human Verification.}
% \textcolor{red}{To select a dispersed subset, we embed each query with Qwen3-8B-Embedding and apply farthest-point sampling per level. The sampled queries are then independently reviewed by three graduate-level NLP experts along four criteria---query quality, answer correctness, evidence-question alignment, and modality-source consistency---with high inter-annotator agreement (Gwet's AC1 $=$ 0.89). This yields $1{,}680$ queries (Level 1--3) that pass all four, a scale comparable to recent QA benchmarks \cite{lei2025spider, huo2025bird}. Beyond NLP expertise, a domain expert with over ten years of industry experience at GS Retail further audited a level-stratified sample of 120 queries drawn from the verified set, judging 90\% (82.5\% at Level-3) plausible as real seller-side analytics questions (see Appendix \ref{appendix:human_annotation} for details).}

\section{RetailRAG: A Retail-Aware Agentic RAG Framework}
\label{sec:framework}
%Alongside \dataname{}, we propose \algname{} as a retail-aware reference framework for evaluating LLMs over heterogeneous retail sources. As shown in Figure \ref{fig:rag_overview}, it addresses domain-specific schemas, SQL-text evidence chains, and costly plan failures through three retail-dedicated stages, namely (i) \emph{auto-schema briefing}, (ii) \emph{source-routed plan chaining}, and (iii) \emph{diagnostic plan revision}.

\begin{wrapfigure}{r}{0.55\textwidth}
\centering
\vspace*{-0.4cm}
\includegraphics[width=\linewidth]{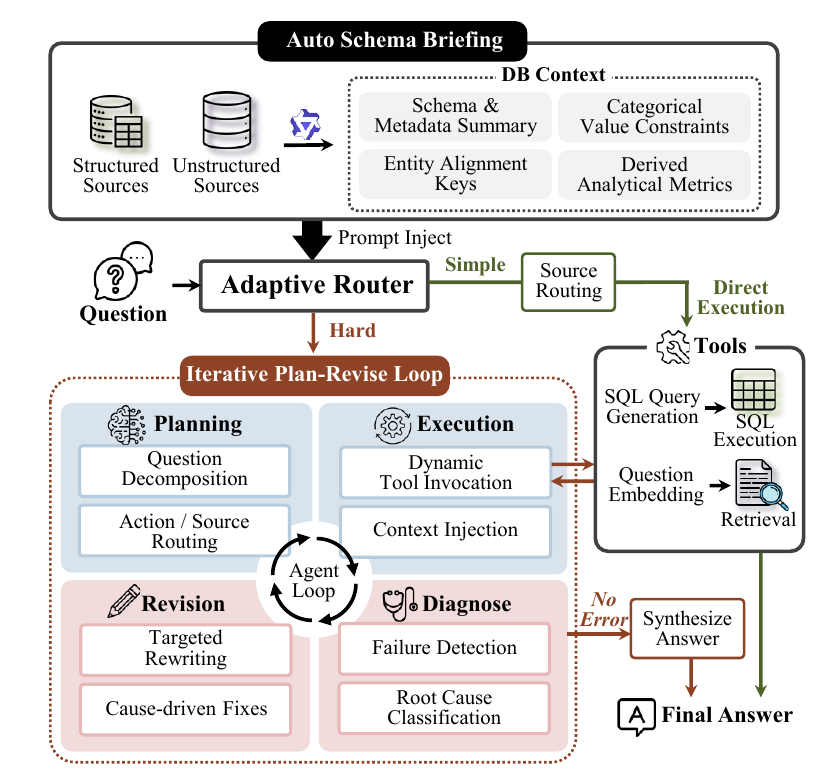}
\vspace{-0.8cm}
\caption{Overview of \algname{}, a retail-aware agentic RAG built on (i) auto-schema briefing, (ii) source-routed plan chaining, and (iii) diagnostic plan revision.}
\vspace{-0.8cm}
\label{fig:rag_overview}
\end{wrapfigure}

Alongside \dataname{}, we propose \algname{} as a retail-aware reference framework for evaluating LLMs over heterogeneous retail sources. Its design is tailored to retail workflows by explicitly modeling retail schemas and entity links, chaining structured records with unstructured sales, inquiry, and review texts, and recovering from failures within these cross-source chains. As in Figure \ref{fig:rag_overview}, it addresses domain-specific schemas, SQL--text evidence chains, and costly plan failures through three stages: (i) \emph{auto-schema briefing}, (ii) \emph{source-routed plan chaining}, and (iii) \emph{diagnostic plan revision}.

%\textcolor{red}{As a reference approach to \dataname{},} we propose \algname{} in Figure \ref{fig:rag_overview}, a three-stage framework for enterprise retail agentic RAG that addresses domain-specific schemas, SQL-text evidence chains, and costly plan failures through (i) \emph{auto-schema briefing}, (ii) \emph{source-routed plan chaining}, and (iii) \emph{diagnostic plan revision}.

\paragraph{Auto-Schema Briefing.}
Effective planning over $\mathcal{D}$ requires structural knowledge---its schemas, cross-source join paths, and KPI derivations (\emph{e.g.}, refund rate)---that a query cannot provide. \algname{} therefore precomputes a compact briefing of $\mathcal{D}$ and injects it into the planner's prompt. To construct this briefing, a deterministic inspection profiles each column as numeric, categorical, or free-text, and an LLM organizes the result into four sections.
\begin{itemize}[leftmargin=*, topsep=1pt]
    \item \textbf{Schema \& Metadata Summary}: the role of each source (\emph{e.g.}, order logs, review texts) and its column types with sample values.
    \item \textbf{Categorical Value Constraints}: valid values per categorical column (\emph{e.g.}, \texttt{status} $\in$ \{\texttt{delivered}, \texttt{refunded}, \texttt{exchanged}\}).
    \item \textbf{Entity Alignment Keys}: shared columns that connect entities across sources (\emph{e.g.}, \texttt{product\_id} linking orders to reviews).
    \item \textbf{Derived Analytical Metrics}: formulas for KPIs that span multiple tables (\emph{e.g.}, refund rate as refunded orders over total orders), automatically derived by the LLM from the schema.
\end{itemize}
Once injected, the briefing provides the planner with the database structure it needs before planning begins; built solely from the schema, it applies to any database beyond retail.

\paragraph{Source-Routed Plan Chaining.}
Multi-source queries chain dependent SQL and text lookups, where a single misrouted step invalidates every step after it. \algname{} therefore determines the chain---each step's source $D_{s,h}$, search query $q_{\text{search}}$, and dependency---in a single pass before execution, through three components:
\begin{itemize}[leftmargin=*, topsep=1pt]
    \item \textbf{Adaptive Router}: for efficiency, an LLM assesses each query's complexity and routes simple ones directly to a single tool call, reserving the planning loop for multi-source ones.
    \item \textbf{Planning}: for complex queries, the planner produces the full chain in a single pass, specifying each step's source $D_{s,h}$ and search query $q_{\text{search}}$, with later steps referencing earlier results as variables, making cross-modality dependencies (\emph{e.g.}, SQL-to-text-to-SQL) explicit.
    \item \textbf{Execution}: each planned step then runs on its declared tool, with prior values deterministically substituted into the next step.
\end{itemize}
This planning keeps every hop aligned with the original question, even as the chain crosses sources and modalities.

% \paragraph{Source-Routed Plan Chaining.}
% Multi-source queries chain SQL and text lookups, where step-by-step decisions cause entity drift. \algname{} therefore plans the entire chain in a single pass for coherent source routing and dependencies.
% \begin{itemize}[leftmargin=*, topsep=1pt]
%     \item \emph{\textbf{Adaptive Router}}: assigns simple queries to a single tool call and hard ones to the plan-execute loop.
    
%     \item \emph{\textbf{Planning}}: the planner produces a full plan, specifying each step's source $D_{s,h}$, and $q_{\text{search}}$.
    
%     \item \emph{\textbf{Execution}}: each step runs on its declared tool, with prior values deterministically substituted into the next step.
% \end{itemize}
% This holistic planning preserves the planner's intended evidence flow across hops.

\paragraph{Diagnostic Plan Revision.}
When a step fails mid-chain, restarting from scratch discards both the completed steps and the plan structure built in the previous stage. \algname{} therefore repairs only the failed step through two stages.
\begin{itemize}[leftmargin=*, topsep=1pt]
    \item \textbf{Diagnose}: an LLM detects failures from execution signals (\emph{e.g.}, empty results, zero aggregates) and assigns a typed cause (\emph{e.g.}, schema mismatch, over-restrictive filter), distinguishing errors that need different repairs.
    \item \textbf{Revision}: guided by the diagnosed cause, an LLM rewrites only the failed step---relaxing a filter, correcting a column, or reformulating the search query---leaving completed steps intact.
\end{itemize}
Localized repair thus recovers from failures at a fraction of the cost of replanning, while keeping the original evidence flow intact.

\section{Evaluation on RetailBridge}
\label{sec:benchmarking}

In evaluation, we first establish \algname{} as the evaluation harness for \dataname{}, as
weaker harnesses conflate framework-induced failures with model limitations. Thus, we compare \algname{} against six representative RAG baselines and ablate its retail-specific designs in Section \ref{subsec:baseline_comparison}. \algname{} outperforms all of them, making it the least confounded harness among those we test. 
Next in Section \ref{subsec:llm_benchmark}, on this basis, we benchmark ten open-source and proprietary LLMs along
\dataname{}. %'s three capability axes---reasoning capability, multi-source
%chaining, and evidence modality ({\color{blue}Section \ref{subsec:llm_benchmark}}).

\paragraph{Benchmark Axes.}
%, comprising 1{,}680 queries
We evaluate along three complementary axes that diagnose distinct capabilities: \emph{(i) reasoning capability}, the depth of multi-hop reasoning required (Level-1, Level-2, Level-3); \emph{(ii) multi-source chaining}, the ability to bridge evidence across sources (Homo.: single source; Hetero.: multiple sources); and \emph{(iii) evidence modality}, the ability to route across structured and unstructured evidence (SQL-only, Text-only, Composite: SQL+Text).

\paragraph{Evaluation Metric.}
We evaluate RAG frameworks and LLMs by accuracy (Acc) and LLM-as-judge, {a standard metric for RAG evaluation \cite{rau2024bergen}.} Acc is a string-containment check (gold answer in prediction), reliable for entity and numeric answers but penalizing valid answers phrased differently; LLM-judge therefore complements it, with Qwen3.6-Plus judging whether each prediction is semantically equivalent to the gold answer. For efficiency, we report the mean number of API calls to the LLM per query. Since agentic frameworks process multi-hop queries through sequential reasoning steps, more API calls increase both latency and cost.

\begin{figure*}[!t]
\centering
\includegraphics[width=\textwidth]{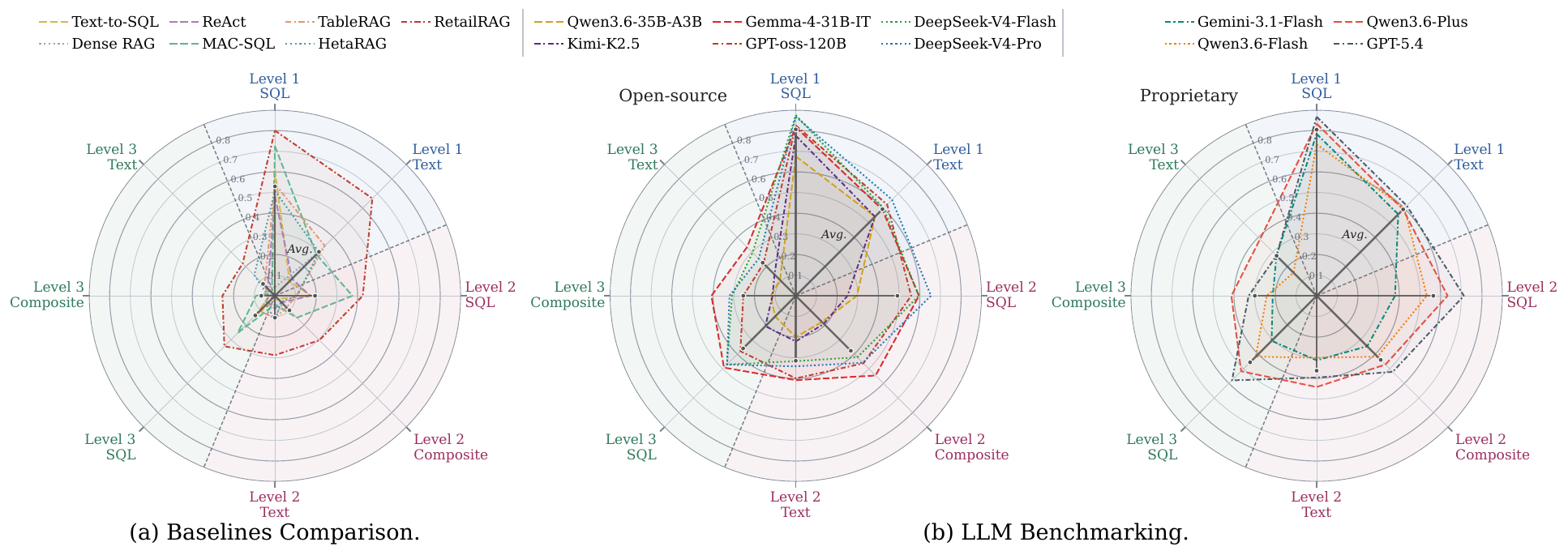}
\definecolor{lvl1bg}{HTML}{EEF4F9}
\definecolor{lvl2bg}{HTML}{F9EEF4}
\definecolor{lvl3bg}{HTML}{EEF9F2}
\definecolor{avgaxis}{HTML}{646566}
\newcommand{\hllvlone}[1]{{\sethlcolor{lvl1bg}\hl{#1}}}
\newcommand{\hllvltwo}[1]{{\sethlcolor{lvl2bg}\hl{#1}}}
\newcommand{\hllvlthree}[1]{{\sethlcolor{lvl3bg}\hl{#1}}}
\newcommand{\avgline}{\textcolor{avgaxis}{\rule[0.3ex]{0.8em}{1pt}}}
\vspace*{-0.7cm}
\caption{LLM-judge accuracy on \dataname{} per evidence modality (SQL / Text / Composite) across reasoning levels (\hllvlone{Level\,1}, \hllvltwo{Level\,2}, \hllvlthree{Level\,3}): (a) \algname{} vs.\ six baselines; and (b) Ten LLM backbones in \algname{}, split into open-source and proprietary. The inner axis (\avgline) indicates the overall performance.}
\label{fig:query_modality}
\end{figure*}

\begin{table*}[t]
\centering
\scriptsize
\caption{Performance on \dataname{}. \emph{Homo.}/\emph{Hetero.}: evidence from one/multiple sources. \emph{Acc}: accuracy. \emph{LLM}: LLM-as-judge accuracy. \emph{API Call}: API calls per query. See Appendix \ref{appendix:stability} for the stability of \algname{} and Appendix \ref{appendix:bird_evaluation} for cross-dataset results.}
\vspace*{-0.25cm}
\setlength{\tabcolsep}{4.3pt}
\renewcommand{\arraystretch}{1.05}
\begin{tabular*}{\textwidth}{l l cc cc cc cc cc cc r}
\toprule
\multirow{3}{*}{Category} & \multirow{3}{*}{Method}
 & \multicolumn{2}{c}{\multirow{2}{*}{Level-1}}
 & \multicolumn{4}{c}{Level-2}
 & \multicolumn{4}{c}{Level-3}
 & \multicolumn{2}{c}{\multirow{2}{*}{Avg.}}
& \multirow{3}{*}{\makecell{API\\Call}} \\
\cmidrule(lr){5-8} \cmidrule(lr){9-12}
 & & &
 & \multicolumn{2}{c}{Homo.} & \multicolumn{2}{c}{Hetero.}
 & \multicolumn{2}{c}{Homo.} & \multicolumn{2}{c}{Hetero.}
 & & & \\
\cmidrule(lr){3-4} \cmidrule(lr){5-6} \cmidrule(lr){7-8} \cmidrule(lr){9-10} \cmidrule(lr){11-12} \cmidrule(lr){13-14}
 & & Acc & LLM & Acc & LLM & Acc & LLM & Acc & LLM & Acc & LLM & Acc & LLM & \\
\midrule
\multirow{2}{*}{Naive}
 & Text-to-SQL & 0.421 & 0.421 & 0.225 & 0.229 & 0.121 & 0.129 & 0.129 & 0.175 & 0.062 & 0.094 & 0.163 & 0.182 & 2.00 \\
 & Dense RAG   & 0.050 & 0.200 & 0.004 & 0.108 & 0.010 & 0.121 & 0.017 & 0.067 & 0.010 & 0.096 & 0.016 & 0.115 & 2.00 \\
 % & Hybrid      & 0.338 & 0.458 & 0.188 & 0.271 & 0.113 & 0.192 & 0.142 & 0.179 & 0.069 & 0.123 & 0.147 & 0.220 & 4.00 \\
\midrule
\multirow{4}{*}{Agentic}
 & ReAct       & 0.196 & 0.258 & 0.058 & 0.096 & 0.033 & 0.079 & 0.008 & 0.021 & 0.006 & 0.029 & 0.049 & 0.085 & 4.20 \\
 & MAC-SQL     & 0.404 & 0.463 & 0.175 & 0.204 & 0.150 & 0.204 & 0.092 & 0.125 & 0.075 & 0.102 & 0.160 & 0.201 & 3.08 \\
 & TableRAG    & 0.163 & 0.400 & 0.054 & 0.188 & 0.035 & 0.175 & 0.037 & 0.087 & 0.015 & 0.133 & 0.051 & 0.185 & 6.55 \\
 & HetaRAG     & 0.379 & 0.504 & 0.188 & 0.336 & 0.108 & 0.263 & 0.133 & 0.263 & 0.062 & 0.212 & 0.149 & 0.293 & 25.21 \\
\midrule
Ours
 & RetailRAG   & 0.688 & 0.725 & 0.342 & 0.463 & 0.279 & 0.479 & 0.179 & 0.250 & 0.158 & 0.302 & 0.298 & 0.429 & 7.45 \\
\bottomrule
\end{tabular*}
\label{tab:baseline_comparison}
\vspace*{-0.05cm}
\end{table*}

\subsection{Comparison with Existing RAG Systems}
\label{subsec:baseline_comparison}

We compare \algname{} against six RAG baselines in two groups, selected to span retrieval strategies from one-shot SQL or dense retrieval to iterative agentic pipelines:
\begin{itemize}[leftmargin=*, topsep=1pt]
    \item \textbf{Non-agentic baselines}: Text-to-SQL \cite{wang2020ratsql} and Dense RAG \cite{lewis2020rag}, adapted to our multi-source setting with LLM-based source routing followed by one-shot SQL generation or top-$k$ retrieval.
    \item \textbf{Agentic baselines}: ReAct \cite{yao2023react}, MAC-SQL \cite{wang2025macsql}, TableRAG \cite{yu2025tablerag}, and HetaRAG \cite{yan2025hetarag}, with control flow and prompts preserved; HetaRAG's multi-store architecture is adapted to our setting. Implementation details are in Appendix \ref{appendix:baselines}.
\end{itemize}
All baselines share the same LLM backbone, embedding model, and top-10 retrieval as \algname{}. For the baseline comparison, \algname{} and all baselines use gpt-oss-120b as the backbone model and Qwen3-Embedding-8B for dense retrieval. 

We compare \algname{} against six existing RAG systems on \dataname{} (Section \ref{subsec:baseline_main}), isolate the contribution of each design through ablations (Section \ref{subsec:baseline_ablation}), and further inject the briefing into baselines to measure its standalone effect (Section \ref{subsec:briefing_injection}).

\subsubsection{Main Results}
\label{subsec:baseline_main} Table \ref{tab:baseline_comparison} compares \algname{} with the six baselines on \dataname{}, which we analyze in turn below.

\paragraph{Reasoning Capability.} 
%Table \ref{tab:baseline_comparison} shows that every system degrades as composition deepens, with the sharpest drops at the Level-1 to Level-2 transition where queries first require multi-hop bridging. Even at deeper levels, \algname{} remains the strongest yet retains only 0.302 at Level-3 Hetero., while the best baseline (HetaRAG) falls to 0.212. \textcolor{red}{These margins indicate that the reasoning axis of \dataname{} separates systems at every depth, and that compositional depth remains a demanding axis far from saturation.}
Existing RAG systems degrade as
composition deepens, with the sharpest drops at the Level-1 to Level-2
transition where queries first require multi-hop bridging. \algname{} remains
the strongest at every depth, reaching 0.302 at Level-3 Hetero.\ against 0.212
for the best baseline (HetaRAG). The reasoning axis therefore separates systems
at every depth while leaving substantial headroom at Level-3, establishing
compositional depth as a discriminative and unsaturated dimension of
\dataname{}.

% \paragraph{Multi-Source Chaining.}
% Within each level, most baselines drop sharply from Homo. to Hetero. sources (in Table \ref{tab:baseline_comparison}), indicating that multi-source chaining---not per-step retrieval---is the dominant bottleneck. \algname{} preserves the largest margin in Hetero settings (0.302 at Level-3 Hetero) at moderate cost (7.45 API calls), confirming that source-routed plan chaining narrows this gap.

\paragraph{Multi-Source Chaining.}
{Table \ref{tab:baseline_comparison} further shows that most baselines drop sharply from Homo. to Hetero. queries within each level, indicating that multi-source chaining rather than per-step retrieval is the dominant bottleneck. Even under this drop, \algname{} preserves the largest Hetero. margin over all baselines at a moderate cost of 7.45 API calls. These results confirm that the Homo./Hetero. axis of \dataname{} isolates cross-source bridging as a failure mode, and that source-routed plan chaining narrows this gap.}

% \paragraph{Evidence Modality.}
% Figure \ref{fig:query_modality} (a) compares frameworks across SQL-only, Text-only, and Composite (SQL+Text) queries at each reasoning level. \algname{} dominates baselines across all three modalities, with the largest margins on Text-only queries. Averaged across levels, most frameworks find SQL-only easiest and Text-only hardest, with Composite in between, exposing text retrieval as the dominant bottleneck of current agentic RAG.

\paragraph{Evidence Modality.}
Figure \ref{fig:query_modality} (a) compares frameworks across SQL-only, Text-only, and Composite (SQL+Text) queries at each reasoning level. Across levels, most frameworks score highest on SQL-only and lowest on Text-only queries, with Composite in between. \algname{} outperforms all baselines on every modality yet follows the same ordering, showing that the gap across modalities holds regardless of framework design. Accordingly, the modality axis of \dataname{} enables per-modality evaluation of how each framework handles structured and unstructured evidence.

% \subsubsection{Ablation Study of RetailRAG Modules}
% \label{subsec:baseline_ablation}
% Figure \ref{fig:retailrag_ablation} ablates \algname{}'s three designs, abbreviated as Schema Briefing, Plan Routing, and Plan Revision for readability (in the same order as Section \ref{sec:framework}). (i) Schema Briefing and Plan Revision
% % Plan Revision and Schema Briefing 
% dominate the average drop, both peaking at Level-2 where the transition from atomic to multi-hop reasoning is demanding. (ii) Plan Routing has its largest impact at Level-1 but nearly vanishes at Level-3, confirming its role as a dispatcher for simple queries rather than a multi-hop driver. (iii) Even with any single design removed, \algname{} still outperforms the strongest baseline on average, showing that the three designs contribute jointly rather than any single one being decisive.

\subsubsection{Ablation Study of RetailRAG}
\label{subsec:baseline_ablation}

\begin{wrapfigure}{r}{0.57\textwidth}
\centering
\definecolor{baselinered}{HTML}{CB2936}
\vspace*{-0.4cm}
\includegraphics[width=\linewidth]{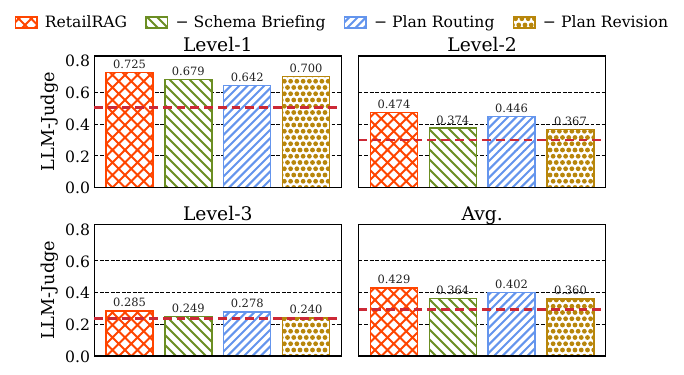}
\vspace*{-0.8cm}
\caption{Ablation of \algname{}'s three retail-specific designs on \dataname{} (LLM-judge). The \textcolor{baselinered}{\textbf{-\,-\,-}} line marks the best baseline (HetaRAG).}
\vspace*{-0.35cm}
\label{fig:retailrag_ablation}
\end{wrapfigure}

Figure \ref{fig:retailrag_ablation} ablates \algname{}'s three designs, abbreviated as Schema Briefing, Plan Routing, and Plan Revision for readability (in the same order as Section \ref{sec:framework}). (i) Schema Briefing and Plan Revision dominate the average drop, peaking at Level-2 where the transition from atomic to multi-hop reasoning is demanding. (ii) Ablating Plan Routing lowers accuracy across levels, most visibly at Level-1, indicating that the routing decision itself contributes to answer quality beyond the planning loop. (iii) Even with any single design removed, \algname{} outperforms the strongest baseline on average, showing that the three designs contribute jointly rather than any single one being decisive.

\subsubsection{Auto-Schema Briefing Injection Analysis}
\label{subsec:briefing_injection}

\begin{wraptable}{r}{0.5\textwidth}
\centering
\scriptsize
\vspace*{-0.4cm}
\caption{Effect of \algname{}'s auto-schema briefing on baselines. \emph{w/ Briefing}: the briefing is injected into each baseline. \emph{LLM}: LLM-as-judge accuracy. \emph{API Call}: API calls per query.}
\vspace*{-0.25cm}
\label{tab:briefing_injection}
\begin{tabular*}{\linewidth}{@{\extracolsep{\fill}}ll cc cc@{}}
\toprule
\multirow{2}{*}{Category} & \multirow{2}{*}{Method} & \multicolumn{2}{c}{Default} & \multicolumn{2}{c}{w/ Briefing} \\
\cmidrule(lr){3-4} \cmidrule(l){5-6}
 & & LLM & \makecell{API\\Call} & LLM & \makecell{API\\Call} \\
\midrule
Naive & Text-to-SQL & 0.182 & 2.00  & 0.296 & 2.00 \\
\midrule
\multirow{4}{*}{Agentic}
 & ReAct       & 0.085 & 4.20  & 0.102 & 4.29 \\
 & MAC-SQL     & 0.201 & 3.08  & 0.205 & 3.13 \\
 & TableRAG    & 0.185 & 6.55  & 0.229 & 6.23 \\
 & HetaRAG     & 0.293 & 25.21 & 0.336 & 23.37 \\
\midrule
Ours & \algname{} & \multicolumn{2}{c}{--} & 0.429 & 7.45 \\
\bottomrule
\end{tabular*}
\vspace*{-0.3cm}
\end{wraptable}

To validate the effectiveness of our auto-schema briefing itself and to isolate its contribution from \algname{}'s planning designs, we inject the identical briefing into all five baselines that accept schema input. Table \ref{tab:briefing_injection} shows that the briefing improves every baseline, with the largest gain on Text-to-SQL, which depends most on schema understanding. Nevertheless, even the strongest baseline remains well below \algname{} while using three times more API calls, confirming that the briefing alone cannot close the gap and that the gains stem from \algname{}'s plan chaining and revision rather than schema access.

\begin{table*}[!t]
\centering
\scriptsize
\caption{LLM benchmark on \dataname{} by reasoning level and source diversity. \emph{Homo.}/\emph{Hetero.}: evidence from one or multiple sources. \emph{Acc}: accuracy. \emph{LLM}: LLM-as-judge accuracy. \emph{API Call}: mean LLM calls per query.}
\vspace*{-0.25cm}
\setlength{\tabcolsep}{3.2pt}
\renewcommand{\arraystretch}{1.05}
\begin{tabular*}{\textwidth}{l l cc cc cc cc cc cc c}
\toprule
\multirow{3}{*}{Category} & \multirow{3}{*}{Model}
 & \multicolumn{2}{c}{\multirow{2}{*}{Level-1}}
 & \multicolumn{4}{c}{Level-2}
 & \multicolumn{4}{c}{Level-3}
 & \multicolumn{2}{c}{\multirow{2}{*}{Avg.}}
& \multirow{3}{*}{\makecell{API\\Call}} \\
\cmidrule(lr){5-8} \cmidrule(lr){9-12}
 & & &
 & \multicolumn{2}{c}{Homo.} & \multicolumn{2}{c}{Hetero.}
 & \multicolumn{2}{c}{Homo.} & \multicolumn{2}{c}{Hetero.}
 & & & \\
\cmidrule(lr){3-4} \cmidrule(lr){5-6} \cmidrule(lr){7-8} \cmidrule(lr){9-10} \cmidrule(lr){11-12} \cmidrule(lr){13-14}
 & & Acc & LLM & Acc & LLM & Acc & LLM & Acc & LLM & Acc & LLM & Acc & LLM & \\
\midrule
\multirow{6}{*}{Open-source}
& Qwen3.6-35B-A3B   & 0.562 & 0.608 & 0.154 & 0.242 & 0.135 & 0.217 & 0.092 & 0.121 & 0.052 & 0.125 & 0.169 & 0.236 & 12.76 \\
& Kimi-K2.5         & 0.621 & 0.658 & 0.179 & 0.254 & 0.129 & 0.204 & 0.113 & 0.158 & 0.090 & 0.154 & 0.193 & 0.255 & 11.81 \\
& Gemma-4-31B-IT    & 0.692 & 0.704 & 0.346 & 0.492 & 0.317 & 0.529 & 0.242 & 0.354 & 0.208 & 0.440 & 0.333 & 0.498 & 10.50 \\
& GPT-oss-120B      & 0.688 & 0.725 & 0.342 & 0.463 & 0.279 & 0.479 & 0.179 & 0.250 & 0.158 & 0.302 & 0.298 & 0.429 &  7.45 \\
& DeepSeek-V4-Flash & 0.671 & 0.738 & 0.308 & 0.433 & 0.269 & 0.450 & 0.242 & 0.338 & 0.177 & 0.369 & 0.302 & 0.449 &  8.88 \\
& DeepSeek-V4-Pro   & 0.713 & 0.762 & 0.329 & 0.475 & 0.306 & 0.490 & 0.212 & 0.321 & 0.183 & 0.362 & 0.319 & 0.466 &  9.68 \\
\midrule
\multirow{4}{*}{Proprietary}
& Gemini-3.1-Flash  & 0.662 & 0.671 & 0.254 & 0.362 & 0.190 & 0.338 & 0.167 & 0.254 & 0.129 & 0.271 & 0.246 & 0.358 &  9.72 \\
& Qwen3.6-Flash     & 0.662 & 0.667 & 0.354 & 0.438 & 0.277 & 0.406 & 0.212 & 0.275 & 0.167 & 0.271 & 0.302 & 0.390 &  7.32 \\
& Qwen3.6-Plus      & 0.700 & 0.713 & 0.379 & 0.533 & 0.312 & 0.506 & 0.250 & 0.375 & 0.244 & 0.471 & 0.349 & 0.511 &  6.92 \\
& GPT-5.4  & 0.700 & 0.742 & 0.383 & 0.546 & 0.335 & 0.542 & 0.254 & 0.367 & 0.202 & 0.408 & 0.345 & 0.508 &  7.28 \\
\bottomrule
\end{tabular*}
\label{tab:llm_benchmark}
\vspace*{-0.1cm}
\end{table*}

\subsection{Benchmarking LLMs on RetailBridge}
\label{subsec:llm_benchmark}

Holding \algname{} fixed as the common backbone, we now shift from frameworks to models and ask how far current LLMs can carry retail workflow queries. Here, we swap ten backbones into \algname{}---open-source and proprietary, small to large in scale---and analyze their accuracy (Section \ref{subsec:llm_main}), source routing performance (Section \ref{subsec:llm_routing}), and efficiency-accuracy trade-off (Section \ref{subsec:llm_efficiency}). 
LM configurations are provided in Appendix \ref{appendix:llm_config}.

\subsubsection{Main Results}
\label{subsec:llm_main} Table \ref{tab:llm_benchmark} reports the performance of ten backbones under \algname{} across
the three axes of \dataname{}, which we analyze in turn below.

\paragraph{Reasoning Capability.}
Every backbone loses about half or more of its Level-1 performance by Level-3. Even the strongest models (Qwen3.6-Plus, GPT-5.4) reach average LLM-judge scores of only 0.511 and 0.508, leaving \dataname{} unsolved by current frontier models. Notably, a small open-source model (Gemma-4-31B-IT, 0.498) nearly matches the proprietary leaders, suggesting that reasoning performance depends more on how a model decomposes and plans than on its size or training regime. As a result, the leveled design of \dataname{} surfaces these capability differences that aggregate or single-hop evaluation would miss.

% \paragraph{Reasoning Capability.}
% Table \ref{tab:llm_benchmark} shows the strongest models (Qwen3.6-Plus, GPT-5.4) reaching average LLM-judge scores of 0.511 and 0.508,
% % LLM Avg., 
% with every backbone losing roughly half of its Level-1 performance by Level-3. Model category and scale offer little signal. A small open-source model (Gemma-4-31B-IT, 0.498) nearly matches the proprietary leaders, suggesting that reasoning performance depends more on how a model decomposes and plans than on its size or training regime.

\paragraph{Multi-Source Chaining.}
Table \ref{tab:llm_benchmark} further shows that at each level, most backbones score lower on Hetero.\ than on Homo.\ queries, confirming that bridging evidence across different sources adds difficulty beyond chaining within a single source at the same depth. Interestingly, this gap is not uniform, as large models narrow or even reverse it at Level-3 regardless of model category (Qwen3.6-Plus, GPT-5.4, Gemma-4-31B-IT, DeepSeek-V4-Pro), whereas smaller models remain below 0.30 on Hetero.\ queries. Cross-source bridging thus emerges as a capability that grows with scale rather than model category. The Homo./Hetero.\ split of \dataname{} makes this capability separately measurable from reasoning depth.

% \paragraph{Multi-Source Chaining.}
% Table \ref{tab:llm_benchmark} shows that Hetero.\ queries amplify capability differences along scale but not category. Large models---proprietary (Qwen3.6-Plus, GPT-5.4) and open-source (Gemma-4-31B-IT, DeepSeek-V4-Pro)---outperform their Homo.\ scores at Level-3, with Qwen3.6-Plus gaining from 0.375 to 0.471. Smaller models show no such advantage and remain below 0.30. Multi-source chaining therefore tracks scale rather than provider, indicating that scale enables effective use of routed multi-source evidence.

\paragraph{Evidence Modality.}
Figure \ref{fig:query_modality} (b) shows the LLM benchmark across SQL-only, Text-only, and Composite queries, separately for open-source and proprietary backbones. Across both groups, backbones degrade less on SQL-only than on Text-only and Composite queries at deeper levels, confirming that the difficulty of a query depends on which evidence modality it requires, not only on its depth. In particular, Text-only queries collapse the steepest, identifying dense text retrieval as the dominant modality bottleneck regardless of model category or scale. The modality split of \dataname{} thus locates each backbone's failures in a specific evidence type, which level-wise scores alone would leave hidden.

% \paragraph{Evidence Modality.}
% Figure \ref{fig:query_modality} (b) shows the LLM benchmark across SQL-only, Text-only, and Composite queries, separately for open-source and proprietary backbones. Both groups handle Level-1 queries comfortably and degrade less on SQL-only queries than on Text-only and Composite queries at Level-2 and Level-3, with Text-only collapsing the steepest. This pattern identifies dense text retrieval over retail records---rather than SQL execution---as the dominant modality bottleneck across LLM backbones, regardless of provider or scale.

\begin{wraptable}{r}{0.55\textwidth}
\centering
\scriptsize
\vspace*{-0.4cm}
\caption{Source routing F1 on \dataname{} (1{,}680 queries) by reasoning level and source diversity. Avg.\ is weighted by the number of queries per cell (L1/L2-Homo/L3-Homo: 240 each; L2-Hetero/L3-Hetero: 480 each).}
\vspace*{-0.25cm}
\setlength{\tabcolsep}{2.5pt}
\begin{tabular*}{\linewidth}{@{\extracolsep{\fill}}l c cc cc c@{}}
\toprule
\multirow{3}{*}{Model}
 & \multirow{3}{*}{Level-1}
 & \multicolumn{2}{c}{Level-2}
 & \multicolumn{2}{c}{Level-3}
 & \multirow{3}{*}{Avg.} \\
\cmidrule(lr){3-4} \cmidrule(lr){5-6}
 & & Homo. & Hetero. & Homo. & Hetero. & \\
\midrule
Qwen3.6-35B-A3B   & 0.606 & 0.359 & 0.326 & 0.241 & 0.273 & 0.343 \\
Kimi-K2.5         & 0.616 & 0.352 & 0.273 & 0.250 & 0.299 & 0.337 \\
Gemma-4-31B-IT    & 0.684 & 0.606 & 0.590 & 0.545 & 0.607 & 0.604 \\
GPT-oss-120B      & 0.694 & 0.588 & 0.533 & 0.474 & 0.484 & 0.541 \\
DeepSeek-V4-Flash & 0.703 & 0.563 & 0.551 & 0.509 & 0.529 & 0.562 \\
DeepSeek-V4-Pro   & 0.722 & 0.560 & 0.515 & 0.435 & 0.485 & 0.531 \\
\midrule
Gemini-3.1-Flash  & 0.685 & 0.604 & 0.489 & 0.495 & 0.438 & 0.520 \\
Qwen3.6-Flash     & 0.683 & 0.684 & 0.557 & 0.640 & 0.519 & 0.594 \\
Qwen3.6-Plus      & 0.692 & 0.708 & 0.631 & 0.647 & 0.648 & 0.658 \\
GPT-5.4           & 0.757 & 0.674 & 0.612 & 0.608 & 0.607 & 0.640 \\
\bottomrule
\end{tabular*}
\label{tab:source_routing}
\vspace*{-0.3cm}
\end{wraptable}

\subsubsection{Source Routing Analysis}
\label{subsec:llm_routing}
Table \ref{tab:source_routing} reports source routing F1, measuring how accurately each model identifies the right sources by comparing its queried sources against the gold source set per query. Cross-referencing with answer accuracy (Table \ref{tab:llm_benchmark}) separates two failure stages, routing to the wrong sources and failing to integrate correctly routed evidence. Routing degrades less sharply than answer accuracy across all backbones, indicating that queries can fail after routing succeeds. Moreover, strong models convert about 70\% of their routing F1 into correct answers while weaker models (\emph{e.g.}, Qwen3.6-35B-A3B) convert under 50\%. Effective multi-source reasoning thus requires both correct routing and the capacity to integrate routed evidence, a decomposition that the gold source annotations of \dataname{} make possible.

% \subsubsection{Source Routing Analysis}
% \label{subsec:llm_routing}
% Table \ref{tab:source_routing} reports source routing F1,  measuring how accurately each model identifies the right sources by comparing its queried sources against the gold source set per query. Cross-referencing with answer accuracy (Table \ref{tab:llm_benchmark}) reveals a two-stage failure pattern.
% % comparing each model's queried sources against the gold source set per query. Cross-referencing with answer accuracy (Table \ref{tab:llm_benchmark}) reveals a two-stage failure pattern.
% Routing degrades less sharply than answer accuracy across all backbones, yet the conversion from routing to answers is itself capacity-dependent---strong models convert about 70\% of their routing F1 into correct answers, while weaker model (\emph{e.g.}, Qwen3.6-35B-A3B) convert under 50\%. Effective multi-source reasoning thus requires both correct routing and the capacity to integrate routed evidence, where larger models show an advantage.
% % Effective multi-source reasoning thus requires both correct routing and the capacity to integrate routed evidence.

\subsubsection{Efficiency-Accuracy Trade-off}
\label{subsec:llm_efficiency}

\begin{wrapfigure}{r}{0.6\textwidth}
\centering
\vspace*{-0.4cm}
\includegraphics[width=\linewidth]{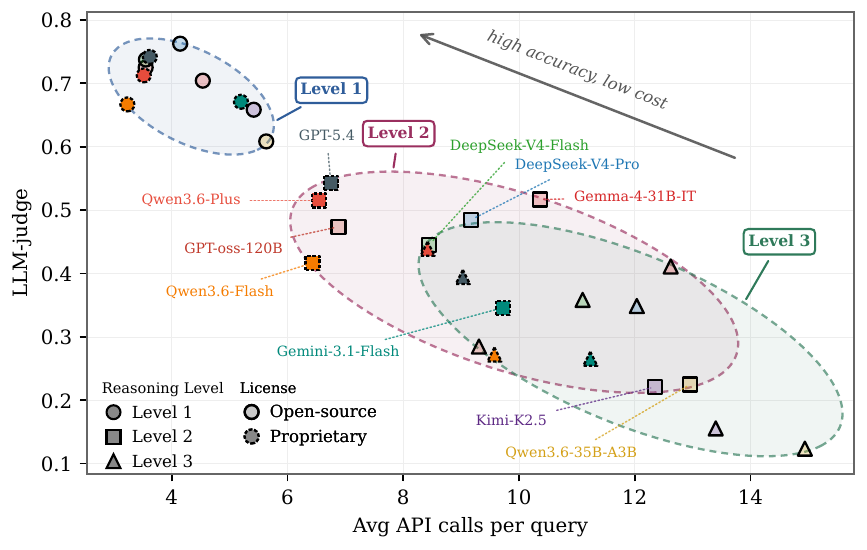}
\vspace*{-0.8cm}
\caption{Efficiency-accuracy trade-off of LLM backbones on \dataname{}. Arrow indicates the ideal.}
\vspace*{-0.3cm}
\label{fig:efficiency}
\end{wrapfigure}

Figure \ref{fig:efficiency} plots each model's trade-off between API calls and answer quality. Across the cost-quality plane, the three reasoning levels form distinct clusters, with Level-1 in the top-left and Level-2/3 in the high-cost region. Within each cluster, additional API calls do not translate into accuracy, as Qwen3.6-35B-A3B and Kimi-K2.5 consume the most calls yet rank lowest, while top backbones reach the ideal corner with substantially fewer. Top open-source models (GPT-oss-120B, DeepSeek-V4-Pro) match the accuracy of the strongest proprietary backbones at a moderately higher cost, indicating that the accuracy gap between providers has largely closed for retail multi-source chaining. Consequently, fewer API calls reflect accurate schema understanding and correct planning, which complete the chain without repeated revisions of failed steps and thereby lead to higher accuracy.

% \subsubsection{Efficiency-Accuracy Trade-off}
% \label{subsec:llm_efficiency}
% Figure~\ref{fig:efficiency} plots each model's trade-off between API calls and answer quality. The three reasoning levels form distinct clusters along the cost-quality plane, with Level-1 in the top-left and Level-2/3 spreading into the high-cost region with partial overlap. More API calls do not help---Qwen3.6-35B-A3B and Kimi-K2.5 consume the most calls yet rank lowest, while top backbones reach the ideal corner with substantially fewer. Notably, top open-source models (GPT-oss-120B, DeepSeek-V4-Pro) sit alongside the strongest proprietary backbones in the top-left, indicating that the gap between providers has largely closed for retail multi-source chaining.

\section{Conclusion}
\label{sec:conclusion}
We introduced \dataname{}, a compositional multi-source retail benchmark that not only measures aggregate accuracy but localizes failures along reasoning depth, source diversity, and evidence modality, together with \algname{}, the first agentic RAG framework natively designed for enterprise retail. \algname{} consistently outperforms baselines, and our LLM benchmark shows that compositional depth and cross-source bridging remain unsolved even for frontier models, with text retrieval as the dominant modality bottleneck. Together, they provide a diagnostic testbed and a reference framework for building LLM agents that operate reliably over real enterprise retail data.

\bibliographystyle{assets/plainnat}
\bibliography{colm2026_conference}

\appendix
% ---------------- Appendix table of contents ----------------
\clearpage
\startcontents[appendix]
\section*{Appendix Contents}
{\hypersetup{linkcolor=black}
\printcontents[appendix]{l}{1}{\setcounter{tocdepth}{2}}}
\clearpage

\section{Ethics and Fairness}
We identify no significant ethical risks in this work, and address the four considerations below.

\paragraph{Data Privacy.}
\dataname{} contains no personally identifiable information or real customer records. All interactions are produced by persona-driven simulation with persona names unlinked to any real individual, and product metadata is drawn from a publicly available dataset \cite{hou2026bridging}. This design eliminates the privacy risks inherent in real-world enterprise data, such as customer re-identification.

\paragraph{Consent.}
Human verification was conducted by graduate-level NLP experts and a retail domain expert, who participated voluntarily, provided informed consent, and received clear annotation instructions; the annotators were compensated above the local statutory minimum wage. Since the task collected no data about the annotators themselves and involved only quality judgments over synthetic instances, it does not constitute human subjects research and did not require IRB review.

\paragraph{Bias.}
\dataname{} is grounded in real-world signals at multiple points. Its product catalog is curated from real Amazon data \cite{hou2026bridging}, its simulator has been validated to reproduce real-world economic regularities \cite{choi2026retailsim}, and its queries are audited by a retail domain expert with 90\% approval (Section \ref{subsec:quality_assurance}) and shown to read on par with human-authored retail queries (Appendix \ref{appendix:human_comparison}). To further reduce model-specific bias in generation, we diversify agent and generator backbones across four LLM families. Under this design, no biases surfaced in our expert verification stages.

\paragraph{Potential Misuse.}
\dataname{} is released solely for evaluating agentic RAG systems, and as it contains no real customer records, it poses no re-identification risk. Our experiments use publicly available LLMs and APIs in accordance with their licenses. See Appendix \ref{appendix:human_annotation} for human annotation details.

\section{Benchmark Construction}
\label{appendix:benchmark_construction}

\subsection{Simulation Setup for RetailBridge}
\label{appendix:simulation_setup}
We produce persona-driven seller--buyer trajectories using RetailSim \cite{choi2026retailsim}, extended with additional scenario-level controls and instantiated at our own scale. These trajectories serve as the raw material from which $\mathcal{D}$ is constructed.

\paragraph{Agents and Role Assignment.}
Each simulation run instantiates $16$ sellers and $96$ buyers from pre-built persona configurations. Each agent is defined by a name, gender, role-specific traits, and an assigned LLM backbone. Seller traits include assertiveness, friendliness, and rationality, while buyer traits include pickiness, price consciousness, and rationality, conditioning the agent's tone and behavior throughout the simulated journey. LLM backbones (DeepSeek-V3.2, Gemini-3-Flash, GPT-oss-120B, Qwen3-235B) are assigned to seller and buyer agents in a balanced manner for each run.

\paragraph{Products.}
The product catalog is curated from Amazon Reviews 2023 \cite{hou2026bridging}, selecting products with valid metadata and normalizing them into four retail categories (Food, Home, Fashion, Electronics) with $50$ products per category. To instantiate a realistic retail environment, each product is further augmented with a flat shipping fee of $\min(0.05 \times \text{price}, \$8.00)$.

\paragraph{Retail Scenario Controls.}
To reflect realistic retail settings, the simulation is governed by scenario-level factors. Each product is promoted by one or two sellers, enabling both exclusive sales pitches and competing offers. Sales pitches are scheduled within bounded promotion windows with assigned discount rates, and each buyer is exposed to only a random subset. These controls introduce variation in seller competition, promotion timing, buyer exposure, and discount-driven purchase incentives.

\begin{figure}[t]
\centering
\includegraphics[width=\textwidth]{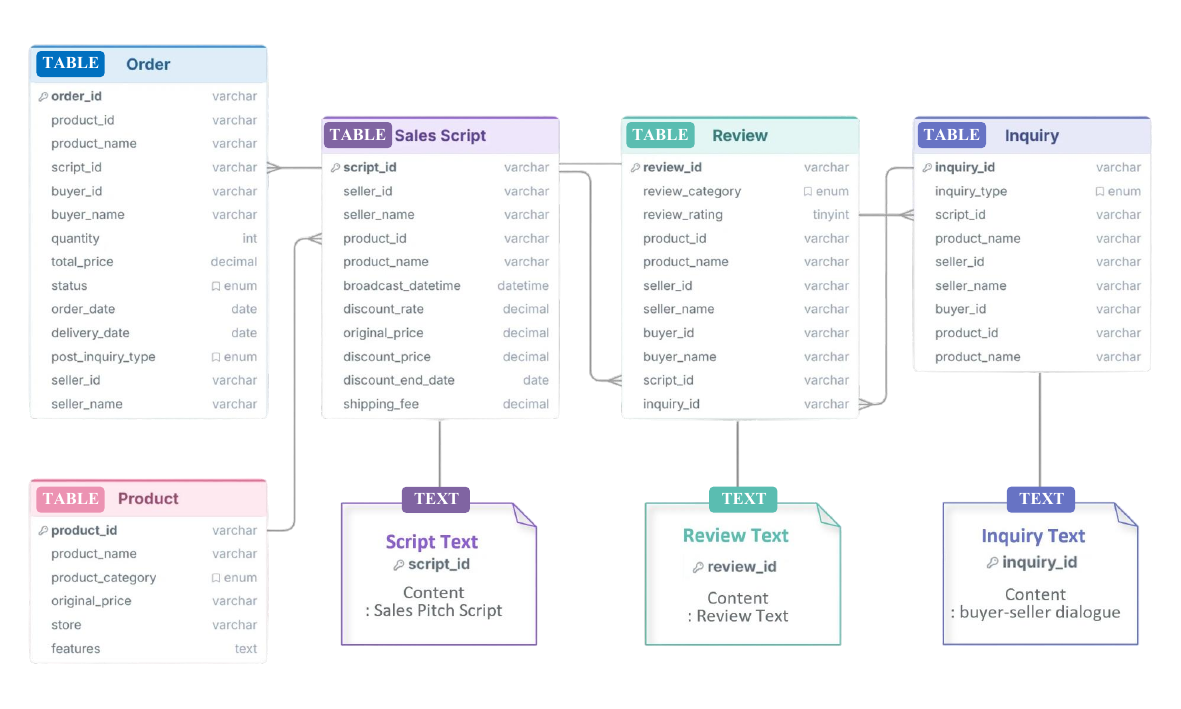}
\vspace{-0.8cm}
\caption{Schema of the heterogeneous retail database $\mathcal{D}$ used in \dataname{}.}
\vspace{-0.2cm}
\label{fig:dbschema}
\end{figure}

\begin{table}[t]
\centering
\small
\caption{Examples of answer-linked composition with $O_\mathrm{ans}$ in \dataname{}, covering entity-set and scalar answers. Colors mark the bridged value (\hlans{$O_\mathrm{ans}$}) and the referring expression \hlcomp{$q^{\text{ref}}$} hiding it in $q_{\ell+1}$.}
\setlength{\tabcolsep}{2.5pt}
\begin{tabularx}{\columnwidth}{@{}c l X@{}}
\toprule
\textbf{Operator} & \textbf{Query} & \textbf{Example} \\
\midrule
% ---------- Set (plural) ----------
\multirow{10}{*}{\makecell{$O_\mathrm{ans}$\\\tiny(Entity Set)}}
& \makecell[tl]{$q_\ell$\\\tiny(Level-$\ell$)}
& Which products sold by Leslie B.\ have the highest refund rate?\quad
  \textit{A:} \{\hlans{Dell 2019 New ..., Herstyle Certain...}\} (2 products) \\
& \makecell[tl]{$\tilde{q}_\ell$\\\tiny(Follow-up)}
& What is the average product review rating for \hlans{these products}?\quad
  \textit{A:} 3.37 \\
& \makecell[tl]{$q_{\ell+1}$\\\tiny(Composite)}
& What is the average product review rating for \hlcomp{the products sold by
  Leslie B. that have the highest refund rate}?\quad\textit{A:} 3.37 \\
& \textit{Effect}
& Set bridge (\textsc{Order} $\rightarrow$ \textsc{Review}); a plural $q^{\text{ref}}$ hides the whole answer set \\
\midrule
\multirow{9}{*}{\makecell{$O_\mathrm{ans}$\\\tiny(Scalar)}}
& \makecell[tl]{$q_\ell$\\\tiny(Level-$\ell$)}
& What is the total amount spent by Lisa J.\ across all her orders?\quad
  \textit{A:} \hlans{\$3{,}161.72} \\
& \makecell[tl]{$\tilde{q}_\ell$\\\tiny(Follow-up)}
& Which buyers have a total order value greater than \hlans{\$3{,}161.72}?\quad
  \textit{A:} \{Adam N., Brent B., \ldots\} (18 buyers) \\
& \makecell[tl]{$q_{\ell+1}$\\\tiny(Composite)}
& Which buyers have a total order value greater than \hlcomp{the total amount
  spent by Lisa J. across all her orders}?\quad
  \textit{A:} \{Adam N., Brent B., \ldots\} (18 buyers) \\
& \textit{Effect}
& Scalar bridge (\textsc{Order} $\rightarrow$ \textsc{Order}); a singular $q^{\text{ref}}$ hides the threshold value \\
\bottomrule
\end{tabularx}
\label{tab:answer_linked_types}
\end{table}

\subsection{Schema of the Heterogeneous Retail Database}
\label{appendix:schema_database}
We map the simulated trajectories onto a heterogeneous database $\mathcal{D}$ of eight sources, spanning five tabular sources (Order, Sales Script, Review, Inquiry, Product) and three textual sources (Script Text, Review Text, Inquiry Text), whose underlying schema is shown in Figure \ref{fig:dbschema}. The tabular sources store structured records of retail activities and are linked through shared identifiers such as \texttt{product\_id}, \texttt{seller\_id}, and \texttt{buyer\_id}, while each textual source pairs with a tabular source via its primary key (\texttt{script\_id}, \texttt{review\_id}, \texttt{inquiry\-\_id}), holding free-text content. This structure captures the heterogeneity of real enterprise retail data, where structured transaction logs coexist with unstructured customer interactions. The full database format and schema files are available at \url{https://github.com/DISL-Lab/RetailBridge}.

% The simulated sales pitch, orders, reviews, and inquiries become five relational tables---product, script, order, review, and inquiry---holding their numeric and categorical attributes, while the sales scripts, review bodies, and inquiry dialogues form three aligned text corpora. 

\subsection{Answer-Linked Composition by Answer Type}
\label{appendix:answertype_qa}
While Table \ref{tab:composition_operators} illustrates answer-linked composition (Section \ref{subsec:multi_source_qa}) with a set-valued intermediate answer, $O_\mathrm{ans}$ also handles scalar answers arising from aggregation or ranking, which likewise admit no noun-phrase referent. Table \ref{tab:answer_linked_types} presents both types side by side. In each case, the literal answer $a_\ell$ in the follow-up question $\tilde{q}_\ell$ is hidden behind the referring expression $q^{\text{ref}}_{\ell}$ derived from $q_\ell$, and only the surface form of $q^{\text{ref}}_{\ell}$ differs, a plural noun phrase for entity sets and a singular one for scalar thresholds.

\begin{table}[t]
\centering
\small
\caption{Statistics of \dataname{}: source-level counts and QA distribution by reasoning capability, multi-source chaining, and evidence modality. \emph{Homo./Hetero.} denote single- or multi-source evidence; \emph{Composite}, SQL+Text questions.}
\vspace*{-0.2cm}
\begin{tabularx}{\linewidth}{@{}XXX@{}}
\toprule
\textbf{Statistic} & & \multicolumn{1}{r@{}}{\textbf{Number}} \\
\midrule
\textbf{Sources} & & \multicolumn{1}{r@{}}{5 tabular / 3 textual} \\
\quad Total rows: 49{,}718 & Total docs: 43{,}140 & \\
\cmidrule{1-3}
\multicolumn{3}{@{}l@{}}{\quad \textit{Tabular} (rows / cols)} \\
\qquad Product: 200 / 6 & Order: 6{,}378 / 14 & Script: 287 / 11 \\
\qquad Review: 24{,}104 / 11 & Inquiry: 18{,}749 / 9 & \\
\cmidrule{1-3}
\multicolumn{3}{@{}l@{}}{\quad \textit{Textual} (docs / avg.\ chars)} \\
\qquad Scripts: 287 / 2{,}286 & Reviews: 24{,}104 / 294 & Inquiries: 18{,}749 / 5{,}730 \\
\midrule
\textbf{Total Questions} & & \multicolumn{1}{r@{}}{1{,}680} \\
\multicolumn{3}{@{}l@{}}{\quad \textit{Generators}} \\
\multicolumn{3}{@{}l@{}}{\qquad Qwen3.6-Max, DeepSeek-V4-Pro, Gemini-3.1-Pro} \\
\multicolumn{3}{@{}l@{}}{\quad \textit{Validators}} \\
\multicolumn{3}{@{}l@{}}{\qquad GPT-5.5, Claude Opus 4.7} \\
\cmidrule{1-3}
\multicolumn{3}{@{}l@{}}{\quad \textit{answer type}} \\
\qquad Entity: 329\,(19.58\%) & Set: 211\,(12.56\%) & Number: 465\,(27.68\%) \\
\qquad Temporal: 84\,(5.00\%) & Text: 591\,(35.18\%) & \\
\cmidrule{1-3}
\multicolumn{3}{@{}l@{}}{\quad \textit{reasoning capability}} \\
\qquad Level-1: 240\,(14.3\%) & Level-2: 720\,(42.9\%) & Level-3: 720\,(42.9\%) \\
\cmidrule{1-3}
\multicolumn{3}{@{}l@{}}{\quad \textit{multi-source chaining}} \\
\qquad Homo: 720\,(42.9\%) & Hetero: 960\,(57.1\%) & \\
\cmidrule{1-3}
\multicolumn{3}{@{}l@{}}{\quad \textit{evidence modality}} \\
\qquad SQL-only: 600\,(35.7\%) & Text-only: 600\,(35.7\%) & Composite: 480\,(28.6\%) \\
\bottomrule
\end{tabularx}
\label{tab:bridgeqa_statistics}
\end{table}

\section{Benchmark Statistics and Validity}
\label{appendix:benchmark_validity}

\subsection{Data Statistics of RetailBridge}
\label{appendix:data_statistics}
Table \ref{tab:bridgeqa_statistics} reports the detailed statistics of $\mathcal{D}$ and the QA distribution.

\paragraph{Source-Level Statistics.}
$\mathcal{D}$ consists of five tabular sources (Product, Order, Script, Review, Inquiry) and three textual sources (Scripts, Reviews, Inquiries). Tabular sources span a product catalog and large interaction logs, while textual sources range from short reviews to long inquiries, reflecting the heterogeneity of real enterprise retail records.

\paragraph{QA Generation Pipeline.}
QA generation is split evenly across three strong LLMs (Qwen3.6-Max, DeepSeek-V4-Pro, Gemini-3.1-Pro) to reduce model-specific bias, and each generated QA is independently verified by two LLM judges (GPT-5.5, Claude Opus 4.7) before being passed to human annotation (Appendix \ref{appendix:human_annotation}).

\begin{table*}[t]
\centering
\small
\caption{Representative \dataname{} examples across reasoning capability, multi-source chaining, and evidence modality. Source gives the source diversity (Homo./Hetero.) and the underlying sources an agent must route to; SQL tables are marked with $^*$ and text stores with $^\dagger$. Long product names in questions are abbreviated; answers are given in full.}
\setlength{\tabcolsep}{3pt}
\renewcommand{\arraystretch}{1.15}
\begin{tabularx}{\textwidth}{@{}c c c >{\raggedright\arraybackslash}X@{}}
\toprule
\textbf{Level} & \textbf{Source} & \textbf{Modality} & \textbf{Example question / answer} \\
\midrule
\multirow{4}{*}{Level-1}
 & \makecell[c]{Homo.\\(Order$^*$)} & SQL & \textbf{Q:} What is the total revenue for Michael K.? \newline \textbf{A:} \$11{,}296.83 \\
 & \makecell[c]{Homo.\\(Script$^\dagger$)} & Text & \textbf{Q:} What input connections did Mark A. mention for the INSIGNIA 2.1-Channel 80W Soundbar \ldots? \newline \textbf{A:} RCA, optical digital, mini-jack, and Bluetooth. \\
\midrule
\multirow{15}{*}{Level-2}
 & \makecell[c]{Homo.\\(Order$^*$)} & SQL & \textbf{Q:} Which sellers have a refund rate greater than Mark A.'s refund rate? \newline \textbf{A:} Natalie H., Joshua H., Joann I., Robert A., Olivia F., Michael K., Diana G., Daniel X., Ashley X., Steven C., Leslie B., Allen G., Alexander G., Michelle W. \\
 & \makecell[c]{Hetero.\\(Order$^*$,\\Product$^*$)} & SQL & \textbf{Q:} What is the original price of the product sold by Robert A. that has the highest conversion rate? \newline \textbf{A:} \$21.31 \\
 & \makecell[c]{Homo.\\(Inquiry$^\dagger$)} & Text & \textbf{Q:} What did Ashley X. tell Gabriel Z. about the return shipping costs for a change-of-mind return of the product that was the subject of a pre-purchase inquiry between Beth A. and Steven C.? \newline \textbf{A:} The buyer must cover the return shipping cost; no prepaid label is provided for a change-of-mind return. \\
 & \makecell[c]{Hetero.\\(Inquiry$^\dagger$,\\Script$^\dagger$)} & Text & \textbf{Q:} What structural features did the seller highlight to assure buyers that the product the buyer asked about (regarding weight limits and dimensions) can support heavy winter coats? \newline \textbf{A:} The rack features a solid wood core and sturdy metal anchors, providing a rock-solid structure designed to securely hold heavy winter coats. \\
 & \makecell[c]{Hetero.\\(Order$^*$,\\Inquiry$^\dagger$)} & Composite & \textbf{Q:} What is the flat-rate shipping cost for Crystal T.'s highest net revenue product? \newline \textbf{A:} \$7.13 per order. \\
\midrule
\multirow{10}{*}{Level-3}
 & \makecell[c]{Homo.\\(Order$^*$)} & SQL & \textbf{Q:} What is the total net revenue (excluding refunds) generated by the sellers with conversion rates greater than Daniel X.'s? \newline \textbf{A:} \$41{,}852.72 \\
 & \makecell[c]{Hetero.\\(Inquiry$^*$,\\Product$^*$,\\Order$^*$)} & SQL & \textbf{Q:} How many orders fall into each status category for the product sold by Michelle W. that has received post-purchase inquiries and has the highest original price? \newline \textbf{A:} delivered: 11 \textbar{} exchanged: 7 \textbar{} refunded: 1 \\
 & \makecell[c]{Hetero.\\(Review$^*$,\\Product$^*$,\\Inquiry$^\dagger$)} & Composite & \textbf{Q:} What did Olivia F. explain about the effect of substituting butter for coconut oil on the carb count of the product with the highest original price among those Beth A. rated the highest? \newline \textbf{A:} Olivia F. explained that switching from coconut oil to butter would not affect the carbohydrate content, as both are non-carbohydrate fats. \\
\bottomrule
\end{tabularx}
\label{tab:retailbridge_qa_examples}
\end{table*}

\paragraph{QA Distribution.}
The final benchmark of $1{,}680$ queries is stratified along three diagnostic axes: reasoning capability (Level-1/2/3), multi-source chaining (Homo./Hetero.), and evidence modality (SQL-only, Text-only, Composite). Answer types span entity, set, number, temporal, and free text, reflecting the variety of retail information needs.

\subsection{QA Examples from RetailBridge}
\label{appendix:qa_example}
Table \ref{tab:retailbridge_qa_examples} presents representative examples from \dataname{}, selected to cover the combinations of the three diagnostic axes, reasoning level, source diversity, and evidence modality. Each row lists the sources an agent must route to, so that the difficulty progression is visible from single-source lookups at Level-1 to cross-modality chains at Level-3. All questions and answers are shown as they appear in the benchmark.

\subsection{Human Annotation}
\label{appendix:human_annotation}
This section details the setup, criteria, and protocol of the human verification introduced in Section \ref{subsec:quality_assurance}.

\begin{figure*}[t]
    \centering
    \includegraphics[width=\textwidth]{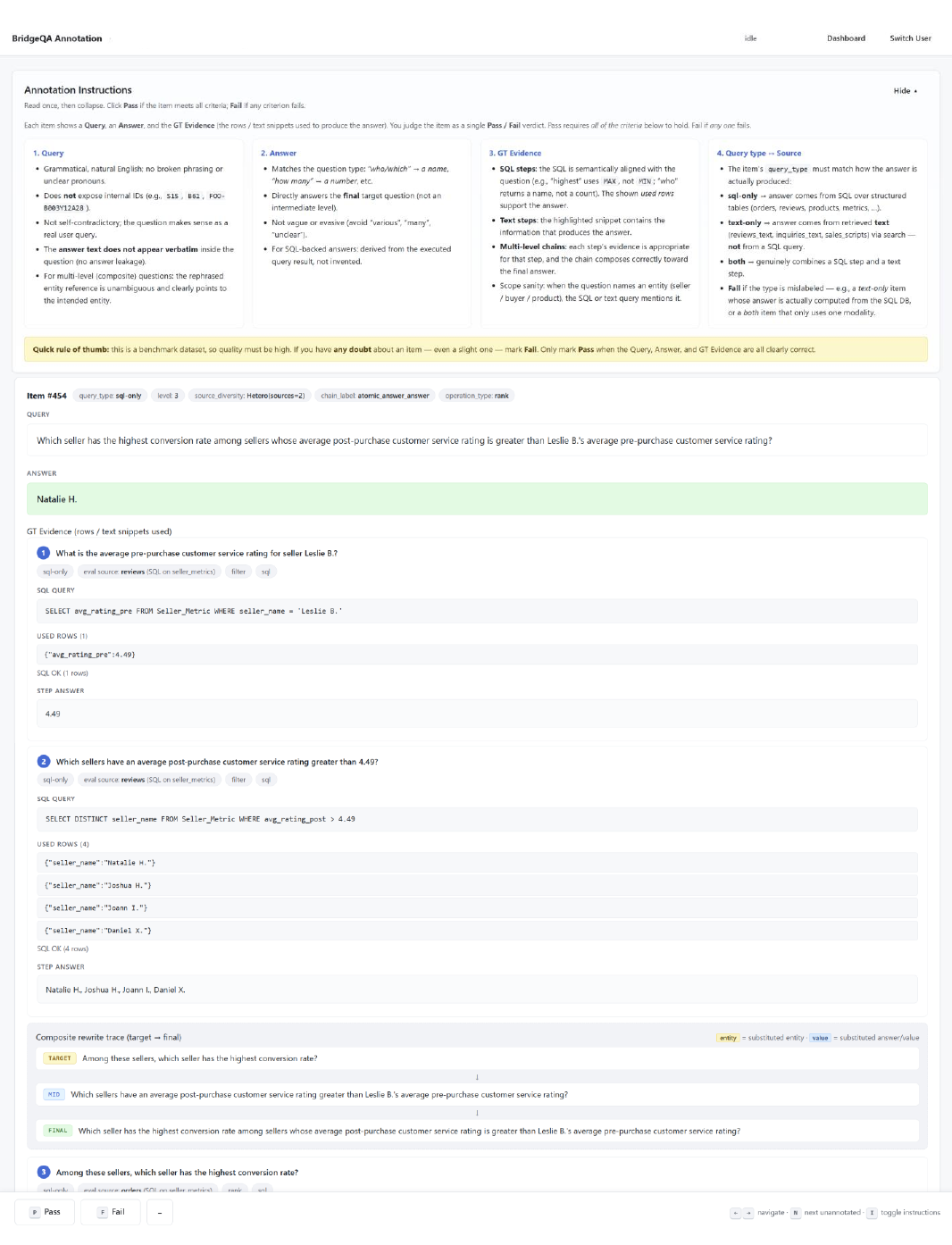}
    \vspace{-0.35cm}
    \caption{Example of annotation template for verifying query quality over 4 criteria (Pass/Fail).}
    \label{fig:annotation_template}
\end{figure*}

\paragraph{Stage 1: NLP-Expert Selection.}
The three annotators review queries using a web-based annotation interface (Figure \ref{fig:annotation_template}). Each query is presented together with its gold answer and the ground-truth evidence used to derive it, including executed SQL rows for structured steps and retrieved text snippets for unstructured steps. Each annotator judges every query along the four criteria below and gives a single Pass/Fail verdict, where a query fails if it violates any criterion.
\begin{itemize}[leftmargin=*, noitemsep, topsep=0pt]
    \item \textbf{Query quality}: whether the query is natural, unambiguous, and free from malformed references.
    \item \textbf{Answer correctness}: whether the gold answer is fully supported by the presented evidence.
    \item \textbf{Evidence-question alignment}: whether the query is answerable from the provided evidence sources alone.
    \item \textbf{Modality-source consistency}: whether the assigned reasoning level, source diversity, and evidence modality match the required evidence path.
\end{itemize}
Since all reviewed queries had already passed the strict verification filters (Section \ref{subsec:atomic_qa}), most clear-cut failures were removed before annotation, which contributes to the high agreement among annotators. Annotators were compensated at \$7.50 per hour, exceeding the U.S.\ federal minimum wage.

\paragraph{Stage 2: Domain-Expert Verification.}
The domain expert holds over ten years of industry experience in retail operations at GS Retail, covering seller-side analytics in practice. For this audit, the 120 queries are drawn evenly across the three levels (40 per level). The expert then gives a binary judgment on each query, whether a seller-side analyst would ask it in practice, and records a brief reason for each query judged otherwise. Throughout, each query is reviewed independently of the Stage 1 results.

\paragraph{Annotation Protocol and Ethics.}
All annotation stages are limited to quality assessment of QA instances and evidence sources, which contain no personally identifiable information or sensitive user data. Annotators judge dataset quality only according to the criteria above, rather than providing personal opinions or evaluating real users, making the task low-risk for all participants.

\subsection{Comparison with Human-Authored Queries}
\label{appendix:human_comparison}
To extend the verification of \dataname{} beyond our own annotators to its fidelity to real retail workloads, we compare its queries against the Spider 2.0-lite retail subset \cite{lei2025spider}, where SQL-expert annotators wrote questions over real retail databases.

\paragraph{Indistinguishability from Human Queries.}
We randomly sample 100 \dataname{} queries per level (300 in total) and pair each with a human-authored query from the subset. For each pair, three LLM judges (Claude Opus 4.7, GLM-5.2, Grok-4.5) independently identify which query appears human-written or mark the two as indistinguishable, and their labels are aggregated by majority vote. The judges select our query as human-written or find the pair indistinguishable in 52\% of cases on average (61\%, 44\%, and 50\% at Level-1, 2, and 3, respectively), indicating that \dataname{} queries read on par with expert-written ones.

% \begin{table}[!t]
% \centering
% \small
% \caption{Pairwise comparison with human-authored queries by three LLM judges. \emph{Win+Tie}: our query judged human-written or the pair indistinguishable.}
% \label{tab:human_comparison}
% \begin{tabular*}{\columnwidth}{@{\extracolsep{\fill}}lcccc@{}}
% \toprule
%  & Level-1 & Level-2 & Level-3 & Avg. \\
% \midrule
% Win+Tie (\%) & 61 & 44 & 50 & 52 \\
% \bottomrule
% \end{tabular*}
% \end{table}

\paragraph{Query-Pattern Correspondence.}
Query patterns in \dataname{} find close counterparts in the same reference. Table \ref{tab:pattern_correspondence} presents illustrative pairs sharing the same underlying pattern, from Level-1 ranking and filtering to Level-3 chained lookups. Beyond these pairs, \dataname{} also includes KPI aggregation (\emph{e.g}., refund and conversion rates), temporal lookups, and free-text review and inquiry questions that fall outside SQL-only workloads.

\begin{table*}[!t]
\centering
\small
\caption{Query-pattern correspondence between \dataname{} and human-authored queries in real retail workloads. Parentheses indicate the source database of each human-authored query in Spider 2.0-lite.}
\label{tab:pattern_correspondence}
\begin{tabularx}{\textwidth}{@{}l >{\raggedright\arraybackslash}p{2.3cm} >{\raggedright\arraybackslash}X >{\raggedright\arraybackslash}X@{}}
\toprule
Level & Shared query pattern & \dataname{} (ours) & Human-authored workload \\
\midrule
L1 & Ranking by frequency & Which buyer has sent the most pre-purchase inquiries to Ashley X.? & Identify the top-3 customers with the highest number of delivered orders. (Olist) \\
\addlinespace[2pt]
L1 & Threshold filtering & How many broadcasts did Michael K. conduct with a discount rate greater than 20\%? & Which liquor categories each contribute at least 1\% of monthly sales volume? (Iowa Liquor) \\
\addlinespace[2pt]
L2 & Best entity within a resolved group & Among the buyers who gave Diana G. the lowest product-quality rating, which buyer has the highest conversion rate? & Find the top-selling product among customers who bought ``YouTube Men's Vintage Henley''. (GA360) \\
\addlinespace[2pt]
L3 & Chained lookups with aggregation & Most common delivery date for orders of the most expensive product Jacob Z. ordered from the seller who received the highest total spending from him. & Most-purchased other products by customers who bought the Google Red Speckled Tee, per month. (GA4) \\
\bottomrule
\end{tabularx}
\end{table*}

\subsection{Audit of Reasoning Paths}
\label{app:path_audit}

Since some questions may admit paths other than the one used during construction, we audit all 16,800 runs to examine whether correct runs follow their labeled sources and reasoning levels.

\paragraph{Source Labels.}
Only 5\% of correct runs reach the answer through a source outside the gold label, mainly when text information also appears in table columns, as in the shipping-cost question in Table \ref{tab:retailbridge_qa_examples}. Since such paths are also valid, we add the identified sources to the gold label as alternative sources rather than treating these runs as label violations. Under both the extended labels and the exclusion of such cases, the level ordering and backbone ranking remain unchanged.

\begin{wraptable}{r}{0.42\textwidth}
\centering
\scriptsize
\vspace*{-0.4cm}
\caption{Number of steps taken by correct runs at each reasoning level.}
\vspace*{-0.25cm}
\label{tab:path_steps}
\setlength{\tabcolsep}{2.5pt}
\begin{tabular*}{\linewidth}{@{\extracolsep{\fill}}lccc@{}}
\toprule
Level & 1 Step & 2 Steps & 3+ Steps \\
\midrule
Level-1 & 89.2\% & 8.8\% & 2.0\% \\
Level-2 & 12.7\% & 64.4\% & 22.9\% \\
Level-3 & 7.9\% & 8.3\% & 83.8\% \\
\bottomrule
\end{tabular*}
\vspace*{-0.3cm}
\end{wraptable}

\paragraph{Reasoning Level Labels.}
We count the steps each correct run takes, where one step is one SQL query or retrieval. As shown in Table \ref{tab:path_steps}, most correct runs take at least as many steps as their level (87.3\% at Level-2 and 83.8\% at Level-3). This suggests that the reasoning paths of correct runs largely align with the intended depth of each level.

\subsection{Generator-Family Analysis}
\label{app:generator}

Since question generators and evaluated backbones partly share model families, we examine whether backbones gain an advantage on questions generated by their own family. Each question is generated by one of three LLMs (Appendix \ref{appendix:data_statistics}), so we compare accuracy across question generators and estimate each family's advantage on its own questions.

\paragraph{Accuracy by Generator.}
As shown in Table \ref{tab:generator}, backbone rankings remain highly consistent across generator groups (Kendall's $W$ = 0.987), so which backbones perform well does not depend on the question generator.

\begin{table*}[t]
\centering
\scriptsize
\begin{minipage}[t]{0.57\textwidth}
\centering
\caption{LLM-judge accuracy of each backbone by question generator (560 questions per generator; rank in parentheses).}
\vspace*{-0.25cm}
\label{tab:generator}
\setlength{\tabcolsep}{2.5pt}
\renewcommand{\arraystretch}{1.05}
\begin{tabular}{lccc}
\toprule
Backbone & Qwen3.6-Max & DeepSeek-V4-Pro & Gemini-3.1-Pro \\
\midrule
Qwen3.6-35B-A3B & 0.212 (\#10) & 0.245 (\#10) & 0.252 (\#10) \\
Kimi-K2.5 & 0.237 (\#9) & 0.266 (\#9) & 0.263 (\#9) \\
Gemma-4-31B-IT & 0.468 (\#3) & 0.490 (\#3) & 0.536 (\#2) \\
GPT-oss-120B & 0.418 (\#6) & 0.421 (\#6) & 0.446 (\#6) \\
DeepSeek-V4-Flash & 0.430 (\#5) & 0.457 (\#4) & 0.461 (\#5) \\
DeepSeek-V4-Pro & 0.446 (\#4) & 0.443 (\#5) & 0.509 (\#4) \\
Gemini-3.1-Flash & 0.346 (\#8) & 0.366 (\#8) & 0.361 (\#8) \\
Qwen3.6-Flash & 0.384 (\#7) & 0.407 (\#7) & 0.380 (\#7) \\
Qwen3.6-Plus & 0.502 (\#1) & 0.491 (\#2) & 0.539 (\#1) \\
GPT-5.4 & 0.477 (\#2) & 0.511 (\#1) & 0.536 (\#2) \\
\midrule
Mean & 0.392 & 0.410 & 0.428 \\
\bottomrule
\end{tabular}
\end{minipage}
\hfill
\begin{minipage}[t]{0.405\textwidth}
\centering
\caption{Self-family advantage of backbones on their own family's questions over other generators' questions.}
\vspace*{-0.25cm}
\label{tab:self_family}
\setlength{\tabcolsep}{3pt}
\renewcommand{\arraystretch}{1.05}
\begin{tabular}{l>{\raggedright\arraybackslash}p{2.5cm}c}
\toprule
Generator & Backbones & \makecell{Advantage\\(\%p)} \\
\midrule
Qwen & Qwen3.6-35B-A3B, Qwen3.6-Flash, Qwen3.6-Plus & +1.0 \\
DeepSeek & DeepSeek-V4-Flash, DeepSeek-V4-Pro & $-$1.1 \\
Gemini & Gemini-3.1-Flash & $-$2.8 \\
\midrule
All families & 6 backbones & $-$0.4 \\
\bottomrule
\end{tabular}
\end{minipage}
\end{table*}

\paragraph{Self-Family Advantage.}
We estimate each family's advantage on its own questions through a regression with question and backbone fixed effects, which removes generator difficulty. As shown in Table \ref{tab:self_family}, a one-sided test with question-clustered standard errors finds no significant advantage for any family (all $p$ > 0.2).

\subsection{Robustness to Data Noise}
\label{app:noise}

Real enterprise records contain noise that our simulated data may not reproduce. To quantify this gap, we inject noise into 25\% of records for every noise type, so that types can be compared fairly, covering name and category variants, mixed date formats, blanked names, and typos and colloquialisms in text. On a stratified 200-question subset, \algname{} drops from 0.410 to 0.230, mostly because the agent treats corrupted names and categories as exact keys and misses records. Realistic noise thus makes \dataname{} harder, and we release this noisy variant alongside the clean benchmark.
% TODO: specify the backbone and the metric (LLM-judge?).

\section{Experimental Setup}
\label{appendix:experimental_setup}

\subsection{LLM Configurations}
\label{appendix:llm_config}
Table \ref{tab:llm_config} lists the LLMs used across simulation, QA generation, QA verification, and \algname{} evaluation, together with their roles. Within each stage, all models receive the same prompt format to ensure fair comparison.

\paragraph{Stage-Specific Models.}
For simulation, seller and buyer agents are instantiated with a diverse set of open-source and proprietary backbones to mirror the variety of real interactions. For QA generation and verification, we use stronger instruction-following models to improve quality, correctness, and consistency. For the LLM benchmark, the listed backbone models are swapped into \algname{} on \dataname{} under a unified setup.

\paragraph{Deployment.}
All proprietary models are accessed through their respective APIs, while open-source models are deployed locally on NVIDIA RTX PRO 6000 Blackwell Server Edition GPUs with 96GB VRAM. We use each model's default decoding configuration.

\begin{table*}[!t]
\centering
\caption{LLM configurations across the three pipeline stages: simulation, benchmark construction (QA generation and verification), and \algname{} evaluation. Models used in multiple stages are listed under each relevant stage.}
\resizebox{\textwidth}{!}{%
\begin{tabular}{cclllll}
\toprule
Stage & Type & Model & Usage & Model Checkpoint & Source & Reference \\
\midrule
\multirow{4}{*}{\makecell[c]{Simu-\\lation}}
& \multirow{3}{*}{\makecell[c]{Open-\\source}}
                & DeepSeek-V3.2   & Simulation            & \texttt{deepseek-ai/DeepSeek-V3.2}     & HuggingFace   & \citet{liu2025deepseek} \\
&               & GPT-oss-120B    & Simulation            & \texttt{openai/gpt-oss-120b}           & HuggingFace   & \citet{agarwal2025gpt} \\
&               & Qwen3-235B      & Simulation            & \texttt{Qwen/Qwen3-235B-A22B-Instruct} & HuggingFace   & \citet{yang2025qwen3} \\
\cmidrule(lr){2-7}
& Proprietary    & Gemini-3-Flash  & Simulation            & \texttt{gemini-3-flash-preview}        & Google API    & \citet{google2025geminiflash} \\
\midrule
\multirow{6}{*}{\makecell[c]{Benchmark\\Construction}}
& Open-source 
                & Qwen3-Embedding-8B & QA Sampling      & \texttt{qwen/qwen3-embedding-8b} & HuggingFace   & \citet{zhang2025qwen3} \\
\cmidrule(lr){2-7}
& \multirow{5}{*}{Proprietary}
                & Qwen3.6-Max      & QA Generation    & \texttt{qwen3.6-max-preview}    & Qwen API      & \citet{qwen2026qwen36maxpreview} \\
&               & DeepSeek-V4-Pro  & QA Generation    & \texttt{deepseek-v4-pro}        & DeepSeek API  & \citet{deepseek2026deepseekv4} \\
&               & Gemini-3.1-Pro   & QA Generation    & \texttt{gemini-3.1-pro-preview} & Google API    & \citet{google2026geminipro} \\
&               & GPT-5.5          & QA Verification  & \texttt{gpt-5.5}                & OpenAI API    & \citet{openai2026gpt55} \\
&               & Claude Opus 4.7  & QA Verification  & \texttt{claude-opus-4.7}        & Anthropic API & \citet{anthropic2026claudeopus47} \\
\midrule
\multirow{10}{*}{\makecell[c]{Retail-\\RAG}}
& \multirow{5}{*}{\makecell[c]{Open-\\source}}
                & GPT-oss-120B      & RetailRAG             & \texttt{openai/gpt-oss-120b}             & HuggingFace   & \citet{agarwal2025gpt} \\
&               & Qwen3.6-35B-A3B   & RetailRAG             & \texttt{qwen/qwen3.6-35b-a3b}            & HuggingFace   & \citet{qwen2026qwen36} \\
&               & Kimi-K2.5         & RetailRAG             & \texttt{moonshotai/kimi-k2.5}            & HuggingFace   & \citet{team2026kimi} \\
&               & Gemma-4-31B-IT    & RetailRAG             & \texttt{google/gemma-4-31b-it}           & HuggingFace   & \citet{google2026gemma4} \\
&               & DeepSeek-V4-Flash & RetailRAG             & \texttt{deepseek/deepseek-v4-flash}      & DeepSeek API  & \citet{deepseek2026deepseekv4} \\
\cmidrule(lr){2-7}
& \multirow{5}{*}{Proprietary}
                & DeepSeek-V4-Pro   & RetailRAG             & \texttt{deepseek-v4-pro}                 & DeepSeek API  & \citet{deepseek2026deepseekv4} \\
&               & Gemini-3.1-Flash  & RetailRAG             & \texttt{gemini-3.1-flash-preview}        & Google API    & \citet{google2026gemini3} \\
&               & Qwen3.6-Flash     & RetailRAG             & \texttt{qwen3.6-flash}                   & Qwen API      & \citet{yang2025qwen3} \\
&               & Qwen3.6-Plus      & RetailRAG; LLM Judge  & \texttt{qwen3.6-plus}                    & Qwen API      & \citet{qwen2026qwen36plus} \\
&               & GPT-5.4           & RetailRAG             & \texttt{gpt-5.4}                         & OpenAI API    & \citet{openai2026gpt54} \\
\bottomrule
\end{tabular}%
}
\label{tab:llm_config}
\end{table*}

\subsection{Implementation Details on Baselines}
\label{appendix:baselines}
This section details the implementation of each baseline in Table \ref{tab:baseline_comparison}. All methods share a common LLM backbone, dense embedder, and LLM-as-judge, with method-specific parameters following each paper's published defaults.

\paragraph{Common Setup.}
All baselines and \algname{} share the same backbone LLM (\texttt{GPT-oss-120B}), embedder (\texttt{Qwen3-Embedding\--8B}), and LLM-as-judge (\texttt{Qwen3.6-Plus}) to ensure fair comparison. SQL execution runs on SQLite endpoints over the five tabular sources, and text retrieval uses a dense index over the three textual sources.

\paragraph{Baseline-Specific Setups.}
Beyond the common setup, each baseline follows its design as summarized below.
\begin{itemize}[leftmargin=*, noitemsep, topsep=0pt]
    \item \textbf{Text-to-SQL}: a one-shot Text-to-SQL baseline with an LLM-based source-routing step that selects one target SQL endpoint per query, providing a lower bound on SQL-only performance.
    \item \textbf{Dense RAG}: a dense retrieval baseline with the same routing step over text sources, retrieving top-$k{=}10$ passages with the shared embedder.
    \item \textbf{MAC-SQL} \cite{wang2025macsql}: the three-agent pipeline of Selector, Decomposer, and Refiner, with prompts following the official repository.
    \item \textbf{TableRAG} \cite{yu2025tablerag}: the iterative explore loop with top-$k$ passages and bounded sub-question iterations routed through a Text-to-SQL tool.
    \item \textbf{HetaRAG} \cite{yan2025hetarag}: the heterogeneous-store pipeline of query rewrite, dense retrieval, Text-to-SQL, rerank, and answer composition.
\end{itemize}
For all baselines, we preserve each paper's core design and prompts, applying only minor adaptations needed to run them in our environment (\emph{e.g.}, swapping the backbone for parity, or adapting tool bodies to our SQL/text stores). These modifications keep the methods faithful to their original intent while ensuring fair comparison.

\subsection{Reliability of LLM-Judge Evaluation}
\label{app:judge}

Since our main evaluation relies on a single LLM judge (Qwen3.6-Plus) that also appears among the evaluated backbones, we validate it from four perspectives, namely agreement with humans, agreement with independent judges, consistency with rule-based scoring, and the separation of correctness from evidential support.

\paragraph{Human Agreement.}
Three graduate-level NLP annotators independently labeled the correctness of answers from the ten backbones on 250 questions (2,500 items), stratified by reasoning level, source diversity, and final-hop modality. 
\begin{wraptable}{r}{0.5\textwidth}
\centering
\scriptsize
\vspace*{-0.4cm}
\caption{Agreement between LLM judges and human labels.}
\vspace*{-0.25cm}
\label{tab:judge_human}
\setlength{\tabcolsep}{2.5pt}
\begin{tabular*}{\linewidth}{@{\extracolsep{\fill}}lcccc@{}}
\toprule
Judge Model & Cohen's $\kappa$ & Precision & Recall & F1 \\
\midrule
Qwen3.6-Plus (Paper) & 0.960 & 0.971 & 0.984 & 0.978 \\
Muse-Spark-1.3 & 0.959 & 0.991 & 0.963 & 0.977 \\
MiMo-V2.6-Flash & 0.939 & 0.944 & 0.990 & 0.966 \\
\bottomrule
\end{tabular*}
\vspace*{-0.3cm}
\end{wraptable}
Given their high agreement (Krippendorff's $\alpha$ = 0.975), we use the majority label as ground truth and compare the paper judge with two independent judges, Muse-Spark-1.3 and MiMo-V2.6-Flash. As shown in Table \ref{tab:judge_human}, the paper judge shows very high agreement with humans (Cohen's $\kappa$ = 0.960), as the gold answers in \dataname{} are well-defined.

\begin{wraptable}{r}{0.59\textwidth}
\centering
\scriptsize
\vspace*{-0.4cm}
\caption{Multi-judge agreement and self-preference bias of the paper judge (Qwen3.6-Plus).}
\vspace*{-0.25cm}
\label{tab:judge_multi}
\setlength{\tabcolsep}{1.5pt}
\begin{tabular*}{\linewidth}{@{\extracolsep{\fill}}lcccc@{}}
\toprule
Reference Judge & \makecell{Agreement\\(Cohen's $\kappa$)} & \makecell{Ranking\\(Kendall's $\tau$)} & \makecell{Bias\\(Qwen family)} & \makecell{Bias\\(Qwen3.6-Plus)} \\
\midrule
Muse-Spark-1.3 & 0.949 & 0.954 & $-$0.1 & +0.7 \\
MiMo-V2.6-Flash & 0.932 & 0.928 & $-$0.8 & $-$1.2 \\
\bottomrule
\end{tabular*}
\vspace*{-0.3cm}
\end{wraptable}

\paragraph{Multi-Judge Agreement and Self-Preference.}
We re-judge all 16,800 answers (10 backbones $\times$ 1,680 questions) with the two independent judges. To test whether the Qwen judge favors Qwen backbones, we define bias as the additional score it assigns to them relative to the reference judge (\%p), where a positive value indicates self-preference. Table \ref{tab:judge_multi} shows that both independent judges agree almost perfectly with the paper judge and preserve the backbone ranking. Question-level bootstrap intervals show no significantly positive bias, and the bias of Qwen3.6-Plus changes sign across reference judges (+0.7 vs. $-$1.2), indicating no detectable self-preference. We further run the paper judge three times independently, where 98.4\% of verdicts are identical across runs and the backbone ranking is unchanged.

\begin{wraptable}{r}{0.55\textwidth}
\centering
\scriptsize
\vspace*{-0.4cm}
\caption{Rule-based vs. LLM-judge accuracy by answer type.}
\vspace*{-0.25cm}
\label{tab:judge_rule}
\setlength{\tabcolsep}{2.5pt}
\begin{tabular*}{\linewidth}{@{\extracolsep{\fill}}lrcc@{}}
\toprule
Answer Type & $n$ & Rule-based & LLM Judge \\
\midrule
Number & 4,650 & 0.412 & 0.417 \\
Temporal & 840 & 0.647 & 0.647 \\
Set & 2,110 & 0.409 & 0.411 \\
Entity & 3,290 & 0.539 & 0.547 \\
\midrule
Total (excl. free text) & 10,890 & 0.468 & 0.473 \\
\bottomrule
\end{tabular*}
\vspace*{-0.3cm}
\end{wraptable}

\paragraph{Rule-Based Scoring by Answer Type.}
For number, temporal, set, and entity answers, we apply deterministic normalization and compare the resulting rule-based scores with those of the LLM judge. This covers 10,890 answers, excluding free-text answers, which have no canonical form for deterministic matching. As shown in Table \ref{tab:judge_rule}, the two differ by less than 0.01 on every answer type, with an overall accuracy of 0.468 under rule-based scoring and 0.473 under the LLM judge. The largest gap appears on entity answers (0.539 vs. 0.547), and temporal answers yield identical scores (0.647). The LLM judge scores equal to or slightly above rule-based scoring on every type, so it does not inflate accuracy beyond what deterministic matching supports.

\begin{wraptable}{r}{0.52\textwidth}
\centering
\scriptsize
\vspace*{-0.4cm}
\caption{Correctness vs. support by retrieved evidence over all 16,800 answers.}
\vspace*{-0.25cm}
\label{tab:judge_support}
\setlength{\tabcolsep}{2.5pt}
\begin{tabular*}{\linewidth}{@{\extracolsep{\fill}}lccc@{}}
\toprule
 & Supported & Not Supported & Total \\
\midrule
Correct & 7,057 (42.0\%) & 148 (0.9\%) & 7,205 (42.9\%) \\
Incorrect & 4,145 (24.7\%) & 5,450 (32.4\%) & 9,595 (57.1\%) \\
\midrule
Total & 11,202 (66.7\%) & 5,598 (33.3\%) & 16,800 (100\%) \\
\bottomrule
\end{tabular*}
\vspace*{-0.3cm}
\end{wraptable}

\paragraph{Correctness versus Evidential Support.}
To separate correctness from grounding, an independent judge checks whether every claim in each answer is supported by the evidence the system actually retrieved, without access to the gold answer. As shown in Table \ref{tab:judge_support}, only 0.9\% of all answers are correct without support from the retrieved evidence, so correctness on \dataname{} reflects grounded answering rather than guessing or prior knowledge.

\section{Additional Analyses}
\label{appendix:additional_analyses}

\begin{table*}[t]
\centering
\scriptsize
\caption{Cross-dataset evaluation on the BIRD benchmark ($n{=}500$) across $11$ domains. \algname{} outperforms all baselines in overall accuracy, demonstrating that its retail-specific designs generalize to other domains.}
\setlength{\tabcolsep}{3.8pt}
\renewcommand{\arraystretch}{1.05}
\begin{tabularx}{\textwidth}{@{}l l ccccccccccc c@{}}
\toprule
\textbf{Category} & \textbf{Method} & \textbf{F1} & \textbf{Cards} & \textbf{Hero} & \textbf{Football} & \textbf{Throm.} & \textbf{Code} & \textbf{Student} & \textbf{Tox.} & \textbf{Fin.} & \textbf{Debit} & \textbf{Schools} & \textbf{Avg.} \\
\midrule
\multirow{2}{*}{Naive}
 & Text-to-SQL  & 0.606 & 0.577 & 0.865 & 0.804 & 0.520 & 0.625 & 0.729 & 0.450 & 0.406 & 0.600 & 0.367 & 0.615 \\
 & Dense RAG    & 0.136 & 0.096 & 0.019 & 0.098 & 0.060 & 0.082 & 0.062 & 0.150 & 0.000 & 0.100 & 0.000 & 0.078 \\
\midrule
\multirow{4}{*}{Agentic}
 & ReAct        & 0.455 & 0.365 & 0.615 & 0.549 & 0.340 & 0.469 & 0.562 & 0.325 & 0.344 & 0.233 & 0.300 & 0.432 \\
 & MAC-SQL      & 0.591 & 0.558 & 0.846 & 0.804 & 0.540 & 0.653 & 0.812 & 0.350 & 0.375 & 0.533 & 0.433 & 0.612 \\
 & TableRAG     & 0.121 & 0.115 & 0.096 & 0.157 & 0.020 & 0.061 & 0.042 & 0.100 & 0.062 & 0.033 & 0.000 & 0.080 \\
 & HetaRAG      & 0.515 & 0.577 & 0.865 & 0.784 & 0.480 & 0.653 & 0.812 & 0.425 & 0.562 & 0.533 & 0.533 & 0.622 \\
\midrule
Ours
 & \algname{}   & 0.545 & 0.596 & 0.942 & 0.800 & 0.560 & 0.653 & 0.854 & 0.475 & 0.406 & 0.567 & 0.533 & 0.645 \\
\bottomrule
\end{tabularx}
\label{tab:bird_minidev}
\end{table*}

\subsection{Baseline Evaluation on BIRD}
\label{appendix:bird_evaluation}
To assess whether \algname{}'s retail-specific designs generalize beyond \dataname{}, we evaluate it on BIRD \cite{li2023bird}, a multi-domain SQL benchmark. We randomly sample $500$ queries spanning $11$ domains and evaluate all baselines under the same setup as Appendix \ref{appendix:baselines}. As BIRD consists of single-source SQL queries, this evaluation examines whether \algname{}, designed for multi-source chaining, remains effective on simpler queries across diverse domains.

\paragraph{Results.}
Table \ref{tab:bird_minidev} reports per-domain LLM-judge scores. \algname{} reaches the highest overall accuracy ($0.645$), outperforming the strongest baseline HetaRAG ($0.622$) and the SQL-specialist Text-to-SQL ($0.615$). Domain-level results further show that \algname{} achieves the best or tied-best score in $7$ of $11$ domains. The margins are naturally modest, as most queries reduce to single-step SQL where \algname{}'s planning advantages have little room to operate; that it still leads even in this regime shows that the retail-specific designs incur no penalty on simple queries and generalize to other structured-data domains.

\subsection{Stability of RetailRAG}
\label{appendix:stability}
\begin{wraptable}{r}{0.52\textwidth}
\centering
\scriptsize
\vspace*{-0.4cm}
\caption{Stability of \algname{} across five runs on \dataname{} (gpt-oss-120b). Bottom rows report the bootstrap mean and 95\% percentile CI ($B{=}10{,}000$). Single-run results in Table \ref{tab:llm_benchmark} lie within all CIs.}
\vspace*{-0.25cm}
\setlength{\tabcolsep}{2.5pt}
\renewcommand{\arraystretch}{0.95}
\begin{tabular*}{\linewidth}{@{\extracolsep{\fill}}l cccccccc@{}}
\toprule
\multirow{2}{*}{Run}
 & \multicolumn{2}{c}{Level-1} & \multicolumn{2}{c}{Level-2}
 & \multicolumn{2}{c}{Level-3} & \multicolumn{2}{c}{Avg.} \\
\cmidrule(lr){2-3} \cmidrule(lr){4-5} \cmidrule(lr){6-7} \cmidrule(lr){8-9}
 & Acc & LLM & Acc & LLM & Acc & LLM & Acc & LLM \\
\midrule
Run 1 & 0.688 & 0.725 & 0.300 & 0.474 & 0.165 & 0.285 & 0.298 & 0.429 \\
Run 2 & 0.662 & 0.717 & 0.287 & 0.449 & 0.174 & 0.296 & 0.292 & 0.421 \\
Run 3 & 0.679 & 0.729 & 0.293 & 0.436 & 0.165 & 0.303 & 0.293 & 0.421 \\
Run 4 & 0.683 & 0.725 & 0.293 & 0.450 & 0.165 & 0.290 & 0.294 & 0.421 \\
Run 5 & 0.688 & 0.721 & 0.304 & 0.461 & 0.179 & 0.300 & 0.305 & 0.429 \\
\midrule
Mean        & 0.680 & 0.723 & 0.295 & 0.454 & 0.170 & 0.295 & 0.297 & 0.424 \\
95\% CI lo  & 0.671 & 0.719 & 0.291 & 0.444 & 0.165 & 0.289 & 0.293 & 0.421 \\
95\% CI hi  & 0.687 & 0.727 & 0.300 & 0.464 & 0.175 & 0.300 & 0.301 & 0.427 \\
\bottomrule
\end{tabular*}
\vspace*{-0.3cm}
\label{tab:stability}
\end{wraptable}

Since \algname{} relies on LLM-driven auto-schema briefing, source-routed plan chaining, and diagnostic plan revision, we examine whether its gains are stable across runs rather than single-seed artifacts. We run \algname{} five times on \dataname{} with the gpt-oss-120b backbone under random seeds and default temperature, then bootstrap the mean and 95\% percentile CI over $B{=}10{,}000$ resamples (Table \ref{tab:stability}). Run-to-run variation stays within $\pm 0.01$ on both scores, and the single-run results in Table \ref{tab:llm_benchmark} lie within all CIs, confirming that the reported gains reflect \algname{}'s design rather than random variation.

\subsection{Baselines with RetailRAG Components}
\label{app:baseline_components}

Section \ref{subsec:briefing_injection} injects only the schema briefing into baselines. To further separate framework design from schema access and recovery budget, we equip each baseline with our schema briefing and retry policy.
All methods share the backbone, embedder, and tools of \algname{}. Since token usage is inherent to each framework, we report it alongside accuracy.

As shown in Table \ref{tab:baseline_components}, every baseline remains below \algname{} even with our components, 
\begin{wraptable}{r}{0.6\textwidth}
\centering
\scriptsize
\vspace*{-0.4cm}
\caption{Baselines equipped with \algname{}'s schema briefing and retry (default $\rightarrow$ briefing + retry). \emph{LLM}: LLM-as-judge accuracy. \emph{API Call}: API calls per query.}
\vspace*{-0.25cm}
\label{tab:baseline_components}
\setlength{\tabcolsep}{1.5pt}
\begin{tabular*}{\linewidth}{@{\extracolsep{\fill}}lcccc@{}}
\toprule
Framework & LLM & \makecell{Input\\Token} & \makecell{Output\\Token} & \makecell{API\\Call} \\
\midrule
ReAct & 0.085 $\rightarrow$ 0.165 & 2.4k $\rightarrow$ 6.7k & 1.3k $\rightarrow$ 2.1k & 4.2 $\rightarrow$ 6.2 \\
MAC-SQL & 0.201 $\rightarrow$ 0.270 & 6.1k $\rightarrow$ 20.4k & 2.4k $\rightarrow$ 4.3k & 3.1 $\rightarrow$ 5.0 \\
TableRAG & 0.185 $\rightarrow$ 0.230 & 11.8k $\rightarrow$ 17.7k & 2.9k $\rightarrow$ 3.2k & 6.6 $\rightarrow$ 6.8 \\
% ReWOO & 0.228 $\rightarrow$ 0.266 & 3.3k $\rightarrow$ 12.9k & 1.7k $\rightarrow$ 2.7k & 4.6 $\rightarrow$ 6.5 \\
% PlanRAG & 0.291 $\rightarrow$ 0.362 & 15.2k $\rightarrow$ 38.3k & 3.6k $\rightarrow$ 4.4k & 6.5 $\rightarrow$ 7.7 \\
HetaRAG & 0.293 $\rightarrow$ 0.368 & 19.3k $\rightarrow$ 90.6k & 14.4k $\rightarrow$ 18.7k & 25.2 $\rightarrow$ 27.6 \\
\midrule
\algname{} & 0.429 & 19.8k & 5.4k & 7.45 \\
\bottomrule
\end{tabular*}
\vspace*{-0.5cm}
\end{wraptable}
while their budgets vary widely, from 3 to 28 API calls and from 2k to 91k input tokens. This indicates that schema access and retry alone do not account for the gap, and that the gains of \algname{} stem from source-routed plan chaining and diagnostic plan revision.

\subsection{Oracle Analysis}
\label{app:oracle}

Final answer scores alone cannot tell whether a failure arises from retrieval, planning, entity binding, or answer generation. We therefore provide gold information to one pipeline stage at a time under four conditions, Oracle-source (gold stores per hop), Oracle-plan (gold step plan), Oracle-intermediate (gold intermediate answers), and Oracle-evidence (gold evidence for every hop).
% TODO: specify the backbone (gpt-oss-120b?) and confirm that the scores are LLM-judge accuracy.

\begin{table*}[t]
\centering
\scriptsize
\caption{LLM-judge accuracy of \algname{} under each oracle condition.}
\vspace*{-0.25cm}
\label{tab:oracle}
\setlength{\tabcolsep}{4.3pt}
\renewcommand{\arraystretch}{1.05}
\begin{tabular*}{\textwidth}{@{\extracolsep{\fill}}l cc ccc ccc@{}}
\toprule
\multirow{2}{*}{Condition} & \multicolumn{2}{c}{Level-1} & \multicolumn{3}{c}{Level-2} & \multicolumn{3}{c}{Level-3} \\
\cmidrule(lr){2-3} \cmidrule(lr){4-6} \cmidrule(lr){7-9}
 & SQL & Text & SQL & Text & Comp. & SQL & Text & Comp. \\
\midrule
\algname{} & 0.857 & 0.625 & 0.561 & 0.400 & 0.464 & 0.381 & 0.221 & 0.255 \\
Oracle-source & 0.867 & 0.633 & 0.527 & 0.438 & 0.490 & 0.393 & 0.267 & 0.272 \\
Oracle-plan & 0.876 & 0.750 & 0.690 & 0.329 & 0.611 & 0.594 & 0.250 & 0.444 \\
Oracle-intermediate & -- & -- & 0.469 & 0.442 & 0.619 & 0.502 & 0.433 & 0.469 \\
Oracle-evidence & 0.971 & 0.990 & 0.975 & 0.992 & 0.992 & 0.983 & 0.971 & 0.987 \\
\bottomrule
\end{tabular*}
\end{table*}

As shown in Table \ref{tab:oracle}, \algname{} answers nearly all questions correctly under Oracle-evidence (0.97 to 0.99), showing that the main difficulty lies in retrieving the evidence, with the largest gap for text, while evidence interpretation and answer generation are not a bottleneck. Oracle-source also improves accuracy, and Oracle-plan and Oracle-intermediate yield even larger gains in most cells, revealing further bottlenecks in deciding how to retrieve information (planning) and in carrying each hop's result to the next (entity binding).

Hetero. questions reach fewer required sources in every matched cell, so crossing sources makes routing harder. Their higher accuracy in some cells comes from the retrieval step after routing, as most Hetero. Level-3 text chains pass through sales scripts, the easiest store to retrieve from, while Homo. SQL questions more often require derived metrics, the hardest to compute. This gap thus stems from source composition rather than easier cross-source reasoning.
% TODO: state whether source usage is balanced in the released version.

\section{Prompt Design}
\label{appendix:prompt}
We describe the prompts used in \dataname{} construction and \algname{} reasoning.

\paragraph{Atomic QA Generation.}
Tables \ref{tab:atomic_generation}--\ref{tab:atomic_validator_text} provide the prompts for constructing and verifying the atomic QA pool. Table \ref{tab:atomic_generation} generates SQL atomics as question--SQL pairs and text atomics as question--search-query pairs conditioned on triplet-scoped text records, using type-specific hints in Tables \ref{tab:base_atomic_sql_hints}--\ref{tab:base_atomic_text_hints}. For text atomics, Table \ref{tab:text_identify} additionally generates identify-style questions whose hidden answer is recoverable from text evidence. After generation, SQL atomics are grounded by query execution, with failures repaired using Table \ref{tab:sql_regen}, while text atomics are grounded from the same scoped records using Table \ref{tab:dense_retrieve}. Tables \ref{tab:atomic_validator_sql}--\ref{tab:atomic_validator_text} then validate SQL and text atomics before they enter the approved pool.

\paragraph{Recursive QA Composition.}
Tables \ref{tab:obfuscate}--\ref{tab:composite_validator} provide the prompts for composing higher-level queries from a lower-level QA pool. Table \ref{tab:obfuscate} rewrites a target question by replacing an entity, set, or scalar placeholder with a referring expression derived from an earlier level, supporting both entity-linked and answer-linked composition. In answer-linked composition, the follow-up atomic QA is generated using the hints in Table \ref{tab:answer_linked_hints} before the intermediate answer is rewritten as a referring expression. Table \ref{tab:composite_validator} then verifies the composed question for grammaticality, faithful substitution, preservation of unreplaced constraints, and absence of answer leakage.

\paragraph{RetailRAG Pipeline.}
Tables \ref{tab:auto_schema_briefing}--\ref{tab:final_answer_generation} provide the prompt templates for \algname{}'s agentic workflow, organized around five components: (i) auto-schema briefing in Table \ref{tab:auto_schema_briefing}, which summarizes heterogeneous retail sources into a planner-facing schema context; (ii) adaptive routing in Tables \ref{tab:complexity_router}--\ref{tab:sql_source_pick}, which determines the execution path before evidence gathering; (iii) source-routed plan generation and execution in Tables \ref{tab:planning}--\ref{tab:sql_tool}, which decomposes hard queries into source-specific evidence-gathering steps; (iv) diagnostic plan revision in Tables \ref{tab:diagnose_template}--\ref{tab:plan_revision}, which identifies failed execution paths and repairs the remaining plan; and (v) final answer generation in Table \ref{tab:final_answer_generation}, which converts executed evidence into a grounded user-facing answer.

%% ── Atomic QA Gen(SQL+Text) ──────────────────────────────────────────

\clearpage
\begin{table*}[!p]
\centering
\caption{Prompt for Atomic QA Generation.}
\begin{promptbox}{Atomic Question Generation}
\begin{systembox}
\scriptsize
You are a QA item author for a retail company dataset. You write exactly one atomic question--answer pair grounded in a single primary source (a SQL table or a free-text field of sales\_scripts/reviews/inquiries).

Generate ONE atomic QA about the given primary source.
\begin{itemize}[leftmargin=*, noitemsep, topsep=2pt]
\item \texttt{atomic\_type=sql} $\rightarrow$ \texttt{action=sql\_query}, single-table SQLite query (no JOINs needed).
\item \texttt{atomic\_type=text} $\rightarrow$ \texttt{action=dense\_retrieve}, natural-language search\_query over the free-text content field of sales\_scripts/reviews/inquiries.
\end{itemize}
 
Rules:
\begin{itemize}[leftmargin=*, noitemsep, topsep=2pt]
\item Question must be semantically about the primary source.
\item Valid ops: filter, aggregate, rank, calculate, search, compare. No single-row ID lookups.
\item SQL must return a CONCISE answer (scalar, small ranked list, or small GROUP BY). If natural result $>$30 rows, reframe the question (top-1, count, avg). No raw dumps.
\item Use EXACT schema columns. Escape apostrophes with \texttt{''} (SQLite).
\item Scope filter: all tables carry denormalized name columns --- use WHERE directly, e.g. \texttt{WHERE seller\_name = 'X'} or \texttt{WHERE buyer\_name = 'Y'} or \texttt{WHERE product\_name = 'Z'}. No JOINs required. The scope entity name is given in the Scope description.
\item Every question must be ABOUT the scope entity. Do not expose internal IDs in the question.
\item If the question uses ``each/every/all $<$category$>$'' phrasing, DO NOT add LIMIT. Add LIMIT only when question says ``top N'' etc.
\item For \texttt{atomic\_type=text}: the search\_query MUST include the scope entity's name so retrieval is properly scoped.
\item For \texttt{atomic\_type=text}: every entity name in question\_entities (seller\_name, buyer\_name, product\_name) must appear verbatim and in full in the question. Never abbreviate a product\_name or replace it with a shorter title.
\item For \texttt{atomic\_type=text}: search\_query must be a natural-language retrieval query, not SQL or a Boolean predicate. Do NOT write strings like \texttt{review\_category='broadcast' AND seller\_name='X'}. Instead write a concise phrase such as ``Mindy M. broadcast review of Steven C.'s Mueller thermometer''.
\item For \texttt{atomic\_type=text}: use record metadata in the provided content headers. reviews\_text may include review\_category (broadcast/pre-cs/post-cs/product) and review\_rating; inquiries\_text may include inquiry\_type (pre/post). Make the question specific to the relevant text type when the content supports it, and do not merge different review/inquiry types unless the question explicitly asks across those types.
\item NEVER return raw IDs (buyer\_id, seller\_id, product\_id, script\_id) as the answer. Use the human-readable name columns: buyer\_name, seller\_name, product\_name.
\item For ``highest/lowest/most/least'' queries, use subquery to catch ties: \texttt{WHERE col = (SELECT MAX(col) FROM ...)} instead of ORDER BY + LIMIT 1.
\item Use SELECT DISTINCT when listing entities to avoid duplicates.
\item For rank queries (most recent, highest, lowest, etc.), use a subquery to handle ties correctly --- do NOT use SELECT DISTINCT with ORDER BY + LIMIT 1 (ambiguous sort after dedup). e.g. most-recent product $\rightarrow$ \texttt{SELECT DISTINCT product\_name FROM Script WHERE seller\_name='X' AND broadcast\_datetime = (SELECT MAX(broadcast\_datetime) FROM Script WHERE seller\_name='X')}. This correctly returns all tied top results.
\end{itemize}

Output JSON only:
\begin{verbatim}
{
 "question": "...", 
 "operation_type": "filter|aggregate|rank|search", 
 "action": "sql_query|dense_retrieve", 
 "sql_query": "..." or "search_query": "..."
}
\end{verbatim}
\end{systembox}
\begin{userbox}
\scriptsize
Scope: \textcolor{blue}{\{scope\_description\}}\\
Primary source: \textcolor{blue}{\{source\_name\}}\\
Atomic type: \textcolor{blue}{\{atomic\_type\}}

\textbf{DB description:}\\
\textcolor{blue}{\{primary\_source\_schema\_description\}}

\textbf{Scope profile} (distribution stats --- use for realistic thresholds/values):\\
\textcolor{blue}{\{scope\_profile\}}

\textcolor{blue}{\{hint\}}

Produce ONE operation-level question + executable query.
\end{userbox}
\end{promptbox}
\label{tab:atomic_generation}
\end{table*}
\clearpage

%% ── Entity-Answer Text Atomic QA Gen ─────────────────────────────────────────

\begin{table*}[!p]
\centering
\caption{Prompt for Entity-Answer Text Atomic Question Generation}
\begin{promptbox}{Entity-Answer Text Atomic Question Generation}
\begin{systembox}
\scriptsize
You are a QA item author specializing in identify-style retrieval questions. You describe distinctive content from one text record so that retrieving that record reveals the target entity name --- without ever naming the target entity in the question.

Generate a TEXT IDENTIFY-style atomic QA over a single text record. The ANSWER must be exactly the target entity name shown in the user prompt. The QUESTION must describe distinctive content from the evidence record so a retrieval model can find the right record and recover the entity name from it.

Rules:
\begin{itemize}[leftmargin=*, noitemsep, topsep=2pt]
\item DO NOT mention the target entity name anywhere in the question or search\_query (any case, any partial form). The model must infer it from retrieval. For person names like ``Steven C.'' or ``Mindy M.'', this includes the first name alone (``Steven'', ``Mindy'') and any possessive form (``Steven's'', ``Mindy's''). Use role words like ``the seller'' / ``the buyer'' instead.
\item The user hint lists REQUIRED ANCHORS (seller\_name / buyer\_name / product\_name) --- these are the non-target triplet strings. Each listed value is the exact database string: copy it character-for-character into the question. Never abbreviate product\_name or replace it with a shorter title; use the full line exactly as given. The search\_query is free-form retrieval text and may be phrased naturally, but should still reference the anchors clearly.
\item Use distinctive details from the evidence: ratings, prices, specific short quotes ($\leq$8 words), review\_category, inquiry\_type, named features, dates. Avoid generic phrasing like ``gave a positive review''.
\item For source=reviews\_text: state the review\_category (broadcast / pre-cs / post-cs / product) when available.
\item For source=inquiries\_text: state the inquiry\_type (pre / post).
\item For source=sales\_scripts: anchor on the product and a specific feature or claim from the script.
\item Keep wording concise where possible; required anchor strings may be long (full product titles). Not a verbatim quote of the whole evidence.
\item DO NOT use IDs (script\_id, review\_id, inquiry\_id, etc.).
\item operation\_type is always ``search''. action is always ``dense\_retrieve''.
\end{itemize}
 
Output JSON only:
\begin{verbatim}
{
 "question": "...", 
 "operation_type": "search", 
 "action": "dense_retrieve", 
 "search_query": "..."
}
\end{verbatim}
\end{systembox}

\begin{userbox}
\scriptsize
Scope: \textcolor{blue}{\{scope\_description\}}\\
Primary source: \textcolor{blue}{\{source\_name\}}

Target answer (the entity to be identified): \textcolor{blue}{\{target\_entity\}}\\
Target role: \textcolor{blue}{\{target\_role\}}

DB description:\\
\textcolor{blue}{\{primary\_source\_schema\_description\}}

Evidence record (the question must point to THIS record):\\
\textcolor{blue}{\{evidence\_section\}}

\textcolor{blue}{\{hint\}}

Produce ONE identify-style question whose answer is exactly: \textcolor{blue}{\{target\_entity\}}
\end{userbox}
\end{promptbox}
\label{tab:text_identify}
\end{table*}

%% ── SQL Query Regen ──────────────────────────────────────────

\begin{table*}[!p]
\centering
\caption{Prompt for SQL Query Regeneration after Execution Failure.}
\begin{promptbox}{SQL Query Regeneration after Execution Failure}
\begin{systembox}
\scriptsize
You are a SQLite repair specialist for the retail company dataset. You diagnose a failed query and rewrite it to run correctly against the exact provided schema.

A SQL query failed or returned an error. Given the original question and the EXACT source table schema, write a corrected SQL query.

Rules:
\begin{itemize}[leftmargin=*, noitemsep, topsep=2pt]
\item Use ONLY columns listed in the provided schema --- do NOT invent columns.
\item SQLite syntax: escape apostrophes by doubling (\texttt{''}) --- NEVER use \texttt{\textbackslash'}.
\item Do NOT SELECT raw ID columns; prefer human-readable name columns (buyer\_name, seller\_name, product\_name).
\item Scope filter: denormalized name columns are present in the table --- use WHERE directly (e.g. \texttt{WHERE seller\_name = 'X'}), no JOIN needed for filtering.
\item For ``highest/lowest/most/least'' queries, use subquery to catch ties: \texttt{WHERE col = (SELECT MAX(col) FROM ...)} instead of ORDER BY + LIMIT 1.
\item Use SELECT DISTINCT when listing entities to avoid duplicates.
\item For rank queries, use subquery instead of ORDER BY + LIMIT 1; do NOT combine SELECT DISTINCT with ORDER BY + LIMIT.
\item SQL must return a CONCISE result (scalar, small ranked list, or GROUP BY).
\end{itemize}
 
Output JSON only: \texttt{\{"sql\_query": "..."\}}
\end{systembox}

\begin{userbox}
\scriptsize
Question: \textcolor{blue}{\{question\}}

Source table schema:\\
\textcolor{blue}{\{source\_schema\}}

\textcolor{blue}{\{entity\_hint\}}\\
Previous failed SQL:\\
\textcolor{blue}{\{failed\_sql\}}

Failure reason:\\
\textcolor{blue}{\{error\}}

Write a corrected SQL query that answers the question.
\end{userbox}
\end{promptbox}
\label{tab:sql_regen}
\end{table*}

%% ── Text-source QA Answerer ──────────────────────────────────────────

\begin{table*}[!p]
\centering
\caption{Prompt for Text Atomic Answer Generation, which generates the grounded answer for a text atomic from its scoped evidence records.}
\begin{promptbox}{Text Atomic Answer Generation}
\begin{systembox}
\scriptsize
You are a retrieval answerer. The records below are ALREADY FILTERED to the relevant scope (specific seller/buyer/product). All records belong to that entity. Given these records and a query, return a concise factual answer grounded in the records. Output only the answer text.

Important --- answer exactly what was asked:
\begin{itemize}[leftmargin=*, noitemsep, topsep=2pt]
\item If the query is about what the BUYER asked/inquired/reviewed, extract the buyer's own words or intent. Do NOT substitute the seller's answer or a product description.
\item If the query is about what the SELLER said/scripted/promised, extract the seller's own words. Do NOT summarize the buyer's reaction or reviews.
\item If the query asks about the PRODUCT (features, specs, etc.), you may cite either party but stay factual.
\item Keep the answer focused on the question; do not pad with unrelated context.
\item Answer only with information explicitly supported by the records.
\item Do not infer unstated details.
\item Prefer short sentences directly grounded in the records.
\item Do not add contextual details not needed to answer the question.
\end{itemize}
\end{systembox}
\begin{userbox}
\scriptsize
Query: \textcolor{blue}{\{search\_query\}}

Records (\textcolor{blue}{\{n\_records\}}): \textcolor{blue}{\{records\_json\}}
\end{userbox}
\end{promptbox}
\label{tab:dense_retrieve}
\end{table*}
\clearpage

%% ── Atomic QA Validation ──────────────────────────────────────────

% =====================================================================
%  Generation Hints for Level-1 SQL Atomic Questions
% =====================================================================
\begin{table*}[t]
\centering
\caption{
Generation hints inserted into the \texttt{\{hint\}} field of the Atomic QA Generation prompt when building Level-1 SQL atomic questions.
}
\scriptsize
\setlength{\tabcolsep}{3pt}
\renewcommand{\arraystretch}{1.08}
\begin{tabularx}{\textwidth}{@{}>{\centering\arraybackslash}p{0.18\textwidth} X@{}}
\toprule
\textbf{Applied case} & \textbf{Hint} \\
\midrule

\makecell[tc]{Level-1\\atomic question\\entity exact-match\\block}
&
\hintcell{
\# tier == "seller":\par
{[!]} CRITICAL - use this EXACT string in SQL, character-for-character:\par
\quad WHERE seller\_name = '\{anchor\_id\}'\par
\quad Do NOT add, remove, or change any character - no trailing quotes, no possessive.\par
\par
\# tier == "buyer":\par
{[!]} CRITICAL - use this EXACT string in SQL, character-for-character:\par
\quad WHERE buyer\_name = '\{anchor\_id\}'\par
\quad Do NOT add, remove, or change any character - no trailing quotes, no possessive.\par
\par
\# tier == "product":\par
{[!]} CRITICAL - exact DB values (copy this FULL title literally into SQL):\par
\quad product\_name: '\{ptitle\}'\par
\quad CORRECT form:  WHERE product\_name = '\{ptitle\}'\par
\quad DO NOT truncate the title, DO NOT add a trailing apostrophe.
}
\\

\midrule

\makecell[tc]{Level-1\\SQL atomic\\question}
&
\hintcell{
\# answer-shape hint constants (ans\_hint):\par
\_HINT\_NAME    = "The answer should be an entity NAME (buyer name, seller name, or product name)."\par
\_HINT\_NUMERIC = "The answer should be a numeric value (count, average, total, ratio, etc.)."\par
\_HINT\_DATE    = "The answer should be a DATE or DATETIME (e.g. broadcast date, order date, delivery date)."\par
\# (sellers source) ans\_hint:\par
"The answer should be a PRODUCT NAME. Ask which product of this seller has the highest or lowest \{col\}."\par
\par
\# \_make\_focus\_hint(focus) -> (note leading space):\par
" Focus the question on \textasciigrave\{focus\}\textasciigrave."\par
\par
\# appended when src == "reviews":\par
" Always filter by one specific review\_category value: 'broadcast', 'product', 'pre-cs', or 'post-cs'."\par
\# appended when src == "inquiries":\par
" Always filter by one specific inquiry\_type value: 'pre' or 'post'."
}
\\

\midrule

\makecell[tc]{Level-1\\SQL atomic\\question with\\triplet constraint}
&
\hintcell{
TRIPLET-COMBINATION MODE - still generate a SINGLE-TABLE atomic question. Use ONLY\par
the \textasciigrave\{src\}\textasciigrave\ primary source table; do NOT use JOINs or any other table. The question\par
must be specifically constrained to this triplet entity combination: \{col\}='\{val\}',\par
\{col\}='\{val\}'. The SQL MUST include these exact filters: \{col\} = '\{val\}' AND \{col\} = '\{val\}'.
}
\\

\bottomrule
\end{tabularx}
\label{tab:base_atomic_sql_hints}
\end{table*}

% =====================================================================
%  Generation Hints for Level-1 Text Atomic Questions
% =====================================================================
\begin{table*}[t]
\centering
\scriptsize
\caption{
Generation hints inserted into the \texttt{\{hint\}} field of the Atomic QA
Generation prompt when building Level-1 text atomic questions.
}
\setlength{\tabcolsep}{3pt}
\renewcommand{\arraystretch}{1.08}
\begin{tabularx}{\textwidth}{@{}>{\centering\arraybackslash}p{0.18\textwidth} X@{}}
\toprule
\textbf{Applied case} & \textbf{Hint} \\
\midrule

\makecell[tc]{Level-1\\text atomic\\question\\sales scripts}
&
\hintcell{
\# src == "sales\_scripts":\par
\par
TRIPLET TEXT CONTEXT - this text source is scoped by seller + product only:\par
- seller: \{seller\}\par
- product: \{product\}\par
- buyer is NOT part of sales\_scripts scope; do not mention the buyer in the question or search\_query.\par
The question must explicitly name the seller (\{seller\}) and the product (\{product\}).\par
Ask about what the seller said, emphasized, described, or promised in the sales script.\par
\# + diversity\_note, appended only when diversity\_idx > 0:\par
\par
This is question variation \#\{idx\} - generate a question from a DIFFERENT angle than you\par
would generate by default. Be specific to the actual content of the script.
}
\\

\midrule

\makecell[tc]{Level-1\\text atomic\\question\\reviews text}
&
\hintcell{
\# src == "reviews\_text":\par
\par
TRIPLET TEXT CONTEXT - this text source is scoped by seller + buyer + product:\par
- seller: \{seller\}\par
- buyer: \{buyer\}\par
- product: \{product\}\par
The question must explicitly name the buyer (\{buyer\}), seller (\{seller\}), and product (\{product\}).\par
\# category\_line: if category\_filter -> first line; else -> second line:\par
Focus ONLY on the '\{category\_filter\}' review\_category record.\par
Use review\_category metadata to make the question specific.\par
\# + diversity\_note (diversity\_idx > 0), with one review angle:\par
\par
This is question variation \#\{idx\} - generate a question from a DIFFERENT angle than the default one.\par
\quad [angle 1] Ask about the buyer's overall opinion, satisfaction, or impression.\par
\quad [angle 2] Ask only about the explicit reason, praised point, criticized point, or concrete\par
\quad\quad\quad detail stated in the review. Do not ask for an overall summary.\par
Avoid near-duplicate wording across variations.
}
\\

\midrule

\makecell[tc]{Level-1\\text atomic\\question\\inquiries text}
&
\hintcell{
\# src == "inquiries\_text":\par
\par
TRIPLET TEXT CONTEXT - this text source is scoped by seller + buyer + product:\par
- seller: \{seller\}\par
- buyer: \{buyer\}\par
- product: \{product\}\par
The question must explicitly name the buyer (\{buyer\}), seller (\{seller\}), and product (\{product\}).\par
\# type\_line: if category\_filter -> first line; else -> second line:\par
Focus ONLY on the '\{category\_filter\}' inquiry\_type record.\par
Use inquiry\_type metadata (pre or post) to make the question specific.\par
\# + diversity\_note (diversity\_idx > 0), with one inquiry angle:\par
\par
This is question variation \#\{idx\} - generate a question from a DIFFERENT angle than the default one.\par
\quad [angle 1] Ask about the buyer's main question, request, or concern.\par
\quad [angle 2] Ask about concrete details discussed, such as price, shipping, timing, or policy.\par
\quad [angle 3] Ask about the seller's response, clarification, or resolution to the buyer's concern.\par
Avoid near-duplicate wording across variations.
}
\\

\bottomrule
\end{tabularx}
\label{tab:base_atomic_text_hints}
\end{table*}
\clearpage
% =====================================================================
%  Dynamic Hints for Answer-Linked Follow-up Atomic Question Generation
% =====================================================================
\begin{table*}[t]
\centering
\caption{
Generation hints inserted into the \texttt{\{hint\}} field of the
Atomic QA Generation prompt when generating follow-up atomic questions.
The hint is selected according to the answer type of the corresponding
Level-1 atomic question: Set, or Scalar.
}
\scriptsize
\setlength{\tabcolsep}{3pt}
\renewcommand{\arraystretch}{1.08}
\begin{tabularx}{\textwidth}{@{}>{\centering\arraybackslash}p{0.18\textwidth} X@{}}
\toprule
\textbf{Applied case} & \textbf{Hint} \\
\midrule

\makecell[tc]{Set-answer\\follow-up\\atomic question\\AGG variant}
&
\hintcell{
{[!]} CRITICAL - \{col\} filter MUST use this exact IN clause (do NOT hardcode names):\par
\quad WHERE \{col\} IN (\{in\_clause\})\par
\par
\{review\_hint\}\par
These \{n\} \{entity\_label\}s form a SET. Generate an AGGREGATION question:\par
\quad - Returns a SINGLE scalar (COUNT / SUM / AVG / MIN / MAX)\par
\quad - About ALL \{n\} \{entity\_label\}s as a group - not about any individual\par
\quad - Use '\{placeholder\}' to refer to the group in the question text\par
\par
SQL examples (adapt to the actual columns in the data above):\par
\quad SELECT COUNT(*) FROM ... WHERE \{col\} IN (\{in\_clause\})\par
\quad SELECT AVG(review\_rating) FROM ... WHERE \{col\} IN (\{in\_clause\})\par
\quad\quad [+ AND review\_category = '...'  when source = reviews]
}
\\

\midrule

\makecell[tc]{Set-answer\\follow-up\\atomic question\\FILTER variant}
&
\hintcell{
{[!]} CRITICAL - \{col\} filter MUST use this exact IN clause (do NOT hardcode names):\par
\quad WHERE \{col\} IN (\{in\_clause\})\par
\par
\{review\_hint\}\par
These \{n\} \{entity\_label\}s form a SET. Generate a FILTERING/NARROWING question:\par
\quad - Returns one or more \{entity\_label\} names from this set (NOT a scalar)\par
\quad - Use a single condition: argmax/argmin OR a numeric threshold\par
\quad - Use '\{placeholder\}' to refer to the group in the question text\par
\quad - Do NOT use LIMIT - use a WHERE condition instead:\par
\quad\quad argmax    -> AND metric = (SELECT MAX(metric) FROM ... WHERE \{col\} IN (...))\par
\quad\quad threshold -> AND metric > value\par
\par
SQL examples:\par
\quad SELECT DISTINCT \{col\} FROM ... WHERE \{col\} IN (\{in\_clause\})\par
\quad\quad AND total\_price = (SELECT MAX(total\_price) FROM ... WHERE \{col\} IN (\{in\_clause\}))\par
\quad SELECT DISTINCT \{col\} FROM ... WHERE \{col\} IN (\{in\_clause\})\par
\quad\quad [+ AND review\_category='...' AND review\_rating >= 4  when source = reviews;\par
\quad\quad\quad else  AND refund\_rate > 20]
}
\\

\midrule

\makecell[tc]{Scalar-answer\\follow-up\\atomic question}
&
\hintcell{
{[!]} SCALAR THRESHOLD HINT - use this numeric threshold from a previous step:\par
\quad Threshold value (MUST appear verbatim in the question): \{scalar\_val\}\par
\quad Threshold measures: '\{measure\_label\}'\par
\par
\{review\_hint\}\par
Generate a question that finds which entities have \{measure\_label\} \{op\} \{scalar\_val\}.\par
\quad - Include the exact numeric value \{scalar\_val\} in the question text\par
\quad - Entity columns to group/filter by: \{entity\_cols\}\par
\par
\# if metric table (buyer\_metrics/seller\_metrics/products/sellers):\par
SQL pattern (metric table - direct WHERE comparison):\par
\quad SELECT DISTINCT \{entity\_example\} FROM \{table\}\par
\quad WHERE metric\_col \{op\} \{scalar\_val\}\par
\par
\# else (raw table):\par
SQL pattern (raw table - GROUP BY + HAVING):\par
\quad SELECT DISTINCT \{entity\_example\} FROM \{table\}\par
\quad WHERE review\_category = '...'          [only when source = reviews]\par
\quad GROUP BY \{entity\_example\}\par
\quad HAVING \{agg\_expr\} \{op\} \{scalar\_val\}
}
\\

\bottomrule
\end{tabularx}
\label{tab:answer_linked_hints}
\end{table*}
\clearpage

%% ── Atomic QA Validation ──────────────────────────────────────────

\begin{table*}[!p]
\centering
\caption{Prompt for SQL Atomic QA Validation.}
\begin{promptbox}{SQL Atomic QA Validation}
\begin{systembox}
\scriptsize
You are a gatekeeper for an atomic QA pair. \\
IMPORTANT CONTEXT: the query was already executed successfully; the shown answer is its actual output.

Domain Context (READ BEFORE JUDGING). \\ This is a RETAIL COMPANY database. Key relationships:
\begin{itemize}[leftmargin=*, noitemsep, topsep=2pt]
\item SELLERS broadcast live shopping shows (scripts) and sell products to BUYERS.
\item Sellers and buyers are completely separate people --- never overlapping.
\end{itemize}

Validator Constitution (applies to ALL cases).
\begin{enumerate}[leftmargin=*, noitemsep, topsep=2pt]
\item NO SPECULATION: never reject based on what ``could/might'' happen (ties, multiple rows, etc.). Judge ONLY the provided Question + Query + Executed Answer.
\item NO STYLE REVIEW: never reject or complain about SQL/search-query style (DISTINCT, GROUP BY, JOIN vs no JOIN, etc.) if it does not clearly change the meaning of the question.
\item REJECT ONLY OBVIOUS SEMANTIC MISMATCH: approve unless there is a clear contradiction between what the question asks and what the query/answer is about.
\end{enumerate}

For SQL atomics (action=sql\_query): TWO criteria.

Criterion 1 --- Semantic Alignment (Q $\leftrightarrow$ SQL $\leftrightarrow$ Answer): Reject only for clear mismatches such as:
\begin{itemize}[leftmargin=*, noitemsep, topsep=2pt]
\item Scope/entity mismatch (question is about seller/buyer/product X but SQL filters different entity or none)
\item Concept mismatch (question asks ``who/which'' but SQL returns a numeric aggregate; asks ``how many'' but SQL returns a name list)
\item Direct contradiction (question says ``most recent/latest'' but SQL clearly selects MIN; says ``highest'' but SQL clearly selects MIN)
\item Schema-impossible claim (using schema to detect: e.g., Buyer\_Metric has no seller\_name/product\_name, or Seller\_Metric has no buyer\_name/product\_name)
\end{itemize}

Criterion 2 --- Realistic \& Safe Question: Reject only if the question:
\begin{itemize}[leftmargin=*, noitemsep, topsep=2pt]
\item Exposes an opaque internal ID (buyer\_id, seller\_id, script\_id, product\_id, etc.)
\item Is nonsensical or self-contradictory
\item Answer leakage: the answer string appears verbatim as a substring of the question.
\end{itemize}

Output JSON only --- list ONLY failed criteria; if approved, \texttt{failed\_criteria} is \texttt{[]}:\\
\texttt{\{"approved": true|false, "failed\_criteria": [], "issues": ["short concrete description of each failure"]\}}\\
Use exact labels \texttt{"criterion\_1"} or \texttt{"criterion\_2"} in failed\_criteria.
\end{systembox}

\begin{userbox}
\scriptsize
Atomic QA:\\
\textcolor{blue}{\{atomic\_qa\_json\}}

Source table schema:\\
\textcolor{blue}{\{source\_schema\}}

Return verdict.
\end{userbox}
\end{promptbox}
\label{tab:atomic_validator_sql}
\end{table*}

\begin{table*}[!p]
\centering
\caption{Prompt for Text Atomic QA Validation.}
\begin{promptbox}{Atomic QA Validation (Text)}
\begin{systembox}
\scriptsize
You are a gatekeeper for a text atomic QA pair. IMPORTANT CONTEXT: the search query was already executed successfully; the shown answer is its actual output.

Domain Context (READ BEFORE JUDGING). \\ This is a RETAIL COMPANY database. Key relationships:
\begin{itemize}[leftmargin=*, noitemsep, topsep=2pt]
\item SELLERS broadcast live shopping shows (scripts) and sell products to BUYERS.
\item Sellers and buyers are completely separate people --- never overlapping.
\end{itemize}

Validator Constitution (applies to ALL cases).
\begin{enumerate}[leftmargin=*, noitemsep, topsep=2pt]
\item NO SPECULATION: judge ONLY the provided Question + Search Query + Evidence + Executed Answer.
\item NO STYLE REVIEW: never reject because the search-query or answer wording is not how you would phrase it.
\item REJECT ONLY CLEAR FAILURES: approve unless there is a clear contradiction, unsupported answer content, or obvious scope drift.
\end{enumerate}

For text atomics (action=dense\_retrieve): THREE criteria.
\begin{itemize}[leftmargin=*, noitemsep, topsep=2pt]
\item Criterion 1 --- Evidence answerability: Using ONLY the provided evidence\_content, determine whether the question can be answered. Reject if the evidence does not contain enough information to answer the question.
\item Criterion 2 --- Evidence grounding / faithfulness: Reject if the answer contains material information that is not explicitly supported by the evidence\_content, or if it contradicts the evidence.
\item Criterion 3 --- Scope sanity: If the question names an entity (seller X, buyer Y, product Z), the search\_query should include it. Reject ONLY if the entity is missing AND the answer is clearly about a different entity.
\item Criterion 4 --- Answer leakage: Reject if the answer string appears verbatim as a substring of the question.
\end{itemize}

Output JSON only --- list ONLY failed criteria; if approved, \texttt{failed\_criteria} is \texttt{[]}:\\
\texttt{\{"approved": true|false, "failed\_criteria": [], "issues": ["short concrete description of each failure"]\}}\\
Use exact labels \texttt{"criterion\_1"}, \texttt{"criterion\_2"}, \texttt{"criterion\_3"}, or \texttt{"criterion\_4"} in failed\_criteria.
\end{systembox}

\begin{userbox}
\scriptsize
Atomic QA:\\
\textcolor{blue}{\{atomic\_qa\_json\}}

Source table schema:\\
\textcolor{blue}{\{source\_schema\}}

Return verdict.
\end{userbox}
\end{promptbox}
\label{tab:atomic_validator_text}
\end{table*}
\clearpage

%% ── Composite Question Composition via Referring-Expression Substitution ───
\begin{table*}[!p]
\centering
\caption{Prompt for Composite Question Composition via Referring-Expression Substitution, shared by the entity-linked and answer-linked (Set, S) pipelines. The \texttt{\{entity\}} field of the user prompt is filled with an entity name, a set placeholder, or a scalar value depending on the variant.}
\begin{promptbox}{Composite Question Composition via Referring-Expression Substitution}
\begin{systembox}
\scriptsize
You are a multi-hop QA composer. You take a target question and rewrite it by replacing exactly one named entity with a natural referring expression derived from an identifier question whose answer is that entity. Your goal is one fluent, unambiguous interrogative sentence that preserves meaning.

Preserve the target meaning and answer. Use the identifier only to identify the replaced entity.\\
 
Rules:
\begin{itemize}[leftmargin=*, noitemsep, topsep=2pt]
\item Replace only the specified entity; preserve all other names, products, dates, times, numbers, and metrics exactly.
\item Keep the replaced entity's role unchanged: seller/broadcaster, buyer/reviewer, product, date/time, or value.
\item Do not let identifier-side products, reviews, broadcasts, dates, or times become confused with the target object/event.
\item If direct substitution creates ambiguity, rewrite the local phrase instead of inserting the referring expression in place.
\item If the entity is immediately followed by a ``that ...'' or ``where ...'' clause in the target question, do NOT simply replace the entity name inline. Instead restructure the whole phrase so the ``that''/``where'' clause clearly attaches to the correct head noun, not to the inserted referring expression. Example: ``the product sold by X that generated the highest sales'' $\rightarrow$ ``the product that generated the highest sales, sold by [referring expression]''.
\item If the entity appears in a possessive position (e.g. ``X's orders''), do NOT produce a sub-clause ending in a participle followed by ``'s''. Instead rewrite using a prepositional form: e.g. ``orders placed by [referring expression]'' or ``orders from [referring expression]''.
\item Put target dates/times next to the target event, not after a long referring expression.
\item When the target question already contains the main action involving another entity, keep that target-side action intact and make the referring expression adjectival or event-based rather than repeating the same action frame. For example, avoid patterns like ``the buyer who asked Steven C. about ... ask Steven C. about ...''; instead use shorter phrases such as ``the buyer who made a pre-purchase inquiry about ...'' or ``the buyer who asked about ...''.
\item If the result is not one clear natural question, return \texttt{\{"combinable": false\}}.
\end{itemize}

Output JSON only: \texttt{\{"combinable": true, "question": "..."\}} or \texttt{\{"combinable": false\}}
\end{systembox}

\begin{userbox}
\scriptsize
Target question: \textcolor{blue}{\{target\_question\}}\\
Target answer: \textcolor{blue}{\{target\_answer\}}

Identifier question: \textcolor{blue}{\{identifier\_question\}}\\
Identifier answer: \textcolor{blue}{\{identifier\_answer\}}

Entity to replace in target: ``\textcolor{blue}{\{entity\}}''

Rewrite the target question in english.

\textit{[For answer-linked level $\geq$2, the following cleft instruction is appended:]}\\
\textit{Rewrite as one natural English question. The entity name must not appear --- replace it fully with the referring expression. Avoid deeply nested noun phrases; restructure the sentence if needed.}
\end{userbox}
\end{promptbox}
\label{tab:obfuscate}
\end{table*}

\begin{table*}[!p]
\centering
\caption{Prompt for Composite QA Validation.}
\begin{promptbox}{Composite QA Validation}
\begin{systembox}
\scriptsize
You are a strict auditor for multi-hop composite QA built by entity substitution.  You judge only substitution quality --- meaning preservation, grammaticality, and unambiguity --- and you reject patterns that other LLMs tend to overlook. The composite question was built by replacing one entity in the target question with a referring expression derived from an earlier chain step (the ``replacement step'').

IMPORTANT: The composite question may intentionally reference multiple different products, events, dates, or entities. Some are only used to identify an intermediate entity, while another is the actual subject of the final answer. Do NOT reject merely because the identifier-side product/entity differs from the target-side product/entity.

Judge ONLY the substitution quality:
\begin{enumerate}[leftmargin=*, noitemsep, topsep=2pt]
\item Naturalness: the composite question must be a grammatically correct interrogative sentence, with no malformed possessives, no awkward repetition, no unclear pronouns, and no cataphora (pronoun preceding its named antecedent in the same sentence).
\item Substitution accuracy: the inserted referring expression must identify the obfuscated entity using the replacement step's question and answer. It is OK if the expression mentions identifier-side products/events/dates, but reject if they are grammatically confused with the final target object/event or if date/time/product clauses attach to the wrong part of the question.
\item Preservation: all other names, products, dates, times, numbers, and metrics from the target question must be preserved exactly, except the replaced entity itself.
\end{enumerate}

Approve if the question is long primarily because target product names are preserved exactly, as long as it remains grammatical, faithful, and unambiguous. Do not reject solely because a required preserved product name is long.

Reject if the rewrite changes the target meaning, adds unsupported constraints, drops target constraints, reveals the target answer through the referring expression, or makes the substitution ambiguous.

\textbf{Output JSON only:} \texttt{\{"approved": true|false, "issues": ["criterion N: reason"]\}}
\end{systembox}

\begin{userbox}
\scriptsize
Composite QA:\\
\textcolor{blue}{\{composite\_qa\_json\}}

Target question before rewrite (the question that was rewritten to produce the composite):\\
\textcolor{blue}{\{target\_question\_before\_rewrite\}}

Replacement step (the identifier question whose answer replaces the obfuscated entity):\\
\textcolor{blue}{\{replacement\_step\_json\}}

Obfuscation context:\\
\textcolor{blue}{\{obfuscation\_context\}}

Return verdict.
\end{userbox}
\end{promptbox}
\label{tab:composite_validator}
\end{table*}
\clearpage

\begin{table*}[!p]
\centering
\caption{Prompt for Auto-Schema Briefing.}
\begin{promptbox}{Auto-Schema Briefing}
\begin{systembox}
\scriptsize
You produce a planner brief from an inspection JSON of one or more data stores. The brief is read by a downstream SQL planner at every query. Output plain text, with no JSON and no fences. Use exact store, table, column, and value names from the inspection; never invent.

Required sections, in this order:

\noindent[Stores]\\
For each store, one line: ``\texttt{[SQL] <id> --- short description}'' or ``\texttt{[TEXT] <id> --- short description}''. \\
Under each SQL store, list every column with one of: 
\begin{itemize}[leftmargin=*, noitemsep, topsep=2pt]
    \item ``\texttt{<col>} $\in$ \texttt{\{v1, v2, ...\}}'' (enum, exact \texttt{distinct\_values});
    \item ``\texttt{<col> (numeric)}'';
    \item ``\texttt{<col> (e.g. <s1>, <s2>, <s3>)}'' (freeform, \texttt{sample\_values});
    \item ``\texttt{<col> (id-shaped)}'' (high-cardinality identifier).
\end{itemize}
Under each TEXT store, one sentence covering topic + distinctive vocabulary.

\noindent[Enums]\\
One line per categorical column:
\texttt{<table>.<col>} $\in$ \texttt{\{v1, v2, ...\}}.

\noindent[Entity-grouping columns]\\
PRIMARY: columns ending in \texttt{\_name}, \texttt{\_label}, or \texttt{\_title} that appear in two or more tables.\\
SECONDARY: columns ending in \texttt{\_id} that appear in two or more tables (join keys, paired with the corresponding \texttt{\_name} column when one exists).

Format:
\begin{itemize}[leftmargin=*, noitemsep, topsep=2pt]
    \item \texttt{PRIMARY:}
    \begin{itemize}[leftmargin=*, noitemsep, topsep=1pt]
        \item \texttt{<col>  (in: <table1>, <table2>, ...)}
    \end{itemize}
    \item \texttt{SECONDARY:}
    \begin{itemize}[leftmargin=*, noitemsep, topsep=1pt]
        \item \texttt{<id\_col>  (in: <tables>)  paired with <name\_col>}
    \end{itemize}
\end{itemize}

\noindent[Derived KPIs]\\
SQL formulas auto-derived from the schema. Use the 4 templates below; substitute placeholders with real names from the inspection. Each KPI is scoped to ONE most-natural entity column -- do not replicate the same KPI across every entity column.

CRITICAL: In every template, \texttt{<entity\_col>} MUST come from the PRIMARY list in [Entity-grouping columns] AND MUST be a column that actually exists in the \texttt{<table>} being aggregated. It MUST NOT be the same column as the \texttt{<enum\_col>} or \texttt{<numeric\_col>} being aggregated. For example, \texttt{refunded\_rate(status) = ... WHERE status='<X>'} is wrong because the filter column equals the enum column. \texttt{avg\_quantity(product\_category) = AVG(quantity) FROM Order WHERE product\_category='<X>'} is wrong because \texttt{Order} has no \texttt{product\_category} column. Emit \texttt{refunded\_rate(seller\_name) = ... WHERE seller\_name='<X>'} instead.

\begin{enumerate}[leftmargin=*, noitemsep, topsep=2pt]
    \item ENUM-RATE --- one KPI per enum value:

    \texttt{<value>\_rate(<entity\_col>) =}\\
    \hspace*{1em}\texttt{SUM(CASE WHEN <enum\_col>='<value>' THEN 1 ELSE 0 END) * 100.0 / COUNT(*)}\\
    \hspace*{1em}\texttt{FROM <table> WHERE <entity\_col>='<X>'}

    \item NUMERIC AGGREGATE --- avg, total, max, and min per numeric column:

    \texttt{avg\_<numeric\_col>(<entity\_col>) = AVG(<numeric\_col>) FROM <table> WHERE <entity\_col>='<X>'}\\
    \texttt{total\_<numeric\_col>(<entity\_col>) = SUM(<numeric\_col>) FROM <table> WHERE <entity\_col>='<X>'}\\
    \texttt{max\_<numeric\_col>(<entity\_col>) = MAX(<numeric\_col>) FROM <table> WHERE <entity\_col>='<X>'}\\
    \texttt{min\_<numeric\_col>(<entity\_col>) = MIN(<numeric\_col>) FROM <table> WHERE <entity\_col>='<X>'}

    \item NUMERIC $\times$ ENUM CROSS --- emit only when semantically meaningful:

    \texttt{avg\_<numeric\_col>\_<enum\_value>(<entity\_col>) = AVG(<numeric\_col>) FROM <table>}\\
    \hspace*{1em}\texttt{WHERE <entity\_col>='<X>' AND <enum\_col>='<enum\_value>'}

    \item CROSS-TABLE CONVERSION --- when two tables share a foreign-key column:

    \texttt{conversion\_rate(<entity\_col>) =}\\
    \hspace*{1em}\texttt{COUNT(DISTINCT child.<shared\_id>) * 100.0 / COUNT(DISTINCT parent.<shared\_id>)}\\
    \hspace*{1em}\texttt{FROM <parent\_table> parent LEFT JOIN <child\_table> child}\\
    \hspace*{1em}\texttt{ON parent.<shared\_id> = child.<shared\_id>}\\
    \hspace*{1em}\texttt{WHERE parent.<entity\_col>='<X>'}
\end{enumerate}

\noindent[Conventions]
\begin{itemize}[leftmargin=*, noitemsep, topsep=2pt]
    \item Date columns: filter with \texttt{LIKE 'YYYY-MM-DD\%'} or \texttt{DATE(col)='YYYY-MM-DD'}.
    \item Long freeform string columns: use \texttt{LIKE '\%keyword\%'} on one to three distinctive words.
    \item For rank-N by KPI queries: compute the KPI per entity in a CTE, then \texttt{ORDER BY ... LIMIT N}. Do not pre-filter by ``above average''.
    \item Use enum values verbatim from [Enums]; do not paraphrase or pluralize.
\end{itemize}
\end{systembox}

\begin{userbox}
\scriptsize
Stores inspection (raw JSON): \textcolor{blue}{\{inspection\_json\}}

Domain knowledge (optional, append verbatim under header): \textcolor{blue}{\{kb\_text\}}

Produce the brief now. List EVERY table that appears in the inspection --- do not skip any.
\end{userbox}
\end{promptbox}
\label{tab:auto_schema_briefing}
\end{table*}
\clearpage

\begin{table*}[!p]
\centering
\caption{Prompt for Adaptive Complexity Routing.}
\begin{promptbox}{Adaptive Complexity Router}
\begin{systembox}
\scriptsize
You classify a retail QA query's reasoning complexity. Output ONE label.

\noindent
simple --- 1 hop. The answer is obtainable from a single source via ONE SQL/retrieval step. No bridge entity, no chaining. Plannable as a single atomic step.

\noindent
hard --- 2+ hop chain. The query requires at least one bridge entity (``the seller of the product that buyer X ordered''), or multiple aggregations across an entity chain. Needs 2 or more atomic steps.

Output JSON only:\\
\texttt{\{ "label": "simple"|"hard", "why": "one short reason" \}}
\end{systembox}

\begin{userbox}
\scriptsize
Query:\textcolor{blue}{\{query\}}

Classify.
\end{userbox}
\end{promptbox}
\label{tab:complexity_router}
\end{table*}

\begin{table*}[!p]
\centering
\caption{Prompt for SQL Source Picker}
\begin{promptbox}{SQL Source Picker}
\begin{systembox}
\scriptsize
You decide which retail SQL databases must be queried to answer a sub-question.

Available databases (logical names, each backed by a separate sqlite file):
\begin{itemize}[leftmargin=*, noitemsep, topsep=2pt]
    \item \texttt{products} --- table \texttt{Product(product\_id, product\_name, product\_category, original\_price, store, features)}
    \item \texttt{scripts} --- table \texttt{Script(script\_id, seller\_id, seller\_name, product\_id, product\_name, broadcast\_datetime, discount\_rate, original\_price, discount\_price, discount\_end\_date, shipping\_fee)}
    \item \texttt{orders} --- table \texttt{"Order"(order\_id, script\_id, buyer\_id, buyer\_name, product\_id, product\_name, quantity, total\_price, status, order\_date, delivery\_date, post\_inquiry\_type, seller\_id, seller\_name)}
    \item \texttt{inquiries} --- table \texttt{Inquiry(inquiry\_id, inquiry\_type, buyer\_id, buyer\_name, seller\_id, seller\_name, product\_id, product\_name, script\_id)}
    \item \texttt{reviews} --- table \texttt{Review(review\_id, review\_category, review\_rating, buyer\_id, buyer\_name, seller\_id, seller\_name, product\_id, product\_name, script\_id, inquiry\_id)}
\end{itemize}

Rules:
\begin{itemize}[leftmargin=*, noitemsep, topsep=2pt]
    \item Pick the MINIMUM set that can answer the sub-question. Do NOT pick all five ``to be safe''.
    \item If two tables must be JOIN'd (e.g. \texttt{products} + \texttt{orders}), include both.
    \item Output JSON: \texttt{\{"sources": ["orders", "reviews"], "why": "one short reason"\}}
    \item \texttt{sources} MUST be a non-empty subset of the five names above.
\end{itemize}
\end{systembox}

\begin{userbox}
\scriptsize
Sub-question:\textcolor{blue}{\{sub\_query\}}

Output JSON.
\end{userbox}
\end{promptbox}
\label{tab:sql_source_pick}
\end{table*}

\begin{table*}[!p]
\centering
\caption{Prompt for Planning.}
\vspace*{-0.2cm}
\begin{promptbox}{Planning}
\begin{systembox}
\fontsize{6.5pt}{7.5pt}\selectfont
\setlength{\columnsep}{8pt}
\begin{multicols}{2}

Decompose the user query into an ordered list of ATOMIC evidence-gathering steps that solve it ONE PIECE AT A TIME.

\noindent RULE 0 --- ENTITY RESOLUTION FIRST

Before writing the plan, mentally list every named entity in the query (persons, long product/work names, specific datetimes, places).

Your plan's FIRST steps must be DEDICATED entity-resolution lookups for those entities --- one entity per step, no other filter or aggregation. Only AFTER all entities are resolved do you write the steps that use those resolved IDs/values as filters.

A resolution step has a SINGLE simple WHERE clause (for SQL) or a short direct query (for \texttt{dense\_retrieve}). NEVER combine an entity lookup with ranking or aggregation in one step.

\noindent RULE 1 --- Atomic decomposition

Each step answers ONE simple sub-question with at most ONE filter, ONE join, OR ONE aggregation. An n-hop query needs exactly n atomic steps --- err on MORE granular, an extra simple step is far better than a compound failing one. The LAST step's output IS the final answer. Do NOT add a separate ``answer'' / ``summarize'' step.

\noindent RULE 2 --- Step actions

Exactly TWO action types are allowed:
\begin{itemize}[leftmargin=*, noitemsep, topsep=2pt]
    \item \texttt{sql\_query} --- single-table or 1-JOIN SQL with at most 1--2 WHERE conditions and at most ONE aggregation/ranking. Each \texttt{sql\_template} is a SINGLE valid SELECT (no DDL/DML, no comments, and no trailing semicolon).
    \item \texttt{dense\_retrieve} --- semantic search over text passages. Each \texttt{search\_template} is a short natural-language query ($\leq$ 20 words).
\end{itemize}

Pick the action that best matches the data store carrying the answer. When the answer lives in SQL columns, prefer \texttt{sql\_query}. When the answer is narrative / sentiment / unstructured, prefer \texttt{dense\_retrieve}.

\noindent RULE 3 --- Placeholder substitution (deterministic chain)

Use \texttt{\{\{prev\_N\}\}} placeholders to reference prior step N's output. At execution time, \texttt{\{\{prev\_N\}\}} is substituted (in Python, NOT by an LLM) with the literal value(s) step N produced --- already SQL-escaped.

Forms:
\begin{itemize}[leftmargin=*, noitemsep, topsep=2pt]
    \item \texttt{\{\{prev\_N\}\}} --- comma-separated SQL literals from step N's primary column.
    \item \texttt{\{\{prev\_N.col\}\}} --- extract a specific column from step N.
    \item single-value filter: \texttt{WHERE col = \{\{prev\_N\}\}}
    \item multi-value filter: \texttt{WHERE col IN (\{\{prev\_N\}\})}
    \item \texttt{dense\_retrieve}: \texttt{\{\{prev\_N\}\}} expands to comma-separated values.  (no SQL quoting). Also valid inside the `metadata\_filter` dict values (see RULE 3b).
\end{itemize}

NEVER copy literal IDs/values from earlier steps into the SQL --- always go through \texttt{\{\{prev\_N\}\}}.

\noindent RULE 3b --- metadata\_filter on \texttt{dense\_retrieve} (hard entity binding)

A \texttt{dense\_retrieve} step MAY include a \texttt{metadata\_filter} dict that hard-restricts the candidate pool BEFORE embedding similarity. Use this when the prior step resolved an entity that the answer must match exactly:

\noindent
\texttt{"metadata\_filter": \{\{ "buyer\_name": "\{\{prev\_1.buyer\_name\}\}" \}\}}

Allowed keys are the metadata fields of the chosen TEXT store (e.g. \texttt{buyer\_name}, \texttt{seller\_name}, \texttt{product\_name}, \texttt{inquiry\_type}, \texttt{review\_category}, \texttt{review\_rating}). Multiple keys are ANDed.

Use \texttt{metadata\_filter} for HARD constraints already resolved upstream (the buyer who did X in step 1 $\to$ restrict step 2 to that buyer). 

\columnbreak

DO NOT put descriptive constraints there (``the buyer who mentioned `soup energy''') --- those belong in \texttt{search\_template}. If the filter value is wrong/missing the call returns 0 hits, so only filter on values you are confident were resolved.

\noindent RULE 4 --- Final-step answer slot

The FINAL step's output column type must match the user query's wh-word:
\begin{itemize}[leftmargin=*, noitemsep, topsep=2pt]
    \item ``who / which seller / buyer'' $\to$ name column (NOT id)
    \item ``how many'' $\to$ \texttt{COUNT(*)}
    \item ``what is the total / average'' $\to$ \texttt{SUM}/\texttt{AVG} aggregate
    \item ``what date / when'' $\to$ date column
    \item PLURAL final answer (query uses ``which products / categories / buyers / texts'', ``list'', ``all'', ``each'', ``every'', ``for each''): the FINAL step must NOT use \texttt{LIMIT 1} --- return EVERY matching row.
\end{itemize}

If the query asks to identify an entity by NAME, the FINAL step's SELECT must surface the human-readable name column, never the internal id.

\noindent RULE 5 --- KPI vocabulary (use [Derived KPIs] formulas verbatim)

The DATA STORES brief below contains a [Derived KPIs] section auto-derived from the schema. When the user query asks for ANY metric (rate, average, total, count-by-X, conversion, ranking by score), find the matching KPI in [Derived KPIs] and copy its formula VERBATIM into the SQL --- substituting only the entity literal (the \texttt{'<X>'} placeholder).

NEVER invent your own KPI definition (e.g. \texttt{status='success'} for \texttt{delivered\_rate}, or ``above average price'' instead of computing the KPI then ranking). NEVER shorten or simplify a KPI formula. NEVER use a column or enum value not in the [Enums] section.

For ``rank-N by <KPI>'' queries, compute the KPI per entity in a CTE then ORDER BY the KPI with LIMIT N.

\noindent DATA STORES (auto-generated; trust these names verbatim)

\textcolor{blue}{\{db\_block\}}

Routing contract (per step): every step must declare which stores it queries. STORE IDENTIFIERS ARE LITERAL --- copy the exact ``[SQL]'' / ``[TEXT]'' identifiers from the DATA STORES block verbatim. Do NOT paraphrase (e.g. ``reviews\_text'' --- never ``Review Corpus'', ``reviews'', or ``Reviews''). A wrong identifier triggers SQL ROUTING ERROR / unknown source.

Use ``sources'' (a list, e.g. \texttt{["orders","reviews"]}) when an SQL step needs to JOIN across multiple SQL DBs; use ``source'' (a single string) for single-store steps. Pick the MINIMUM set of stores that answers the step. Cross-DB JOIN is only allowed across stores listed in ``sources'' for that step. If a step's SQL must reference a table from another store, that store MUST be declared in ``sources'' --- otherwise the call fails with SQL ROUTING ERROR.

Output JSON only:\\
{\scriptsize
\ttfamily
\noindent
\{\{ "steps": [ \{\{ "step": 1,\\
"sub\_query": "short natural-language description (for trace only)",\\
"action": "sql\_query" | "dense\_retrieve",\\
"source": "<one store name>", "sources": ["<store\_a>", "<store\_b>"] | null,\\
"sql\_template": "SELECT ...   (with \{\{prev\_N\}\} placeholders)" | null,\\
"search\_template": "free-text query with \{\{prev\_N\}\}" | null,\\
"metadata\_filter": null | \{\{"<meta\_key>": "<value or \{\{prev\_N.col\}\}>"\}\},\\
"depends\_on": null | [<step ids>]\\
\}\} ],\\
"rationale": "one short paragraph" \}\}
\par}

\end{multicols}
\end{systembox}

\begin{userbox}
\fontsize{6.5pt}{7.5pt}\selectfont
User query: \textcolor{blue}{\{query\}}

Step 1 of your thinking: list every named entity in the query (persons, dates, long product/work names) before deciding the plan order. Resolution steps for those entities come FIRST.

Output the plan JSON only (no entity-list section).
\end{userbox}
\end{promptbox}
\label{tab:planning}
\end{table*}

\begin{table*}[!p]
\centering
\caption{Prompt for SQL Query Generation in the SQL Tool.}
\begin{promptbox}{SQL Query Generation in the SQL Tool}
\begin{systembox}
\scriptsize
SQL writer.

Tables, enums, and auto-derived KPI formulas are in the DATA STORES brief below. Use the EXACT names listed there --- case-sensitive, no plural / singular flips.

Generic rules:
\begin{itemize}[leftmargin=*, noitemsep, topsep=2pt]
    \item ONE SELECT (or WITH...SELECT). No DDL/DML.
    \item If a table name is a SQL reserved word, quote it with double-quotes (NOT backticks).
    \item Use only listed columns. Always spell out the full column name from the schema after a table alias --- single-letter or two-letter abbreviations after a dot are never valid.
    \item Cross-table $\to$ JOIN on shared id columns (whatever IDs appear in both).
    \item Use enum values exactly as listed (case + spacing). Do not paraphrase.
    \item Return ONLY SQL inside \texttt{\textasciigrave\textasciigrave\textasciigrave sql ... \textasciigrave\textasciigrave\textasciigrave}.
\end{itemize}

Filter precision:
\begin{itemize}[leftmargin=*, noitemsep, topsep=2pt]
    \item DATETIME: never match exact seconds/minutes. Use one of:\\
    \texttt{DATE(col) = 'YYYY-MM-DD'}\\
    \texttt{col LIKE 'YYYY-MM-DD\%'}\\
    \texttt{col LIKE 'YYYY-MM-DD HH:\%'}
    \item LONG STRING FILTERS (long strings with delimiters or punctuation): use \texttt{col LIKE '\%distinctive-keyword\%'} on 1--3 most unique words instead of exact equality.
    \item SHORT NAMES / IDs: use exact match.
    \item If the question references a SPECIFIC entity, your WHERE clause MUST filter by that entity.
\end{itemize}

KPI derivation:
\begin{itemize}[leftmargin=*, noitemsep, topsep=2pt]
    \item Use the auto-derived KPI formulas in [Derived KPIs] VERBATIM, substituting only the entity literal. Do not invent your own metric definitions.
    \item For ``rank-N by KPI'' queries, compute the KPI per entity in a CTE then ORDER BY the KPI with LIMIT N.
\end{itemize}

\noindent DATA STORES (auto-derived; trust these names verbatim)

\textcolor{blue}{\{db\_block\}}
\end{systembox}
\begin{userbox}
\scriptsize
Sub-question: \textcolor{blue}{\{sub\_query\}}

ROUTING: This call is restricted to target\_sources=\textcolor{blue}{\{target\_sources\}}.\\ Use ONLY tables from these sources: \textcolor{blue}{\{schema\_block\_for(target\_sources)\}}

Cross-DB JOIN is allowed only across the selected sources.

Upstream context: \textcolor{blue}{\{previous\_steps\}}

Write the SQL now. Output JSON.
\end{userbox}
\end{promptbox}
\label{tab:sql_tool}
\end{table*}

\begin{table*}[!p]
\centering
\caption{Prompt for Step Diagnosis.}
\begin{promptbox}{Step Diagnosis}
\begin{systembox}
\scriptsize
A plan step did not produce a usable result. Identify the single most likely root cause.

Causes (SQL-side):
\begin{itemize}[leftmargin=*, noitemsep, topsep=2pt]
    \item \texttt{schema\_error} --- wrong table or column name was used.
    \item \texttt{filter\_too\_narrow} --- WHERE clause matched 0 rows.
    \item \texttt{empty\_upstream} --- a depends\_on step returned nothing.
    \item \texttt{wrong\_source} --- the answer lives in a different store.
    \item \texttt{target\_sources\_too\_narrow} --- the SQL referenced a table whose store is NOT in this step's \texttt{sources} list. Status will start with ``SQL ROUTING ERROR''.
    \item \texttt{other} --- describe briefly.
\end{itemize}

Causes (TEXT-side):
\begin{itemize}[leftmargin=*, noitemsep, topsep=2pt]
    \item \texttt{query\_too\_specific} --- search query was too narrow; 0 hits or off-topic.
    \item \texttt{query\_too\_generic} --- search query missed the key entity; hits drift.
    \item \texttt{wrong\_text\_source} --- picked the wrong text store (multiple available).
    \item \texttt{insufficient\_evidence} --- top hits exist but don't contain the answer.
\end{itemize}

Output JSON only:\\
{\ttfamily
\noindent
\{\{ "cause": "schema\_error"|"filter\_too\_narrow"|"empty\_upstream"|"wrong\_source"|\\
"target\_sources\_too\_narrow"|"query\_too\_specific"|"query\_too\_generic"|\\
"wrong\_text\_source"|"insufficient\_evidence"|"other",\\
"explanation": "one sentence" \}\}
\par}
\end{systembox}
\begin{userbox}
\scriptsize
Step: \textcolor{blue}{\{step\_json\}}\\
action: \textcolor{blue}{\{action\}}\\
output\_size: \textcolor{blue}{\{N | 'scalar'\}}\\
status: \textcolor{blue}{\{status\}}\\
auto\_reason: \textcolor{blue}{\{auto\_reason\}}
\end{userbox}
\end{promptbox}
\label{tab:diagnose_template}
\end{table*}

\begin{table*}[!p]
\centering
\caption{Prompt for Plan Revision.}
\begin{promptbox}{Plan Revision}
\begin{systembox}
\scriptsize
A step in the plan failed. Rewrite the plan from the failed step onward (the failed step + all subsequent steps). Prior successful steps are kept as-is.

Each revised step MUST follow the same \texttt{sql\_template} / \texttt{search\_template} + \texttt{\{\{prev\_N\}\}} placeholder schema as the original plan. Dense-retrieve steps may also include \texttt{metadata\_filter} (see PLAN\_TEMPLATE RULE 3b) --- use it to hard-bind an entity already resolved by a prior step.

Common fixes by cause:
\begin{itemize}[leftmargin=*, noitemsep, topsep=2pt]
    \item \texttt{schema\_error} --- use the correct table / column name (see DATA STORES below).
    \item \texttt{filter\_too\_narrow} --- relax the WHERE clause; LIKE '\%...\%' or drop a restrictive condition.
    \item \texttt{empty\_upstream} --- find an alternative bridge entity, or fold upstream logic into a single JOIN.
    \item \texttt{wrong\_source} --- switch to a different store from DATA STORES.
    \item \texttt{target\_sources\_too\_narrow} --- expand the step's \texttt{sources} list to include the missing store, OR rewrite the SQL to use a table that IS already in \texttt{sources}. Do not silently drop the reference.
    \item \texttt{query\_too\_specific} --- broaden the search\_template (drop a keyword).
    \item \texttt{query\_too\_generic} --- add the entity name from the user query into the search\_template.
    \item \texttt{wrong\_text\_source} --- switch to a different TEXT store from DATA STORES that better fits the question topic.
    \item \texttt{insufficient\_evidence} --- re-formulate the search\_template to target the missing fact (e.g., add the relation word from the user query).
\end{itemize}

DATA STORES (use these names verbatim): \textcolor{blue}{\{db\_block\}}

Output JSON only:\\
{\scriptsize
\ttfamily
\noindent
\{\{ "new\_remaining\_steps": [<step objects>], "rationale": "one short sentence" \}\}
\par}
\end{systembox}

\begin{userbox}
\scriptsize
User query: \textcolor{blue}{\{query\}}\\
Failed step (index \textcolor{blue}{\{index\}}): \textcolor{blue}{\{failed\_step\}}\\
Diagnose: \textcolor{blue}{\{diagnosis\}}\\
Prior outputs (compact): \textcolor{blue}{\{prior\_step\_outputs\}} \\ 
Original full plan: \textcolor{blue}{\{plan\}}
\end{userbox}
\end{promptbox}
\label{tab:plan_revision}
\end{table*}

\begin{table*}[!p]
\centering
\caption{Prompt for Final Answer Generation.}
\begin{promptbox}{Final Answer Generation}
\begin{systembox}
\scriptsize
Produce the final answer to the user query from the executed plan's outputs.

How to choose the answer:
\begin{enumerate}[leftmargin=*, noitemsep, topsep=2pt]
    \item Identify what TYPE of value the query asks for (a name, date, number, count, passage). Call this the ``answer slot''.
    \item Walk the steps in order. Identify which step's INTENT matches the answer slot --- that step's OUTPUT is the answer.
    \item SINGULAR vs PLURAL --- decide BEFORE writing the answer:
    \begin{itemize}[leftmargin=*, noitemsep, topsep=1pt]
        \item SINGULAR cues: ``the highest'', ``the most'', ``which one'', aggregations. Pick ONE row.
        \item PLURAL cues: ``which products / categories / buyers / texts / reviews'', ``list'', ``all'', ``each'', ``every'', ``for each''. Return ALL matching rows as a comma-joined list.
    \end{itemize}
\end{enumerate}

Output style --- STRICT:
\begin{itemize}[leftmargin=*, noitemsep, topsep=2pt]
    \item ``answer'' is a STRING. For singular $\to$ bare value (name, number, date, short phrase). For plural $\to$ comma-joined bare values. No prefix, no full sentences.
    \item For dates use the format that appears in the context.
    \item For numbers return a bare number.
    \item NEVER return an internal ID (FOO-\ldots, ORD-\ldots) when the query asks for a name. Cross-look earlier steps for the corresponding name.
\end{itemize}

Dense-retrieve outputs (rule for using retrieved docs):
\begin{itemize}[leftmargin=*, noitemsep, topsep=2pt]
    \item Each retrieved doc has two parts: metadata (structured fields: \texttt{buyer\_name}, \texttt{seller\_name}, \texttt{product\_name}, \texttt{review\_category}, \texttt{review\_rating}, \texttt{broadcast\_datetime}, \texttt{inquiry\_type}, etc.) and content (free-form narrative text like review text, dialogue, broadcast script).
    \item Decide where the answer lives BEFORE looking. If the answer slot is a NAME / ENTITY / CATEGORY / RATING / DATE (a structured field), read it from the metadata of the single best-matching doc --- NOT from the content text body. If the answer slot is a QUOTE / OPINION / EXPLANATION / REASON / DESCRIPTION (free-form prose), read it from the content text body of the relevant doc --- quote or paraphrase the relevant span; never return an identifier-shaped token.
    \item When multiple persons appear in the same passage (e.g., a review written BY buyer X about seller Y for product Z), use the metadata role labels to pick the entity that matches the question's role.
    \item The answer is ALWAYS user-facing prose or a metadata value. Never return a list of capitalized identifier tokens (e.g. FOO-XXX-0001).
\end{itemize}

INSUFFICIENT is a LAST-RESORT fallback. Return it only when NO step produced any useful output.

Output JSON only:\\
{\scriptsize
\ttfamily
\noindent
\{\{ "answer": "...",\\
"evidence": [\{\{"step": <int>, "source": "...", "snippet": "..."\}\}],\\
"confidence": "high"|"medium"|"low" \}\}
\par}
\end{systembox}
\begin{userbox}
\scriptsize
User query: \textcolor{blue}{\{query\}}

Executed plan:\\
{}[step \textcolor{blue}{\{n\}}] INTENT: \textcolor{blue}{\{sub\_query\}}\\
\hspace*{2em}STATUS: \textcolor{blue}{\{status\}}\\
\hspace*{2em}OUTPUT (\textcolor{blue}{\{type\}}): \textcolor{blue}{\{step\_output\}}\\
\textcolor{blue}{\{\dots one block per executed step \dots\}}

Produce the final JSON answer.
\end{userbox}
\end{promptbox}
\label{tab:final_answer_generation}
\end{table*}

\end{document}